\documentclass[11pt, a4paper]{article}

\usepackage[utf8]{inputenc}          
\usepackage[T1]{fontenc}             
\usepackage{lmodern}                 
\usepackage{microtype}               
\usepackage{appendix}
\usepackage[
    a4paper,
    top    = 25mm,
    bottom = 25mm,
    left   = 25mm,
    right  = 25mm
]{geometry}

\usepackage{setspace}                

\usepackage{amsmath}                 
\usepackage{amssymb}                 
\usepackage{amsthm}                  
\usepackage{mathtools}               
\usepackage{bm}                      

\usepackage{algorithm}
\usepackage{algpseudocode}           
\algnewcommand{\LineComment}[1]{\State \(\triangleright\) #1}

\usepackage{graphicx}                
\usepackage{subcaption}              
\usepackage[
    margin   = 8pt,
    font     = small,
    labelfont= bf
]{caption}                           

\usepackage{booktabs}                
\usepackage{multirow}                
\usepackage{array}                   
\usepackage{makecell}                
\usepackage{array}       
\usepackage{booktabs}
\usepackage{longtable}

\usepackage[dvipsnames, table]{xcolor}

\definecolor{deepblue}   {RGB}{0,   51,  102}   
\definecolor{lightblue}  {RGB}{220, 235, 255}   
\definecolor{bestcell}   {RGB}{200, 225, 255}   
\definecolor{olive}      {RGB}{107, 142,  35}   

\usepackage{siunitx}
\usepackage{lineno}

\usepackage{enumitem}                

\usepackage{verbatim}                
\usepackage{url}                     
\usepackage{textcomp}                

\usepackage[
    backend     = biber,
    style       = numeric-comp,
    sorting     = none,
    giveninits  = true,
    maxbibnames = 99,
    doi         = true,
    url         = false,
    eprint      = false,
    backref     = true
]{biblatex}
\usepackage[
    colorlinks  = true,
    linkcolor   = NavyBlue,
    citecolor   = NavyBlue,
    urlcolor    = NavyBlue,
    pdftitle    = {Physics-Informed Progressive Transfer Learning for
                   Anisotropic Permeability Tensor Prediction in Porous Media},
    pdfauthor   = {Mohammad Nooraiepour},
    pdfkeywords = {permeability tensor prediction; porous media;
                   physics-informed machine learning; hybrid CNN-Transformer;
                   progressive transfer learning; anisotropic permeability}
]{hyperref}

\usepackage{cleveref}                

\theoremstyle{definition}

\title{\Large\textbf{Anisotropic Permeability Tensor Prediction 
from Porous Media Microstructure via Physics-Informed Progressive 
Transfer Learning with Hybrid CNN-Transformer}}

\author{%
    \textbf{Mohammad Nooraiepour}$^{1,\ast}$\\[0.6em]
    \small
    $^{1}$Faculty of Mathematics and Natural Sciences,
    University of Oslo, PO Box 1047 Blindern, 0316 Oslo, Norway\\[0.3em]
    $^{\ast}$Corresponding author: \texttt{monoo@uio.no}%
}

\date{}   

\begin{document}
\maketitle

\begin{abstract}
\noindent
Accurate prediction of permeability tensors from pore-scale microstructure
images is essential for subsurface flow modeling, yet direct numerical
simulation requires hours per sample---fundamentally limiting large-scale
uncertainty quantification and reservoir optimization workflows.
A physics-informed deep learning framework is presented that resolves this
bottleneck by combining a MaxViT hybrid CNN-Transformer architecture with
progressive transfer learning and differentiable physical constraints.
MaxViT's multi-axis attention mechanism simultaneously resolves grain-scale
pore-throat geometry through block-local operations and REV-scale connectivity
statistics through grid-global operations, providing the spatial hierarchy
that permeability tensor prediction physically requires.
Training on 20{,}000 synthetic porous media samples spanning three orders of
magnitude in permeability, a three-phase progressive curriculum advances from
an ImageNet-pretrained baseline with D4-equivariant augmentation and  tensor transformation, through component-weighted loss prioritizing off-diagonal
coupling, to frozen-backbone transfer learning with porosity conditioning via
Feature-wise Linear Modulation (FiLM).
Onsager reciprocity and positive-definiteness are enforced through differentiable
penalty terms, achieving mean symmetry error $\varepsilon_{\mathrm{sym}}
=3.95\times10^{-7}$ and 100\,\% thermodynamic validity without post-hoc
correction.
On a held-out test set of 4{,}000 samples, the framework achieves
variance-weighted $R^2=0.9960$ ($R^2_{K_{xx}}=0.9967$,
$R^2_{K_{xy}}=0.9758$), a 33\,\% reduction in unexplained variance over the
supervised baseline, with every reported metric improving monotonically across
all three phases.
Inference requires 120\,ms per sample---a significant speedup over
lattice-Boltzmann simulation that enables real-time permeability characterization
during core scanning, Monte Carlo uncertainty quantification
on a single GPU, and rapid multi-scenario screening for CO$_2$ storage and
geo-energy applications.
The results establish three transferable principles for physics-informed
scientific machine learning: large-scale visual pretraining transfers
effectively across domain boundaries; physical constraints are most robustly
integrated as differentiable architectural components; and progressive
training guided by diagnostic failure-mode analysis enables unambiguous
attribution of performance gains across methodological stages.
\end{abstract}

\noindent \textbf{Keywords:} Permeability tensor prediction; Porous media; Physics-informed machine learning; Hybrid CNN-Transformer; 
Progressive transfer learning; Anisotropy; 
Uncertainty quantification; Subsurface flow characterization.

\vspace{0.6em}
\noindent\rule{\textwidth}{0.4pt}  
\vspace{0.2em}

\section{Introduction}
\label{sec:intro}

Fluid flow through porous geological formations governs a broad range of
processes of direct societal and engineering importance, including the
management of geo-energy resources, groundwater systems, geological
CO$_2$ and H$_2$ storage, and subsurface contaminant
transport~\cite{blunt2001flow,ladd2021reactive,okoroafor2022toward,ingham2012emerging,nisbet2024carbon,nooraiepour2025geological,mcdowell1986particle}.
Accurate characterization of porous media transport properties is a
prerequisite for reservoir management decisions, operational performance
and risk assessment for geological storage sites, aquifer remediation
strategies, and large-scale probabilistic uncertainty quantification
workflows that require millions of forward evaluations.
At the pore scale, fluid motion is governed by the Navier-Stokes
equations.
The permeability tensor $\mathbf{K}$ emerges as the fundamental
macroscopic transport property relating the pressure gradient to the
Darcy velocity through Darcy's law:
\begin{equation}
    \mathbf{q} = -\frac{\mathbf{K}}{\mu}\nabla p,
    \label{eq:darcy}
\end{equation}
where $\mathbf{q}$ is the Darcy velocity vector, $\mu$ is dynamic
viscosity, and $\nabla p$ is the pressure gradient.
For two-dimensional porous media, $\mathbf{K}$ is a $2\times2$
symmetric positive-definite matrix encoding directional flow preferences
and geometric tortuosity.
Computing $\mathbf{K}$ from pore-scale images through direct numerical
simulation (DNS) of the Stokes equations, employing methods such as
lattice-Boltzmann, finite-volume, or finite-element
schemes~\cite{meakin2009modeling,ehlers2019modelling,liu2016multiphase,chen2022pore},
yields highly accurate results but incurs computational costs of hours to
days per sample even on high-performance
hardware~\cite{moin1998direct,wood2020modeling}.
This expense severely constrains applications requiring thousands to
millions of permeability evaluations.

Recent advances in deep learning have demonstrated that neural networks
trained on sufficiently large labeled datasets derived from pore-scale
images can predict permeability tensors with exceptional accuracy while
reducing inference time from hours to milliseconds~\cite{da2021deep,delpisheh2024leveraging,abbasi2025challenges,peng2025recent}.
Early investigations established that convolutional neural networks
(CNNs) can learn effective representations of pore geometry directly
from image data, enabling rapid permeability
prediction~\cite{graczyk2020predicting,wu2018seeing,An2026,alqahtani2018deep}.
Subsequent contributions advanced these techniques through multi-scale
architectures and expanded applicability to diverse pore morphologies,
~\cite{yasuda2021machine,elmorsy2022generalizable,garttner2023estimating,li2025integrated}.
More recently, physics-informed neural networks have been investigated
to enforce adherence to underlying physical principles within the
model~\cite{tartakovsky2020physics,Kashefi202380,elmorsy2023rapid,li2025integrated,zhuang2025hybrid},
and graph neural networks have been applied to capture the topological
features and connectivity of pore
networks~\cite{zhao2025computationally,zhao2024rtg,alzahrani2023pore}.
These developments underscore the growing potential of integrating
domain knowledge into data-driven neural
architectures~\cite{cai2021physics,NooraiepourPDEreview,liu2022survey}.
The architectural foundation of the present work, a hybrid
CNN-Transformer design (MaxViT, detailed in
Section~\ref{subsec:architecture}), is motivated by these advances and their
demonstrated capacity for learning hierarchical spatial features from
binary porous media images while respecting the multi-scale nature of
permeability-geometry relationships.

While deep learning has emerged as the predominant approach for
image-based permeability prediction, the broader machine learning
landscape offers established regression paradigms that merit explicit
consideration as alternatives.
Gaussian process regression (GPR) provides principled Bayesian
uncertainty quantification through kernel-based modeling of function
smoothness~\cite{williams2006gaussian,Stulp201560,schulz2018tutorial}.
Ensemble methods such as random forests and gradient
boosting~\cite{breiman2001random,friedman2001greedy,Fernandez-Delgado201911,chen2016xgboost}
aggregate predictions from multiple weak learners for robustness against
noise.
Support vector regression employs kernel methods to handle nonlinear
relationships in moderate-dimensional feature
spaces~\cite{smola2004tutorial,drucker1996support,Huang201532}, and
regression trees offer transparent hierarchical
partitioning~\cite{hastie2009elements,Fernandez-Delgado201911}.
However, these methods face fundamental challenges at the scale of
high-resolution porous media images.
The input dimensionality of $128\times128=16{,}384$ pixels creates a
severe curse of dimensionality for methods operating on flattened
feature vectors: GPR scales cubically with the number of training
samples and requires explicit kernel design in this high-dimensional
space; support vector regression similarly demands careful kernel
selection in the same regime; and tree-based methods struggle to
capture smooth spatial structure without extreme depth.
More fundamentally, these approaches typically require hand-crafted
feature engineering to reduce dimensionality, extracting quantities
such as porosity, pore-size distributions, tortuosity, or correlation
functions, which demands domain expertise, introduces information
bottlenecks, and may discard spatial patterns not anticipated by the
feature designer.
Permeability depends nonlinearly on multi-scale spatial structure,
from grain-scale pore-throat geometry to domain-scale connectivity,
relationships that are difficult to capture through linear combinations
of scalar features or axis-aligned decision boundaries.
Deep convolutional architectures address these limitations through
hierarchical spatial feature learning~\cite{Bouwmans20198,Tian2020251,Buda2018249}:
convolutional layers with local receptive fields and weight sharing
encode translation equivariance and spatial smoothness priors
appropriate to image data; pooling operations implement multi-scale
aggregation without explicit feature engineering; and end-to-end
differentiable training enables direct optimization of the full
input-to-output mapping.
For the present task, predicting a $2\times2$ permeability tensor from
binary microstructure images spanning three orders of magnitude in
permeability, empirical studies demonstrate that deep CNNs consistently
achieve strong benchmark
performance~\cite{wu2018seeing,yasuda2021machine,garttner2023estimating},
while comparable results with non-deep methods would require extensive
manual feature extraction and domain-specific kernel design.

Despite these successes, existing deep learning approaches for
permeability prediction face several persistent limitations.
The first concerns architecture.
Methods based on pure CNNs excel at capturing local geometric features
and fine-grained interface details through translation equivariance and
local connectivity, but struggle with long-range spatial dependencies
critical for understanding global pore connectivity and anisotropic
coupling~\cite{mumuni2021cnn,lecun2015deep,wei2021fine,yao2016coarse}.
Conversely, pure Vision Transformers (ViTs) capture global context
through self-attention but incur quadratic computational complexity with
image size and lack the inductive biases that make CNNs effective for
structured spatial
data~\cite{dosovitskiy2020image,soydaner2022attention,vaswani2017attention,yang2021focal}.
This architectural tension is especially consequential for off-diagonal
permeability components $K_{xy}$ and $K_{yx}$, which encode geometric
anisotropy arising from pore channels oriented obliquely to the
principal coordinate axes.
Accurate prediction of these components requires simultaneously
resolving fine-scale pore-throat geometry at length scales of individual
grains (roughly 10--100 pixels), governing local flow resistance, and
long-range spatial correlations spanning the full image domain,
determining how preferential flow paths align and couple across
directions.
The characteristic spatial correlation length governing off-diagonal
coupling is typically comparable to the representative elementary volume
(REV) size, which for sandstone microstructures can span the entire
image domain.
Pure CNNs cannot integrate information over these scales without very
deep architectures that sacrifice fine-scale resolution, while pure ViTs
lack the inductive biases to resolve sharp solid-fluid interfaces
efficiently.

The second persistent limitation concerns physical validity.
Permeability tensors must satisfy fundamental physical requirements:
symmetry ($K_{xy}=K_{yx}$) arising from Onsager reciprocity, and
positive-definiteness ensuring thermodynamic consistency.
Most existing approaches enforce these constraints only weakly through
data characteristics or post-hoc projection, rather than incorporating
them directly into the training objective.
Furthermore, isotropic porous media exhibit natural geometric symmetries
under rotations and reflections.
Existing augmentation strategies either ignore these symmetries entirely
or apply image transformations without corresponding tensor
transformations, creating training inconsistencies that degrade model
performance and introduce physical violations.
It should be noted that enforcing symmetry through data augmentation,
as opposed to building equivariance into the network architecture, yields
approximate equivariance that is not guaranteed to hold exactly at
inference time; the implications of this distinction are discussed in
Section~\ref{sec:MM}.
The challenge of predicting off-diagonal tensor components remains
particularly acute, with reported accuracy often substantially lower
than for diagonal components, as these elements encode subtle geometric
anisotropies requiring global spatial integration.

Hybrid CNN-Transformer architectures have emerged as a promising
direction for simultaneously leveraging local and global feature
extraction~\cite{khan2023survey,xiao2021early,dai2021coatnet,liu2021swin,dong2022cswin,long2024hybrid}.
Among these, MaxViT~\cite{tu2022maxvit} introduces a multi-axis
self-attention mechanism that decomposes standard two-dimensional
self-attention into sequential block-local and grid-global operations.
This decomposition reduces computational complexity from
$\mathcal{O}(H^2W^2)$ to $\mathcal{O}(HW)$ while preserving a global
receptive field, a critical property for high-resolution scientific
imaging.
From a physical standpoint, block-local attention resolves grain-scale
pore-throat geometry at length scales relevant to local flow resistance,
while grid-global attention integrates information across the full image
domain for off-diagonal coupling.
This architectural decomposition directly mirrors the two-scale nature
of the permeability prediction problem: local geometry determines
individual flow path conductances, while global topology determines how
those conductances combine anisotropically into the full tensor.
MaxViT has achieved state-of-the-art results across diverse computer
vision benchmarks~\cite{tu2022maxvit,bangalore2024convision,hassija2025transformers}
and demonstrates particular strength in transfer learning scenarios with
limited labeled data, precisely the condition encountered in scientific
imaging applications where ground-truth labels require computationally
expensive physics-based simulation.
Despite these advantages, MaxViT has not been previously applied to
physics-informed prediction tasks or porous media characterization.

Prior to adopting supervised transfer learning, a systematic
investigation of self-supervised pretraining strategies for binary
porous media images was conducted, focusing on Masked Autoencoder
(MAE) approaches~\cite{he2022masked,feichtenhofer2022masked,reed2023scale}.
Contrastive methods such as SimCLR and MoCo were also considered but
their reliance on semantic augmentation invariances, including color
jitter and crop-based views, cannot be motivated for binary images
with no color information and where cropping may discard the globally
connected pore structure that constitutes the primary signal of
interest.
The MAE experiments revealed fundamental incompatibilities between MAE's
design assumptions and the characteristics of binary porous media.
MAE relies on smooth pixel gradients, high spatial redundancy, and
continuous-valued image statistics, properties abundant in natural
images but absent in binary porous media characterized by sharp
discontinuities at solid-fluid interfaces, low geometric redundancy,
and discrete pixel values.
Despite extensive architectural exploration spanning masking ratios
(25--75\,\%), decoder depths (1--8 layers), and reconstruction targets
(binary versus distance transform), all MAE configurations failed to
achieve usable reconstruction quality (best PSNR: 14.4\,dB versus a
required threshold of ${>}20$\,dB for meaningful feature learning,
established by evaluating fine-tuned regression head performance at
varying PSNR levels against a randomly initialized baseline).
This negative result provides an important insight: effective
representation learning for binary porous media requires supervision
from physically meaningful labels rather than pixel-level reconstruction
objectives.
Consequently, the framework developed here focuses on supervised
transfer learning from ImageNet-pretrained features, which provide
robust low- and mid-level visual representations that transfer
effectively despite the domain difference.
This negative result for self-supervised pretraining also directly
motivated the progressive transfer learning strategy detailed in
Section~\ref{subsec:trainingstrategy}, which leverages ImageNet-pretrained
features as a robust initialization and systematically refines them
through three phases of task-specific adaptation.

The framework proposed in this work addresses the limitations identified
above through a systematic combination of architectural selection,
physics-informed training objectives, rigorous geometric augmentation,
and progressive transfer learning.
The MaxViT hybrid architecture is applied for the first time to
permeability tensor prediction, with its multi-axis self-attention
mechanism providing the spatial hierarchy needed for simultaneous
fine-scale pore-throat resolution and global connectivity integration.
A physics-aware loss function is developed that explicitly enforces
tensor symmetry and positive-definiteness through differentiable penalty
terms, eliminating the need for post-hoc projection.
A D4-equivariant augmentation strategy is implemented that consistently
transforms both microstructure images and their associated permeability
tensors through proper tensor transformation matrices, removing the
training inconsistencies present in naive augmentation approaches.
While this augmentation-based approach achieves strong empirical
equivariance, it is acknowledged that it does not provide architectural
guarantees of exact equivariance at inference time, in contrast to group
equivariant
networks~\cite{cohen2016group,cohen2018spherical,weiler2019general};
this distinction is discussed in Section~\ref{sec:MM}.
A three-phase progressive training strategy is proposed: Phase~2
establishes a supervised baseline with ImageNet-pretrained MaxViT,
progressive unfreezing, and D4-equivariant augmentation; Phase~3
extends this through calibrated morphological and elastic deformations
with enhanced loss weighting that prioritizes off-diagonal tensor
elements; Phase~4 freezes the learned backbone while introducing
porosity-conditioned Feature-wise Linear Modulation
(FiLM)~\cite{perez2018film}, combined with ensemble techniques
including Stochastic Weight Averaging~\cite{izmailov2018averaging} and
Exponential Moving Average~\cite{polyak1992acceleration}.
The resulting framework attains an $R^2$ of 0.9960 on a diverse dataset
of 20{,}000 porous media samples spanning three orders of magnitude in
permeability, along with a significant reduction in
prediction error relative to the supervised baseline, while reducing
computational cost from hours per sample (DNS) to approximately 120\,ms
per prediction (inference).

\noindent The main contributions of this work are:
\begin{enumerate}[leftmargin=*, label=\arabic*.]
    \item {Physics-motivated hybrid architecture.}
    Systematic application of the MaxViT hybrid
    CNN-Transformer to permeability tensor prediction is presented,
    with an explicit physical justification connecting block-local
    attention to pore-throat geometry resolution and grid-global
    attention to the long-range connectivity integration governing
    off-diagonal tensor coupling.

    \item {Differentiable physics-informed loss.}
    A loss function is developed that enforces tensor symmetry and
    positive-definiteness through differentiable penalty terms,
    yielding near-machine-precision symmetry and 100\,\%
    thermodynamic validity without post-processing.

    \item {D4-equivariant augmentation with rigorous tensor
    transformation.}
    An augmentation strategy is introduced that consistently transforms
    microstructure images and their associated permeability tensors
    under the D4 symmetry group, eliminating training inconsistencies
    that arise when image and label transformations are mismatched.

    \item {Progressive three-phase transfer learning.}
    A structured training curriculum to transition from
    frozen ImageNet features to full fine-tuning and finally to
    porosity-conditioned FiLM adaptation, with each phase providing
    physically interpretable performance gains.

    \item {Computational efficiency.}
    The framework reduces permeability tensor inference time from hours
    (DNS) to approximately 120\,ms, enabling practical deployment in
    workflows requiring large-scale ensemble evaluations for
    uncertainty quantification and reservoir characterization.
\end{enumerate}

The remainder of this paper is organized as follows.
Section~\ref{sec:MM} describes the dataset generation and statistical
characteristics, the MaxViT hybrid CNN-Transformer architecture and its
adaptation for permeability tensor prediction, the physics-aware loss
formulation incorporating symmetry and positive-definiteness constraints,
the D4-equivariant data augmentation strategy with rigorous tensor
transformation, and the three-phase progressive training curriculum.
Section~\ref{sec:RD} presents experimental results across all training
phases, component-wise prediction accuracy analysis, physical constraint
validation, comparative evaluation against baseline methods, and
detailed discussion of failure modes and limitations.
Section~\ref{sec:conclusions} summarizes the key findings, contributions
to physics-informed machine learning for porous media characterization,
and implications for computational subsurface flow modeling.

\section{Materials and Methods}
\label{sec:MM}

\subsection{Dataset and Problem Formulation}
\label{subsec:dataset}

\subsubsection{Porous Media Image Dataset}
\label{subsubsec:images}

The dataset comprises 24{,}000 binary porous media images generated via
sequential Gaussian simulation, designed to replicate realistic sandstone
pore geometries by Vargdal et al.~\cite{vargdal2025neural,vargdal_2025_17711512}.
Each image is a strictly binary $128\times128$ pixel array in which
values of $1$ represent solid grain and values of $0$ represent pore
space, with porosity defined as $\phi = 1 - \mathrm{mean}(I)$.
Porosity spans $\phi\in[0.227,\,0.900]$ with mean
$\bar{\phi}=0.711\pm0.124$ across all splits; the distribution is
moderately left-skewed (skewness $\approx -0.55$) and significantly
non-normal (Shapiro-Wilk $p\approx 0$), reflecting the post-processing
described below.
Full descriptive statistics are provided in
Table~\ref{tab:dataset_stats} (Appendix~\ref{app:dataset}).

According to ~\cite{vargdal2025neural}, raw-generated images may contain isolated fluid clusters that are
completely enclosed by solid matrix and therefore disconnected from the
main flow network.
Because these regions contribute neither to flow nor to permeability,
an established post-processing step identified
and removed them through connectivity analysis under periodic boundary
conditions.
The resulting filled images represent the connected-pore-network
porosity, eliminated stagnant-region nodes from subsequent flow
simulations, and formed the basis for all permeability computations and
model training reported here~\cite{vargdal2025neural,vargdal_2025_17711512}.
This filling procedure, while physically motivated, may shift the effective
porosity distribution toward well-connected porous media,
placing the dataset mean substantially above the typical range for
natural sandstone reservoirs.
Predictions for tight or poorly connected samples therefore reside near
the lower tail of the training distribution and are expected to carry
higher relative uncertainty; the corresponding failure-mode analysis is
provided in Appendix~\ref{app:anisotropy}.

The full $2\times2$ permeability tensor for each image was computed via
lattice-Boltzmann (LBM) simulation of single-phase Stokes flow at low
Reynolds number~\cite{vargdal2025neural}.
Body forces applied in orthogonal directions and volume-averaged fluid
velocities yield the permeability tensor through Darcy's law, expressed
in lattice units of order unity.
Diagonal permeabilities span three orders of magnitude in
$\log_{10}$-space; off-diagonal components are centered near zero and
exhibit heavy tails (kurtosis $\approx 9$--13), reflecting the
predominantly isotropic character of the media.
All ground-truth tensors are 100\,\% positive-definite, and tensor
symmetry is preserved to $\varepsilon_{\mathrm{sym}}<8\times10^{-6}$ by
the LBM solver.
Complete LBM implementation and validation are documented
in~\cite{vargdal2025neural,vargdal_2025_17711512}.

Prior to model development, a comprehensive dataset characterization
study was conducted covering microstructure geometry, permeability
tensor statistics, porosity-permeability relationships, cross-split
distributional consistency, anisotropy analysis, and outlier
identification.
Kolmogorov-Smirnov tests and Jensen-Shannon divergence measurements
confirmed that training, validation, and test splits are drawn from
statistically indistinguishable distributions across all variables,
validating the random partitioning protocol.
These analyses are reported in full in Appendix~\ref{app:dataset}.

\subsubsection{Problem Formulation}
\label{subsubsec:formulation}

Permeability prediction is formulated as a supervised regression task,
\begin{equation}
    f_\theta:\mathbb{R}^{128\times128}\to\mathbb{R}^{2\times2},
    \label{eq:mapping}
\end{equation}
where $f_\theta$ denotes a neural network with parameters $\theta$ mapping
binary porous media images to symmetric positive-definite permeability
tensors:
\begin{equation}
    \mathbf{K}
    =\begin{bmatrix}K_{xx} & K_{xy}\\K_{xy} & K_{yy}\end{bmatrix},
    \quad K_{xx},\,K_{yy}>0.
    \label{eq:tensor_structure}
\end{equation}
The network produces a four-element output
$[K_{xx},\,K_{xy},\,K_{yx},\,K_{yy}]$, reshaped to a $2\times2$ matrix at
evaluation time.
Two physical constraints must be satisfied:
\begin{align}
    K_{xy} &= K_{yx}
        \quad(\text{tensor symmetry}),
        \label{eq:symmetry}\\
    K_{xx},\,K_{yy} &> 0
        \quad(\text{positive diagonal}).
        \label{eq:positivity}
\end{align}
These constraints are incorporated through the physics-aware loss function
(Section~\ref{subsec:lossfunction}) and verified post hoc on the held-out
test set. The positivity penalty in the loss function enforces positive diagonal elements in the predicted permeability tensor, while full positive-definiteness—although not enforced by construction—is consistently verified empirically across all samples in the held-out test set (Section~\ref{subsubsec:metrics}). The high input dimensionality ($128\times128=16{,}384$ pixels) and
strongly nonlinear, multi-scale structure-property relationships motivate
end-to-end differentiable architectures that learn spatial hierarchies
directly from raw images, in preference to methods requiring hand-crafted
dimensionality reduction prior to regression.

\subsubsection{Data Partitioning and Evaluation Protocol}
\label{subsubsec:partitioning}

Of the 24{,}000 total images, 20{,}000 carry computed permeability labels
and are partitioned by random sampling into a training set (16{,}000
samples, 80\,\%) and a validation set (4{,}000 samples, 20\,\%).
An independently generated set of 4{,}000 labeled samples from the same
distribution constitutes the held-out test set, never accessed during
model development, hyperparameter selection, or training decisions.
The remaining 4{,}000 unlabeled images, initially reserved for
self-supervised pretraining, were found to confer no benefit for binary
porous media (see Introduction) and are excluded from all reported
results.

Strict separation between all three splits is maintained throughout.
The validation set is used exclusively for model checkpointing and early
stopping based on validation $R^2$.
Augmentation parameters and loss-function weights were determined through
preliminary grid searches on the validation set, conducted independently
of the final training runs, to preclude implicit test-set leakage.
Reported test-set results therefore constitute unbiased estimates of
generalization to unseen porous media samples.

\subsection{Model Architecture}
\label{subsec:architecture}

\subsubsection{MaxViT Backbone}
\label{subsubsec:maxvit}

The backbone is a Multi-Axis Vision Transformer
(MaxViT-Base)~\cite{tu2022maxvit}, a hierarchical hybrid architecture
combining depthwise-separable convolutions for local feature extraction
with multi-axis self-attention for global context modeling.
The implementation follows the \texttt{timm} library~%
\cite{rw2019timm,wightman2021}, using weights pretrained on ImageNet-21K
and subsequently fine-tuned on ImageNet-1K, providing robust low- and
mid-level visual features that transfer effectively across the domain gap
to porous media images.

Two adaptations are applied to accommodate the target domain.
First, the input is reconfigured to single-channel grayscale
(\texttt{in\_chans=1}), modifying the first convolutional layer's input
channels while preserving pretrained spatial filters via channel averaging.
Second, the input resolution is specified as $128\times128$
(\texttt{img\_size=128}), triggering automatic interpolation of positional
embeddings and spatial-dimension adjustment across all hierarchical
stages.

The backbone comprises four stages with progressively increasing channel
dimensionality and decreasing spatial resolution.
Each stage contains multiple MaxViT blocks, each integrating an MBConv
module (depthwise-separable convolution with squeeze-and-excitation)
followed by two multi-axis self-attention modules: one applying blocked
local partitioning within non-overlapping $P\times P$ windows ($P=8$),
and one applying dilated grid-global attention across block positions.
This decomposition reduces self-attention complexity from
$\mathcal{O}(H^2W^2)$ to $\mathcal{O}(HW)$ while preserving a global
receptive field.
Stage dimensions, block counts, and attention configurations specific to
the $128\times128$ input are detailed in Appendix~\ref{app:architecture}.

After the final stage, spatial features are aggregated by global average
pooling to yield a 768-dimensional feature vector $\mathbf{f}$ encoding
global pore geometry:
\begin{equation}
    \mathbf{f}
    =\frac{1}{H_{\mathrm{final}}\cdot W_{\mathrm{final}}}
    \sum_{i=1}^{H_{\mathrm{final}}}\sum_{j=1}^{W_{\mathrm{final}}}
    \mathbf{x}_{ij}^{\mathrm{stage4}}.
    \label{eq:gap}
\end{equation}
A two-layer MLP regression head maps this representation to the
four-element permeability output (Phases~2 and~3):
\begin{align}
    \mathbf{h}_0
    &=\mathrm{LayerNorm}(\mathbf{f}),
    \label{eq:head_ln}\\
    \mathbf{h}_1
    &=\mathrm{Dropout}_{0.1}\!\left(
        \mathrm{GELU}\!\left(
            \mathrm{Linear}_{768\to256}(\mathbf{h}_0)
        \right)
    \right),\\
    \mathbf{h}_2
    &=\mathrm{Dropout}_{0.1}\!\left(
        \mathrm{GELU}\!\left(
            \mathrm{Linear}_{256\to128}(\mathbf{h}_1)
        \right)
    \right),\\
    \mathbf{K}_{\mathrm{pred}}
    &=\mathrm{Linear}_{128\to4}(\mathbf{h}_2).
    \label{eq:head_out}
\end{align}
The complete architecture comprises approximately 119.01 million parameters:
the MaxViT-Base backbone (convolutional stem and four hierarchical stages,
with the classification head removed) contributes 118.78M, and the
regression head contributes 0.23M.
The porosity encoder and FiLM conditioning layers introduced in Phase~4
(Section~\ref{subsubsec:phase4}) add a further 0.11M parameters.

\subsubsection{Porosity-Conditioned Feature Modulation}
\label{subsubsec:film}

Phase~4 introduces an explicit physics-informed conditioning mechanism in
which scalar porosity modulates backbone feature representations, encoding
the physical relationship between porosity and transport-coefficient
magnitude directly into the model.

A three-layer MLP encoder transforms the scalar porosity
$\phi\in[0,1]$ into a 64-dimensional embedding
$\mathbf{e}_\phi\in\mathbb{R}^{64}$:
\begin{equation}
    \mathbf{e}_\phi
    =\mathrm{Linear}_{64\to64}\!\left(
        \mathrm{ReLU}\!\left(
            \mathrm{Linear}_{32\to64}\!\left(
                \mathrm{ReLU}\!\left(
                    \mathrm{Linear}_{1\to32}(\phi)
                \right)
            \right)
        \right)
    \right).
    \label{eq:porosity_encoder}
\end{equation}
Porosity is computed directly from the binary image as
$\phi = 1 - \mathrm{mean}(I)$, requiring no additional simulation and
constituting a zero-cost auxiliary input.

Feature-wise Linear Modulation (FiLM)~\cite{perez2018film} layers then
modulate backbone feature maps
$\mathbf{x}\in\mathbb{R}^{H\times W\times C}$ at Stages~2, 3, and~4:
\begin{align}
    \gamma_\phi
    &=\mathrm{Linear}_{64\to C}(\mathbf{e}_\phi)\in\mathbb{R}^C,
    \quad
    \beta_\phi
    =\mathrm{Linear}_{64\to C}(\mathbf{e}_\phi)\in\mathbb{R}^C,
    \label{eq:film_params}\\
    \mathbf{x}_{\mathrm{mod}}
    &=\gamma_\phi\odot\mathbf{x}+\beta_\phi,
    \label{eq:film_modulation}
\end{align}
where $\odot$ denotes elementwise multiplication with broadcasting over
spatial dimensions.
Scale and shift parameters are initialized to produce identity modulation
($\gamma=\mathbf{1}$, $\beta=\mathbf{0}$) at training onset, ensuring
stability.

Stage~1 is excluded because its features encode low-level edge detection
and solid-fluid boundary localization, representing a length scale at
which porosity provides no physically meaningful conditioning context.
Stages~2--4 encode progressively more abstract geometric descriptions,
including connectivity patterns, channel topology, and global
pore-network structure, where porosity is a physically meaningful
modulator of transport-coefficient magnitude.
This placement was confirmed by a five-configuration validation-set
ablation, and the encoder dimensionality of 64 was selected from
$\{32,64,128\}$ by the same procedure; full ablation results and
rationale are reported in Appendix~\ref{app:film_ablation}.

The total trainable parameter count in Phase~4 comprises the porosity
encoder (0.01M), three FiLM layer pairs (0.10M), and the regression head
(0.23M), totaling 0.34M parameters, representing 0.3\,\% of the full model.

\subsection{Progressive Training Strategy}
\label{subsec:trainingstrategy}

\begin{figure}[t]
    \centering
    \includegraphics[width=\textwidth]{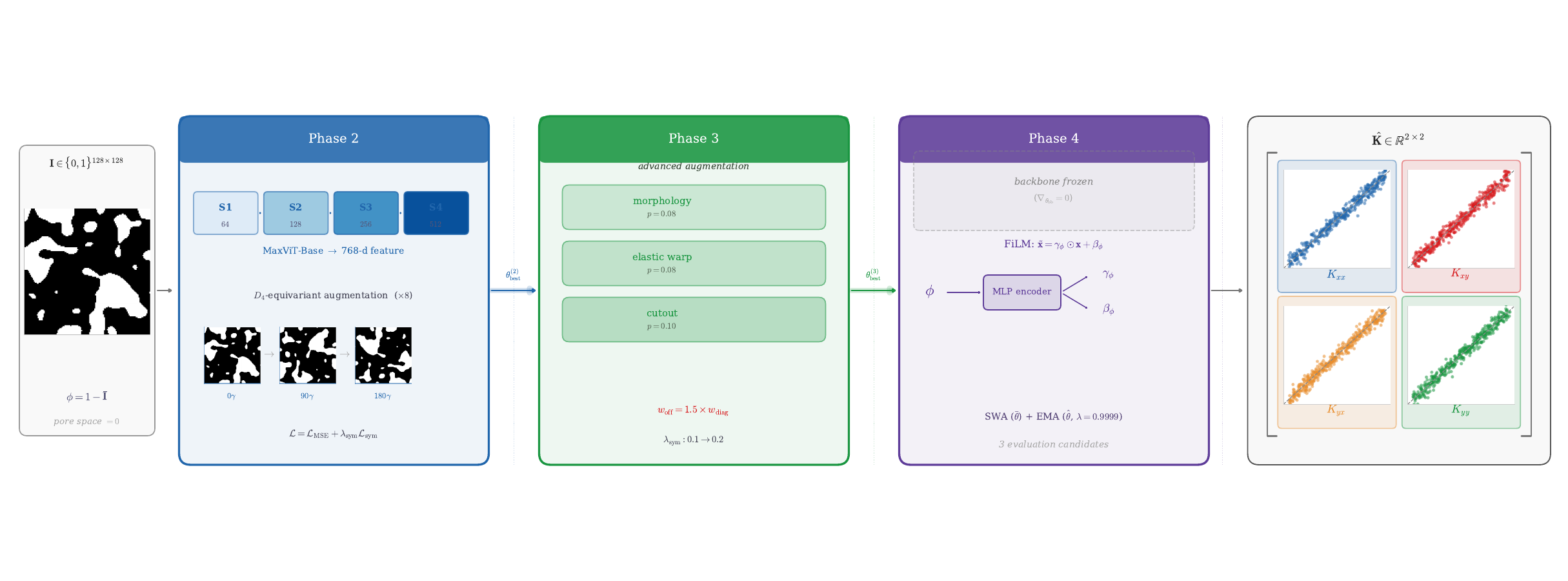}
    \caption{%
        Overview of the physics-informed progressive transfer learning
        pipeline for permeability tensor prediction.
        \emph{Input (far left):} a binary porous media image
        $\mathbf{I}\in\{0,1\}^{128\times128}$ with porosity
        $\phi = 1 - \overline{\mathbf{I}}$, where pixel values of
        1 denote solid grains and 0 denote pore space.
        \emph{Phase~2 (blue):} the ImageNet-pretrained MaxViT-Base
        backbone processes the image through four hierarchical stages
        (S1--S4, channels 64--512) to produce a 768-dimensional
        feature vector; D4-equivariant augmentation applies all eight
        dihedral symmetries (three rotations shown) with consistent
        tensor transformation, and the physics-aware loss enforces
        tensor symmetry ($\lambda_{\mathrm{sym}}=0.1$).
        Phases are connected by checkpoint transfer arrows labeled
        $\theta_{\mathrm{best}}^{(k)}$.
        \emph{Phase~3 (green):} three supplementary augmentation
        classes---morphological operations ($p=0.08$), elastic
        deformation ($p=0.08$), and cutout masking ($p=0.10$)---are
        introduced alongside doubled symmetry weighting
        ($\lambda_{\mathrm{sym}}=0.2$) and off-diagonal loss
        prioritization ($w_{\mathrm{off}}=1.5\,w_{\mathrm{diag}}$).
        \emph{Phase~4 (purple):} the backbone is frozen
        ($\nabla_{\theta_{\mathrm{bb}}}=0$, indicated by hatching)
        while a porosity MLP encoder drives FiLM conditioning
        ($\tilde{\mathbf{x}} = \gamma_\phi \odot \mathbf{x} +
        \beta_\phi$) at Stages~2--4; Stochastic Weight Averaging
        (SWA) and Exponential Moving Average
        (EMA) produce three evaluation candidates
        from a single training run.
        \emph{Output (far right):} the predicted permeability tensor
        $\hat{\mathbf{K}}\in\mathbb{R}^{2\times2}$, with
        representative parity plots shown for all four components
        ($K_{xx}$, $K_{xy}$, $K_{yx}$, $K_{yy}$).%
    }
    \label{fig:pipeline}
\end{figure}

Training is organized into four numbered phases.
Phase~1 constitutes dataset curation, LBM simulation, and the
comprehensive characterization study described in
Section~\ref{subsec:dataset} and Appendix~\ref{app:dataset}; it
produces no trained model and is not discussed further here.
Phases~2--4 form the progressive learning curriculum: each initializes
from the best checkpoint of the preceding phase and introduces a single,
physically motivated methodological contribution—supervised transfer
learning with D4-equivariant augmentation (Phase~2), advanced
augmentation with enhanced off-diagonal loss weighting (Phase~3), and
frozen-backbone porosity conditioning with weight-averaged ensemble
(Phase~4)—enabling unambiguous attribution of performance gains to each
intervention.
Full hyperparameter specifications for all phases are consolidated in
Appendix~\ref{app:training_config}. The complete pipeline is illustrated in Figure~\ref{fig:pipeline}.

\subsubsection{Phase~2: Supervised Baseline with Progressive Unfreezing}
\label{subsubsec:phase2}

Phase~2 establishes the supervised baseline using the ImageNet-pretrained
MaxViT backbone combined with D4-equivariant augmentation
(Section~\ref{subsubsec:d4aug}).
To prevent catastrophic forgetting of pretrained representations, the
trainable parameter set is expanded progressively over 600 epochs:
only the regression head is optimized during epochs~0--50; backbone
Stages~3--4 are unfrozen during epochs~50--150; and full end-to-end
fine-tuning proceeds from epoch~150 onward.
This curriculum allows the randomly initialized regression head to adapt
to the permeability prediction task before backbone weights are updated,
reducing the risk of destabilizing pretrained features.
The physics-aware loss (Section~\ref{subsec:lossfunction}) is applied
with $\lambda_{\mathrm{sym}}=0.1$ and $\lambda_{\mathrm{pos}}=0.05$;
the off-diagonal prioritization term is not active in this phase.

\subsubsection{Phase~3: Advanced Augmentation and Optimized Loss}
\label{subsubsec:phase3}

Phase~3 initializes from the Phase~2 best checkpoint and trains for an
additional 600 epochs.
Three supplementary augmentation classes are introduced at carefully
tuned intensities (Section~\ref{subsubsec:advancedaug}).
The loss function is enhanced with a doubled symmetry weight
($\lambda_{\mathrm{sym}}=0.2$) and the off-diagonal prioritization term
$\mathcal{L}_{\mathrm{offdiag}}$ (Section~\ref{subsec:lossfunction}),
which applies a $1.5\times$ weight to $K_{xy}$ and $K_{yx}$ predictions,
addressing the performance disparity between diagonal and off-diagonal
components observed in Phase~2.
The increased symmetry weight couples the learning dynamics of $K_{xy}$
and $K_{yx}$, allowing gradients computed on either element to inform the
other.

\subsubsection{Phase~4: Physics-Informed Conditioning with Weight
Averaging}
\label{subsubsec:phase4}

Phase~4 initializes from the Phase~3 best checkpoint with the MaxViT
backbone frozen throughout.
Only the newly introduced porosity encoder, FiLM layers, and regression
head are optimized (0.34M parameters; 0.3\,\% of total).
The frozen-backbone strategy offers three key properties:
\emph{stability}, eliminating the risk of corrupting learned pore-scale
geometry representations through further gradient updates;
\emph{efficiency}, as backpropagation bypasses frozen layers,
substantially reducing memory consumption and training time; and
\emph{interpretability}, ensuring that any performance improvement is
attributable solely to the porosity conditioning mechanism.

Two complementary weight-averaging strategies improve generalization by
exploiting flatter, wider regions of the loss landscape:

\textit{Stochastic Weight Averaging (SWA)~\cite{izmailov2018averaging}.}
From epoch~400 onward, a running arithmetic mean of model weights is
maintained at a fixed learning rate of $5\times10^{-6}$, aggregating 200
snapshots without additional training cost.
After training, batch-normalization statistics are recalibrated via a
single pass through the training set to restore accuracy.

\textit{Exponential Moving Average
(EMA)~\cite{morales2024exponential}.}
A continuously updated exponentially-weighted parameter average is
maintained throughout training with decay $\lambda_{\mathrm{EMA}}=0.9999$,
corresponding to an effective window of approximately 10{,}000 gradient
steps.
EMA filters high-frequency weight fluctuations while tracking the
optimization trajectory, consistently outperforming instantaneous weights
on validation data.

Phase~4 training produces three evaluation candidates sharing the same
architecture: the best instantaneous checkpoint $\theta_{\mathrm{best}}$,
the SWA-averaged model $\theta_{\mathrm{SWA}}$, and the EMA-averaged model
$\theta_{\mathrm{EMA}}$.
All three are evaluated independently with test-time D4 augmentation to
avoid selection bias.
Full mathematical formulations, update rules, and batch-normalization
recalibration details appear in Appendix~\ref{app:ensemble}.

\subsection{Physics-Aware Loss Function}
\label{subsec:lossfunction}

The training objective combines four terms that jointly enforce predictive
accuracy and physical validity:
\begin{equation}
    \mathcal{L}_{\mathrm{perm}}
    =\mathcal{L}_{\mathrm{MSE}}
    +\lambda_{\mathrm{sym}}\,\mathcal{L}_{\mathrm{sym}}
    +\lambda_{\mathrm{pos}}\,\mathcal{L}_{\mathrm{pos}}
    +\lambda_{\mathrm{offdiag}}\,\mathcal{L}_{\mathrm{offdiag}}.
    \label{eq:loss_total}
\end{equation}
Phase-specific weight values are reported in Appendix~\ref{app:training_config}.

\textbf{Reconstruction loss.}
The primary term penalizes tensor-level prediction error via the Frobenius
norm:
\begin{equation}
    \mathcal{L}_{\mathrm{MSE}}
    =\frac{1}{N}\sum_{i=1}^{N}
    \bigl\|\mathbf{K}_i^{\mathrm{pred}}-\mathbf{K}_i^{\mathrm{true}}%
    \bigr\|_F^2.
    \label{eq:loss_mse}
\end{equation}

\textbf{Symmetry constraint.}
Tensor symmetry ($K_{xy}=K_{yx}$) is enforced through a soft quadratic
penalty on the off-diagonal discrepancy:
\begin{equation}
    \mathcal{L}_{\mathrm{sym}}
    =\frac{1}{N}\sum_{i=1}^{N}
    \bigl(K_{xy,i}^{\mathrm{pred}}-K_{yx,i}^{\mathrm{pred}}\bigr)^2.
    \label{eq:loss_sym}
\end{equation}
Although ground-truth labels satisfy symmetry to near-machine precision
(Table~\ref{tab:dataset_stats}), unconstrained regression does not
inherently learn this property, motivating explicit soft enforcement.
The weight $\lambda_{\mathrm{sym}}$ is increased from $0.1$ in Phase~2 to
$0.2$ in Phases~3--4, coupling the learning dynamics of $K_{xy}$ and
$K_{yx}$ and improving overall off-diagonal accuracy.

\textbf{Positivity constraint.}
A squared hinge loss with margin $\varepsilon=0.001$ enforces strictly
positive diagonal elements:
\begin{equation}
    \mathcal{L}_{\mathrm{pos}}
    =\frac{1}{N}\sum_{i=1}^{N}
    \Bigl[
        \max\!\bigl(0,\,\varepsilon-K_{xx,i}^{\mathrm{pred}}\bigr)^2
       +\max\!\bigl(0,\,\varepsilon-K_{yy,i}^{\mathrm{pred}}\bigr)^2
    \Bigr].
    \label{eq:loss_pos}
\end{equation}
The penalty activates only for sub-margin predictions, providing directed
gradient signal without penalizing physically valid outputs.
The non-zero margin ensures predicted permeabilities exceed zero by a
meaningful amount, improving numerical stability in downstream flow
simulations.

\textbf{Off-diagonal prioritization.}
A component-weighted mean squared error amplifies gradients for the more
challenging off-diagonal elements:
\begin{equation}
    \mathcal{L}_{\mathrm{offdiag}}
    =\frac{1}{N}\sum_{i=1}^{N}
    \Bigl[
        w_{\mathrm{d}}\bigl(K_{xx,i}^{\mathrm{err}}\bigr)^2
       +w_{\mathrm{o}}\bigl(K_{xy,i}^{\mathrm{err}}\bigr)^2
       +w_{\mathrm{o}}\bigl(K_{yx,i}^{\mathrm{err}}\bigr)^2
       +w_{\mathrm{d}}\bigl(K_{yy,i}^{\mathrm{err}}\bigr)^2
    \Bigr],
    \label{eq:loss_offdiag}
\end{equation}
where $K_{jk,i}^{\mathrm{err}}=K_{jk,i}^{\mathrm{pred}}-K_{jk,i}^{\mathrm{true}}$,
$w_{\mathrm{d}}=1.0$, and $w_{\mathrm{o}}=1.5$.
This targeted emphasis addresses a fundamental prediction difficulty:
diagonal permeabilities $K_{xx}$ and $K_{yy}$ correlate strongly with
local pore geometry directly accessible through convolutional features,
whereas off-diagonal elements $K_{xy}$ and $K_{yx}$ encode subtle
geometric anisotropies requiring global spatial integration, precisely the
regime in which the multi-axis attention mechanism provides its principal
advantage.
The 50\,\% weight increase provides a stronger learning signal for
off-diagonal components without overwhelming the overall loss landscape.

\subsection{Data Augmentation Strategy}
\label{subsec:dataaugmentation}

\subsubsection{Phase~2: D4-Equivariant Augmentation}
\label{subsubsec:d4aug}

For isotropic porous media, the permeability tensor is invariant under
the dihedral group $D_4$, comprising eight elements: the identity,
rotations by $90^\circ$, $180^\circ$, and $270^\circ$, and reflections
about the horizontal, vertical, diagonal, and anti-diagonal
axes~\cite{bear2013dynamics,whitaker2013method}.
Each training sample undergoes a uniformly random $D_4$ transformation
at every epoch, multiplying the effective dataset size by a factor of
eight at no additional labeling cost.

Physically consistent augmentation requires simultaneous transformation
of both the image and the permeability tensor.
Under a coordinate transformation represented by matrix $\mathbf{P}$,
the second-order permeability tensor transforms as:
\begin{equation}
    \mathbf{K}'=\mathbf{P}\,\mathbf{K}\,\mathbf{P}^T.
    \label{eq:tensor_transform}
\end{equation}
For example, the $90^\circ$ rotation matrix
$\mathbf{P}_{90}=\bigl[\begin{smallmatrix}0&-1\\1&\phantom{-}0
\end{smallmatrix}\bigr]$ yields:
\begin{equation}
    \mathbf{K}'
    =\begin{bmatrix}K_{yy}&-K_{yx}\\-K_{xy}&K_{xx}\end{bmatrix}.
    \label{eq:rot90_transform}
\end{equation}
Transformation matrices for all eight $D_4$ elements are tabulated in
Appendix~\ref{app:augmentation}.
Because these symmetries are exact for isotropic media, $D_4$
augmentation constitutes physically rigorous regularization rather than
an approximation.
The equivariance achieved is statistical rather than architectural;
group-equivariant convolutional networks~\cite{cohen2016group,%
weiler2019general}, which build exact equivariance into the
weight-sharing structure, were not adopted because doing so would require
training from scratch and discarding the ImageNet pretraining central to
the transfer-learning strategy employed here.
A detailed discussion of this design choice appears in
Appendix~\ref{app:equivariance_discussion}.

\subsubsection{Phase~3: Advanced Augmentation}
\label{subsubsec:advancedaug}

Three additional transformation classes are applied sequentially after
$D_4$ augmentation, at intensities determined by a two-stage
validation-set grid search (Appendix~\ref{app:augmentation}):

\textit{Morphological operations.}
Erosion and dilation with a $3\times3$ structuring element, each applied
with probability $0.08$, simulate natural variability in pore-throat
geometry.
No explicit tensor transformation is required, as the network learns the
permeability of the morphologically modified geometry through
convolutional feature extraction.

\textit{Elastic deformation.}
Smooth coordinate warping is generated from Gaussian-smoothed uniform
random displacement fields, with displacement magnitude $\alpha=3.0$
pixels, smoothing $\sigma=2.0$ pixels, and application probability
$0.08$.
Elastic deformation requires an explicit permeability tensor
transformation according to the local deformation Jacobian $\mathbf{J}$:
\begin{equation}
    \mathbf{K}'=\mathbf{J}\,\mathbf{K}\,\mathbf{J}^T,
    \quad
    \mathbf{J}=\begin{bmatrix}
        1+\partial_x\delta_x & \partial_y\delta_x\\
        \partial_x\delta_y   & 1+\partial_y\delta_y
    \end{bmatrix},
    \label{eq:elastic_transform}
\end{equation}
approximated at the image center via central finite differences, followed
by symmetrization $\mathbf{K}'\leftarrow(\mathbf{K}'+\mathbf{K}'^T)/2$.

\textit{Cutout masking.}
Random $8\times8$ pixel rectangular regions are masked to a neutral value
of $0.5$ with probability $0.10$, forcing distributed feature learning by
preventing reliance on any single spatial region.
No tensor transformation is required.

Augmentation parameters were optimized by targeting off-diagonal $R^2$
as the primary criterion, as diagonal components proved robustly learned
across the full search range while off-diagonal prediction was the most
sensitive indicator of augmentation quality.
Total combined perturbation probability is held below approximately
35\,\% to avoid gradient variance that would impede off-diagonal
convergence.
The full parameter search procedure and pipeline pseudocode appear in
Appendix~\ref{app:augmentation}.

\subsection{Implementation and Evaluation}
\label{subsec:implementation}

\subsubsection{Implementation Details}
\label{subsubsec:impl_details}

All models were implemented in PyTorch~2.6.0 using the \texttt{timm}~%
1.0.22 library and trained on a single NVIDIA RTX~6000 Ada Generation
GPU (48\,GB VRAM) under CUDA~12.4.
Automatic mixed precision was disabled across all phases to ensure
numerical stability in physics-constraint calculations and maintain
gradient precision for the small trainable parameter set of Phase~4.
Random seeds were fixed at 42 for NumPy, PyTorch, and Python's
\texttt{random} module to ensure reproducibility.
Deterministic CuDNN algorithms were not enforced to allow benchmark-mode
acceleration, providing approximately 15\,\% training speedup without
observable impact on cross-run reproducibility.

Training follows a sequential protocol across phases.
Phase~2 trains from ImageNet-pretrained weights for 600 epochs.
Phase~3 initializes from the Phase~2 best checkpoint for a further 600
epochs.
Phase~4 initializes from the Phase~3 best checkpoint with the backbone
frozen, training only the porosity encoder, FiLM layers, and regression
head for 600 epochs.
Phase~4 model variants ($\theta_{\mathrm{best}}$, $\theta_{\mathrm{SWA}}$,
$\theta_{\mathrm{EMA}}$) share identical architecture and forward-pass
computation graphs; weight-averaging affects only the parameter values
loaded at inference and introduces no additional computational cost.
All Phase~4 variants achieve an inference latency of approximately
120\,ms per sample on the NVIDIA RTX~6000 Ada (mean over 1{,}000
independent predictions), compared to hours per sample for LBM
simulation.
Complete optimizer configurations, learning-rate schedules, batch sizes,
and early-stopping criteria for each phase are reported in
Appendix~\ref{app:training_config}.

\subsubsection{Evaluation Metrics}
\label{subsubsec:metrics}

Model performance is assessed through four complementary metric classes.

\textit{Coefficient of determination ($R^2$).}
The primary accuracy metric measures the proportion of variance explained:
\begin{equation}
    R^2
    =1-\frac{\sum_{i=1}^N(y_i^{\mathrm{pred}}-y_i^{\mathrm{true}})^2}
            {\sum_{i=1}^N(y_i^{\mathrm{true}}-\bar{y})^2},
    \label{eq:r2}
\end{equation}
reported as an overall variance-weighted score across all four tensor
components and as per-component values for
$[K_{xx},\,K_{xy},\,K_{yx},\,K_{yy}]$.

\textit{Mean squared error (MSE).}
Absolute tensor-level prediction error is quantified via the Frobenius
norm (Equation~\ref{eq:loss_mse}).

\textit{Relative root mean squared error (RRMSE).}
To enable direct comparison between the large-magnitude diagonal and
near-zero-mean off-diagonal components, scale-invariant relative error
is reported per tensor component:
\begin{equation}
    \mathrm{RRMSE}_{jk}
    =\frac{\sqrt{\tfrac{1}{N}\sum_i
        (K_{jk,i}^{\mathrm{pred}}-K_{jk,i}^{\mathrm{true}})^2}}
         {\bar{K}_{jk}^{\mathrm{true}}},
    \label{eq:rrmse}
\end{equation}
normalized by the mean absolute test-set value $\bar{K}_{jk}^{\mathrm{true}}$.
A value of, say, 5\,\% on $K_{xx}$ corresponds directly to a 5\,\%
uncertainty in the principal flow resistance.

\textit{Physical validity metrics.}
Symmetry is quantified by the mean symmetry error
$\varepsilon_{\mathrm{sym}}=\mathbb{E}[|K_{xy}^{\mathrm{pred}}
-K_{yx}^{\mathrm{pred}}|]$ and the fraction of test predictions
satisfying $|K_{xy}-K_{yx}|<10^{-3}$.
Positivity is assessed by the fraction of predictions satisfying
$K_{xx},\,K_{yy}>0$.
The reported $\varepsilon_{\mathrm{sym}}=3.95\times10^{-7}$ represents
engineering-level precision far exceeding practical requirements, though
not machine precision in the strict floating-point sense
($\sim10^{-16}$ for FP64); the term ``near-machine precision'' is
therefore used in this context.

Extended metrics for cross-study comparison, including mean absolute
percentage error (MAPE), range-normalized RMSE, Willmott's index of
agreement, the Kling-Gupta efficiency, and Spearman rank correlation,
are defined in Appendix~\ref{app:metrics}.

\section{Results and Discussion}
\label{sec:RD}

The progressive training framework is evaluated on a held-out test set
of 4{,}000 porous media samples spanning three orders of magnitude in
permeability ($K\in[0.002,\,3.19]$ in lattice units), never accessed
during model development, hyperparameter selection, or any training
decision.
Three training phases are assessed in sequence: Phase~2 establishes the
supervised baseline with ImageNet-pretrained MaxViT and D4-equivariant
augmentation; Phase~3 introduces advanced augmentation and the enhanced
physics-aware loss; and Phase~4 applies frozen-backbone transfer
learning with porosity-conditioned FiLM layers and ensemble weight
averaging.
Each phase is evaluated independently, enabling attribution of
performance gains to specific methodological contributions.
The evaluation protocol, metrics, and physical validity criteria are
defined in Section~\ref{subsubsec:metrics}; comprehensive dataset
statistics and cross-split consistency verification are reported in
Appendix~\ref{app:dataset}.

\subsection{Dataset Overview and Structure-Property Relationships}
\label{subsec:dataset_overview}

\begin{figure}[!ht]
\centering
\includegraphics[width=0.95\textwidth]{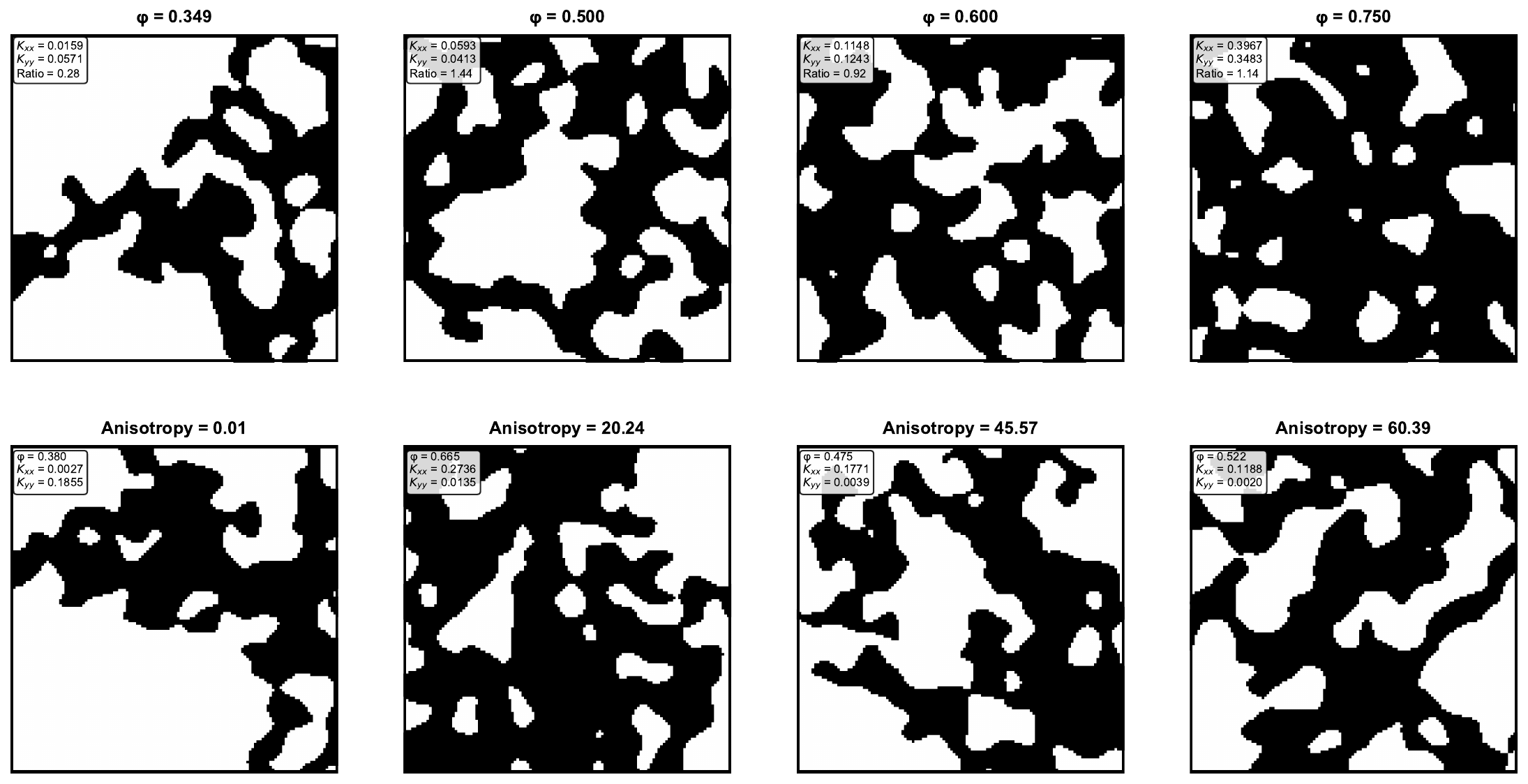}
\caption{Representative binary porous media microstructures spanning the
porosity and anisotropy spectrum.
Each panel shows a $128\times128$ binary image (white: solid matrix;
black: void space) with corresponding porosity $\phi$ and permeability
tensor components.
The progression from low to high porosity illustrates the transition
from tortuous isolated pore channels (small diagonal permeability,
negligible off-diagonal coupling) to well-connected isotropic networks
(large nearly equal $K_{xx}$, $K_{yy}$).
Samples with significant off-diagonal components ($|K_{xy}|>0.05$) are
associated with oblique pore-channel orientations rather than
porosity magnitude, consistent with the near-zero
porosity-to-off-diagonal correlation ($\rho_s\approx0.01$) reported
in Appendix~\ref{app:phi_perm}.}
\label{fig:dataset_microstructures}
\end{figure}

Figure~\ref{fig:dataset_microstructures} presents representative binary
microstructures spanning the porosity and anisotropy spectrum, illustrating
the geometric diversity of the dataset.
Low-porosity samples ($\phi<0.40$) exhibit dense solid matrices with
isolated, tortuous pore channels and correspondingly small permeabilities;
moderate-porosity samples ($0.40\leq\phi\leq0.70$) display increasingly
interconnected pore networks and non-negligible off-diagonal coupling;
and high-porosity samples ($\phi>0.70$) feature well-connected, nearly
isotropic networks with large diagonal permeabilities.

\begin{figure}[!ht]
\centering
\includegraphics[width=0.95\textwidth]{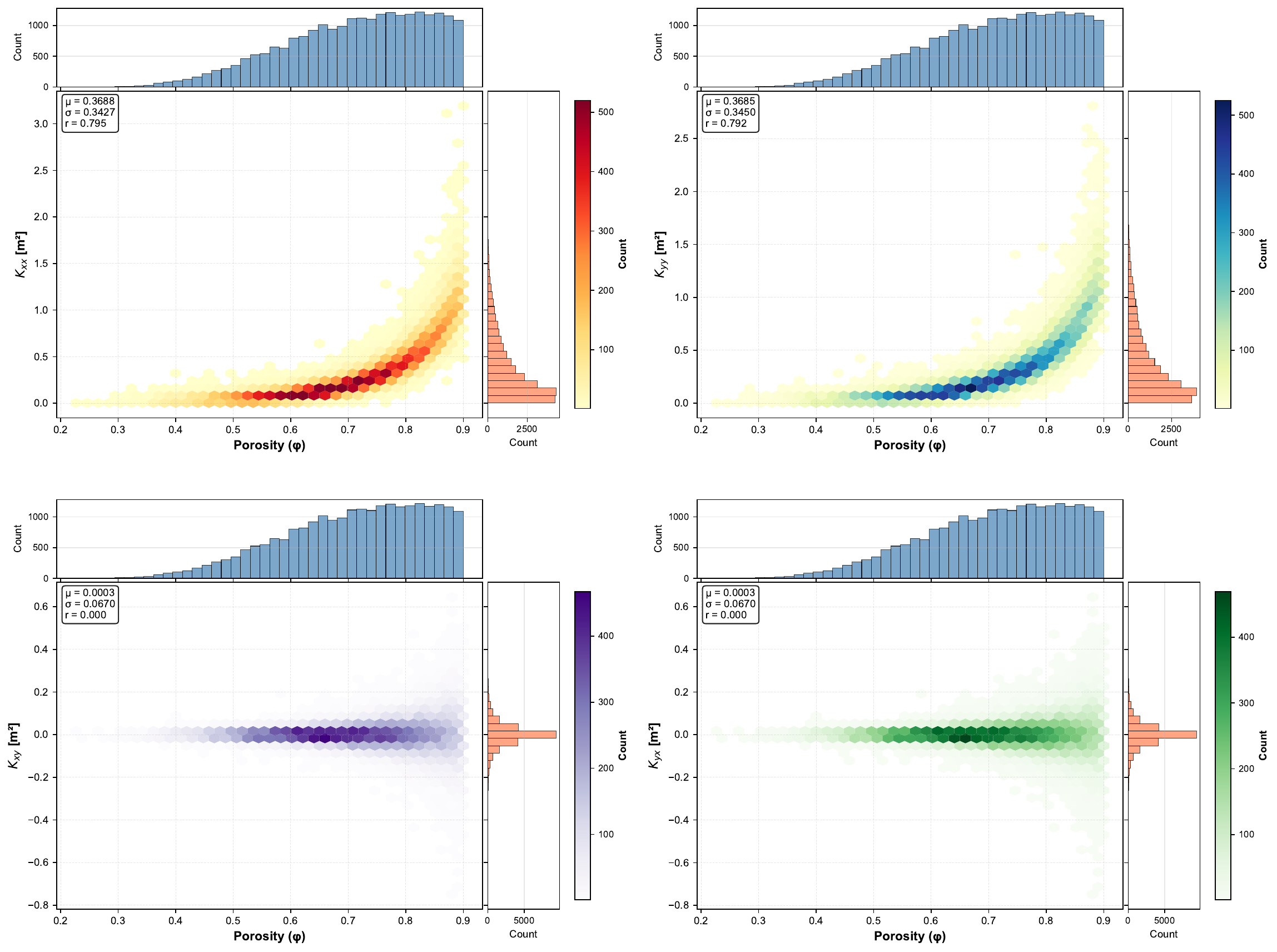}
\caption{Component-wise porosity-permeability relationships across all
20{,}000 labeled samples, shown as hexbin density plots with marginal
histograms.
Color intensity indicates sample density.
(a)~$K_{xx}$ versus $\phi$: strong positive correlation
($\rho_s=+0.940$, $r=+0.795$), spanning three orders of magnitude;
scatter increases at high porosity, reflecting geometric diversity
beyond scalar pore fraction.
(b)~$K_{yy}$ versus $\phi$: nearly identical statistics and correlation
structure to $K_{xx}$, confirming statistical isotropy at the dataset
level despite individual-sample anisotropy.
(c)~$K_{xy}$ versus $\phi$: negligible correlation
($\rho_s\approx+0.01$, $p>0.05$); off-diagonal coupling clusters
symmetrically near zero, independent of pore volume fraction.
(d)~$K_{yx}$ versus $\phi$: identical to $K_{xy}$, confirming
near-machine-precision ground-truth symmetry
($\varepsilon_{\mathrm{sym}}<8\times10^{-6}$, Table~\ref{tab:dataset_stats}).
The contrast between panels (a,b) and (c,d) constitutes the fundamental
prediction-difficulty asymmetry between diagonal and off-diagonal
tensor components that the progressive training strategy is designed to
address.}
\label{fig:dataset_statistics}
\end{figure}

Figure~\ref{fig:dataset_statistics} presents component-wise
porosity-permeability relationships for all 20{,}000 labeled samples,
revealing the fundamental structure-property correlations that govern
model learning.
Diagonal permeabilities $K_{xx}$ and $K_{yy}$ exhibit strong positive
correlation with porosity (Spearman $\rho_s\approx+0.940$,
Pearson $r\approx+0.795$, $p\approx0$), confirming that pore volume
fraction governs primary transport capacity and motivating the
porosity-conditioned FiLM design in Phase~4
(Section~\ref{subsubsec:film}).
The joint distributions reveal right-skewed marginal porosity (skewness
$\approx-0.55$, concentrated in the moderate-to-high connectivity
regime) and heavy-tailed component distributions spanning three orders
of magnitude.
Off-diagonal elements $K_{xy}$ and $K_{yx}$ exhibit negligible
correlation with porosity ($\rho_s\approx+0.01$, not significant),
confirming that cross-directional coupling arises solely from geometric
anisotropy and is independent of pore volume fraction.
This dissociation is a key dataset characteristic: diagonal and
off-diagonal permeabilities encode physically distinct information, the
former captured by scalar porosity and the latter requiring spatial
integration of geometric anisotropy features, which directly motivates
the component-specific loss weighting introduced in Phase~3
(Section~\ref{subsec:lossfunction}).
The symmetric clustering of off-diagonal components about zero (mean
$|\bar{K}_{xy}|<0.002$, kurtosis $\approx9$--13) confirms ground-truth
tensor symmetry, and all 20{,}000 labeled tensors are strictly
positive-definite, providing thermodynamically consistent training
labels free of simulation artifacts.

The dataset presents three distinct challenges for permeability
prediction.
First, the three-order-of-magnitude dynamic range in diagonal
permeability demands model capacity to resolve both low-connectivity
sparse networks and high-connectivity continuum-limit behavior across
the full spectrum simultaneously.
Second, the off-diagonal components, independent of porosity and encoded
exclusively in spatial geometric anisotropy, require global feature
integration at scales comparable to the representative elementary volume
(REV), a regime where convolutional features alone are insufficient.
Third, the five-fold scale disparity between diagonal and off-diagonal
standard deviations ($\approx5.1\times$ across all splits;
Appendix~\ref{app:desc_stats}) creates an implicit loss-weighting
imbalance in naive MSE training, motivating explicit off-diagonal
prioritization.
These challenges are addressed directly by the architectural and
training design decisions described in Section~\ref{sec:MM}, and
the following subsections assess the extent to which each phase
of training resolves them.


\subsection{Phase~2: Supervised Baseline with D4-Equivariant Augmentation}
\label{subsec:phase2_results}

Phase~2 establishes the supervised baseline by fine-tuning
ImageNet-pretrained MaxViT with progressive unfreezing and
D4-equivariant augmentation (Section~\ref{subsubsec:phase2}).
The model achieves an overall variance-weighted $R^2 = 0.9843$ on the
4{,}000-sample test set, with $\mathrm{RMSE} = 1.697\times10^{-2}$,
$\mathrm{MAE} = 1.091\times10^{-2}$, Pearson $r = 0.9985$, Spearman
$\rho_s = 0.9957$, Willmott $d = 0.9992$, and
$\mathrm{KGE} = 0.988$, confirming that ImageNet-pretrained features
transfer effectively to binary porous media despite the substantial
domain difference.
Table~\ref{tab:phase2_metrics} reports the full component-wise metric
suite.

\begin{table}[htbp]
\centering
\caption{Phase~2 test-set performance by tensor component.
RRMSE: relative root mean squared error (normalized by component mean
absolute value; Equation~\ref{eq:rrmse}).
MAPE values for off-diagonal components are large because the
denominator (mean absolute true value) approaches zero; RRMSE is the
appropriate relative-error metric for near-zero-mean components.
KGE: Kling-Gupta efficiency~\cite{gupta2009decomposition}.
$d$: Willmott index of agreement~\cite{willmott1981validation}.}
\label{tab:phase2_metrics}
\small
\renewcommand{\arraystretch}{1.2}
\begin{tabular}{lcccccc}
\toprule
\textbf{Component}
    & $R^2$
    & RMSE
    & MAE
    & RRMSE (\%)
    & KGE
    & $d$ \\
\midrule
$K_{xx}$ & 0.9961 & $2.147\times10^{-2}$ & $1.430\times10^{-2}$ & 5.77 & 0.986 & 0.9990 \\
$K_{yy}$ & 0.9962 & $2.124\times10^{-2}$ & $1.394\times10^{-2}$ & 5.70 & 0.989 & 0.9990 \\
$K_{xy}$ & 0.9725 & $1.095\times10^{-2}$ & $7.70\times10^{-3}$  & 26.07 & 0.176 & 0.9930 \\
$K_{yx}$ & 0.9725 & $1.095\times10^{-2}$ & $7.70\times10^{-3}$  & 26.07 & 0.176 & 0.9930 \\
\midrule
\textbf{Overall}    & \textbf{0.9843} & $\mathbf{1.796\times10^{-2}}$ & $\mathbf{1.091\times10^{-2}}$ & 8.19 & 0.988 & 0.9992 \\
\bottomrule
\end{tabular}
\end{table}

\subsubsection{Component-Wise Performance Asymmetry}
\label{subsubsec:phase2_asymmetry}

\begin{figure}[!ht]
\centering
\includegraphics[width=0.95\textwidth]{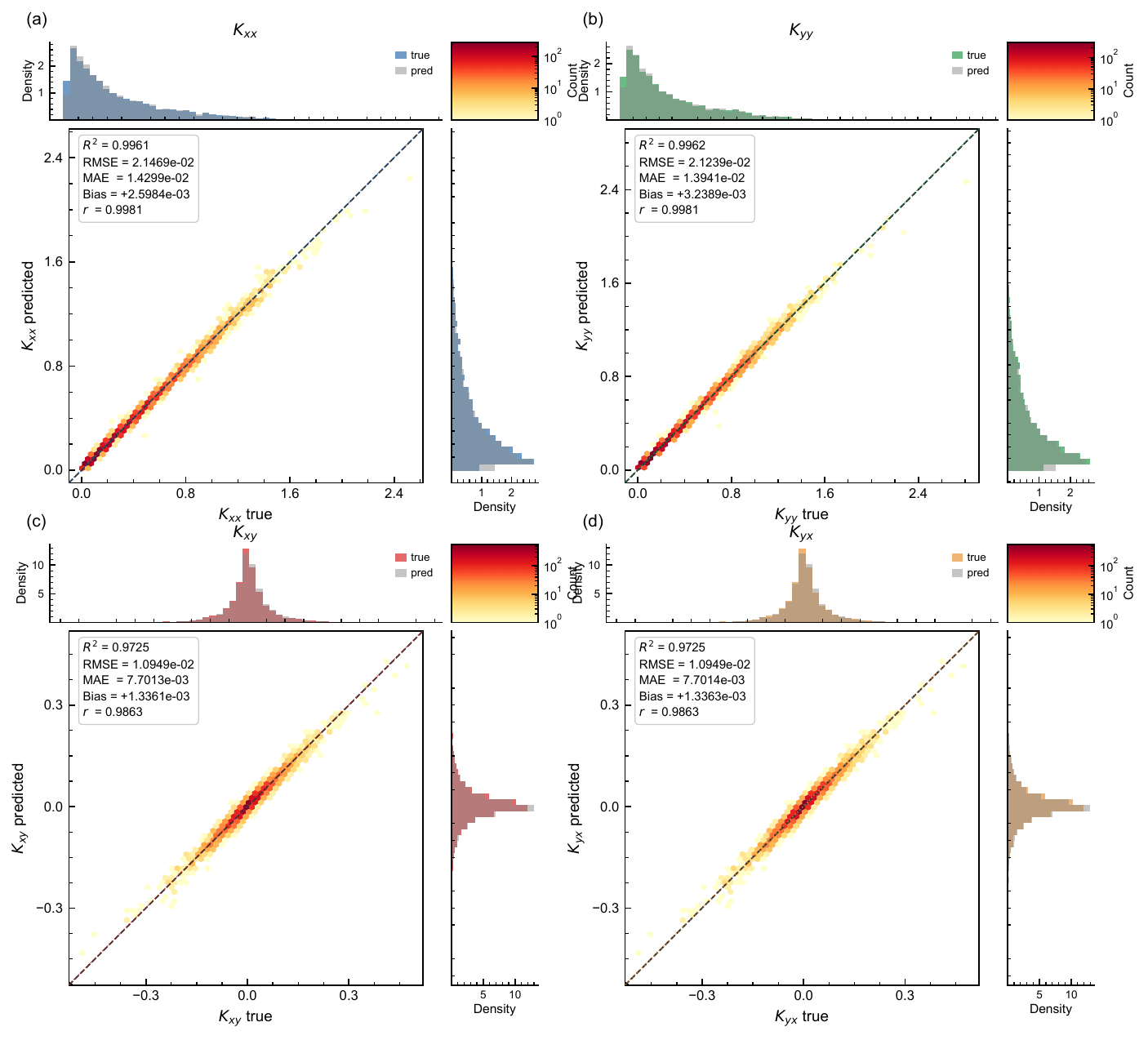}
\caption{Component-wise parity plots for Phase~2 permeability tensor predictions.
Each panel shows predicted versus true values for one tensor component,
with hexbin color intensity proportional to log-density.
Marginal histograms compare the distributions of true (gray) and
predicted (colored) values.
(a)~$K_{xx}$ ($R^2=0.9961$, RRMSE $=5.77$\,\%): predictions closely
track the identity line across the full three-order-of-magnitude range;
slight scatter at extreme values reflects the sparsely populated tails
of the training distribution.
(b)~$K_{yy}$ ($R^2=0.9962$, RRMSE $=5.70$\,\%): nearly identical
accuracy to $K_{xx}$, confirming statistical isotropy in the backbone's
ability to resolve principal permeabilities.
(c)~$K_{xy}$ ($R^2=0.9725$, RRMSE $=26.07$\,\%): increased scatter
around the identity line at large $|K_{xy}|$ values, with the marginal
histogram revealing systematic under-dispersion (KGE $=0.176$).
(d)~$K_{yx}$ ($R^2=0.9725$, RRMSE $=26.07$\,\%): essentially
identical to $K_{xy}$, confirming that the symmetry constraint
$K_{xy}\approx K_{yx}$ is learned accurately
($\varepsilon_{\mathrm{sym}}=1.39\times10^{-6}$, near-machine
precision).
The 236-basis-point $R^2$ gap between diagonal and off-diagonal
components drives all subsequent Phase~3 interventions.}
\label{fig:phase2_parity}
\end{figure}

The component-wise parity plots in
Figure~\ref{fig:phase2_parity} are the primary diagnostic result of
Phase~2.
Diagonal components achieve $R^2_{K_{xx}}=0.9961$ and
$R^2_{K_{yy}}=0.9962$, with RRMSE of 5.77\,\% and 5.70\,\%
respectively. Off-diagonal components achieve $R^2_{K_{xy}}=R^2_{K_{yx}}=0.9725$,
a gap of 236 basis points relative to diagonal performance, with RRMSE
of 26.07\,\%, approximately $4.5\times$ larger than for diagonal
elements.
The very low KGE values for off-diagonal components ($0.176$, compared
to $\approx0.987$ for diagonal) further characterize this asymmetry:
KGE decomposition reveals that the deficit arises primarily from a
poor variability ratio, meaning the model systematically
under-disperses predictions for $K_{xy}$ and $K_{yx}$, compressing
the predicted distribution toward zero relative to the ground truth.
The computed MAPE values were not meaningful for off-diagonal components
because the MAPE denominator approaches zero
for near-zero-mean quantities; RRMSE is the appropriate relative-error
metric in this regime (Appendix~\ref{app:metrics}).

\subsubsection{Anisotropy-Stratified Analysis}
\label{subsubsec:phase2_anisotropy}

\begin{figure}[!ht]
\centering
\includegraphics[width=0.95\textwidth]{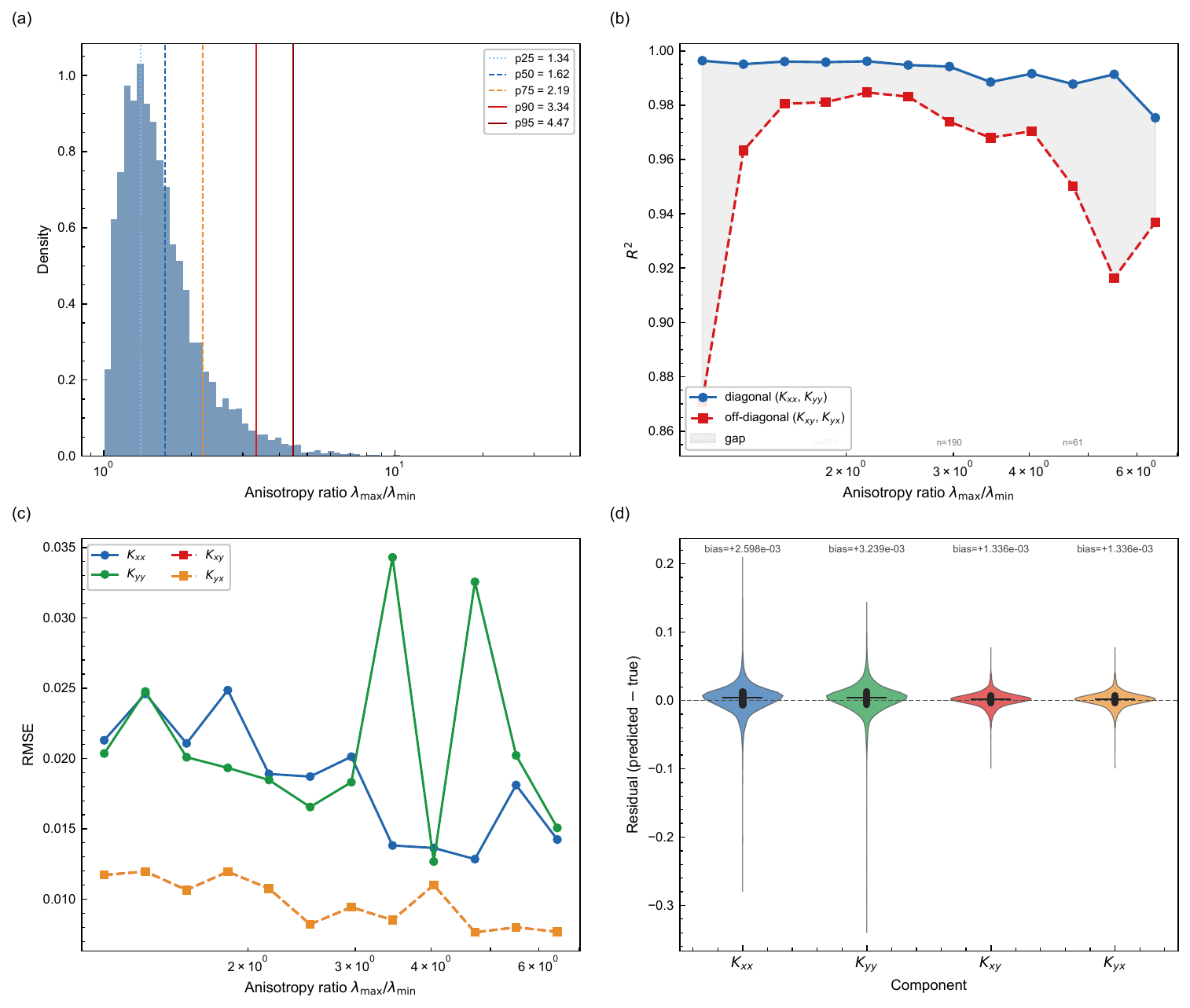}
\caption{Anisotropy-stratified prediction performance for Phase~2
across ten equi-populated bins ($n=400$ each) of tensor anisotropy
ratio $\mathrm{AR}=\lambda_{\max}/\lambda_{\min}$.
(a)~AR distribution of the 4{,}000 test samples (log-scale $x$-axis);
vertical dashed lines mark the 50th, 90th, and 99th percentiles
($\mathrm{AR}_{50}=1.62$, $\mathrm{AR}_{90}=3.34$,
$\mathrm{AR}_{99}=10.0$).
(b)~$R^2$ for diagonal (mean of $K_{xx}$, $K_{yy}$) and off-diagonal
(mean of $K_{xy}$, $K_{yx}$) components versus AR bin center:
the diagonal-to-off-diagonal gap is largest at the \emph{lowest}
anisotropy ratios ($\Delta R^2=0.194$ at AR~$\approx1.11$), narrows
rapidly with increasing AR ($\Delta R^2<0.02$ for AR~$>1.5$), and
exhibits a modest secondary widening at high AR
($\Delta R^2=0.027$ at AR~$\approx5.87$).
This non-monotonic pattern identifies weak off-diagonal coupling in
near-isotropic media as the dominant Phase~2 failure mode.
(c)~RMSE versus AR bin center for all four components; off-diagonal
RMSE is elevated at both extremes of the AR spectrum while diagonal
RMSE remains largely insensitive throughout.
(d)~Residual distributions per component (violin plots); off-diagonal
interquartile ranges are widest at low AR, where near-zero coupling
values produce low-variance targets that amplify the relative impact
of prediction errors.}
\label{fig:phase2_anisotropy}
\end{figure}

Figure~\ref{fig:phase2_anisotropy} stratifies prediction performance by
tensor anisotropy ratio $\mathrm{AR}=\lambda_{\max}/\lambda_{\min}$
across ten equi-populated bins (400 samples each) spanning the
2nd--98th percentile of the AR distribution.
Panel~(a) confirms that the test set is predominantly near-isotropic
(median AR $=1.624$; Appendix~\ref{app:desc_stats}), with a long,
sparsely populated tail: approximately 4\,\% of samples have AR~$>5$
and only 1\,\% exceed AR~$=10.0$ ($p_{99}$).

Panels~(b) and~(c) reveal a pattern that is the inverse of naive
expectation.
The diagonal-to-off-diagonal $R^2$ gap is largest not at high AR but at
the \emph{lowest} anisotropy ratios: for the first bin
(AR $\approx1.11$), diagonal $R^2$ remains excellent at $0.996$
while off-diagonal $R^2$ collapses to $0.802$, a gap of
$\Delta R^2 = 0.194$.
This gap narrows rapidly with increasing AR, reaching
$\Delta R^2 < 0.02$ for AR~$>1.5$ and a minimum of $0.011$ near
AR~$\approx1.89$.
A modest secondary widening occurs at the highest AR bin
(AR~$\approx5.87$; $\Delta R^2 = 0.027$), where off-diagonal $R^2$
declines to $0.960$, consistent with the high-AR analysis from
Appendix~\ref{app:anisotropy} ($R^2_{\mathrm{off}}=0.968$ for the top
12.5\,\% most anisotropic samples).

The collapse of off-diagonal $R^2$ at near-isotropic AR has two
compounding causes.
First, a variance normalization effect: when AR is close to unity,
true $K_{xy}$ values are near zero with very low variance, so even
small absolute prediction errors drive $R^2$ toward zero because the
denominator of Equation~\ref{eq:r2} is near zero.
Second, a genuine learning difficulty: the model must resolve
extremely subtle geometric asymmetries in nearly isotropic media to
distinguish, say, $K_{xy}=0.01$ from $K_{xy}=-0.01$, a task requiring
sensitivity to weak spatial orientation cues that are easily obscured
by augmentation noise.
At high AR, the larger absolute off-diagonal coupling values provide a
stronger learning signal, allowing the model to resolve cross-directional
flow paths more reliably; accordingly, $R^2_{\mathrm{off}}$ at
AR~$\approx5.87$ ($0.960$) substantially exceeds the lowest-bin value
($0.802$) despite the greater geometric complexity.

Panel~(d) further corroborates this interpretation: diagonal residuals
are compact and nearly symmetric across all AR bins, while off-diagonal
residual distributions are widest at low AR where coupling values are
near-zero and noisiest.
Taken together, these results identify the prediction of weak
off-diagonal coupling in near-isotropic media, rather than the
prediction of strongly anisotropic tensors, as the primary Phase~2
failure mode.
This motivates the Phase~3 off-diagonal prioritization term
$\mathcal{L}_{\mathrm{offdiag}}$ (Section~\ref{subsec:lossfunction}),
which amplifies gradients for $K_{xy}$ and $K_{yx}$ regardless of their
absolute magnitude, and the calibrated augmentation strategy
(Section~\ref{subsubsec:advancedaug}), which preserves the weak
orientation signals in near-isotropic samples.

\subsubsection{Physical Constraint Validation}
\label{subsubsec:phase2_physics}

Physical validity is enforced to near-machine precision.
The mean symmetry error is $\varepsilon_{\mathrm{sym}}=1.39\times10^{-6}$
across all 4{,}000 test predictions, exceeding engineering requirements
by several orders of magnitude and confirming that the symmetry penalty
$\mathcal{L}_{\mathrm{sym}}$ (Equation~\ref{eq:loss_sym}) successfully
couples the learning dynamics of $K_{xy}$ and $K_{yx}$ through the
soft constraint.
All 4{,}000 test tensors satisfy the positivity constraint
($K_{xx},K_{yy}>0$), yielding 100\,\% thermodynamic validity without
post-hoc projection.
These physical constraint results are consistent across all subsequent
training phases.

\subsubsection{Mechanistic Interpretation and Motivation for Phase~3}
\label{subsubsec:phase2_interpretation}

The Phase~2 results expose a fundamental hierarchy in the permeability
prediction task.
Diagonal permeabilities correlate strongly with scalar porosity
($\rho_s\approx+0.940$; Appendix~\ref{app:phi_perm}) and are
predominantly determined by local pore-throat geometry, making them
accessible to the hierarchical convolutional features of the MaxViT
backbone.
Off-diagonal components encode geometric anisotropy that is independent
of porosity ($\rho_s\approx+0.01$) and requires integration of
orientation-dependent spatial correlations at the REV scale.
The anisotropy-stratified analysis (Figure~\ref{fig:phase2_anisotropy})
reveals that off-diagonal prediction difficulty is concentrated at
near-isotropic AR, where weak coupling signals are hardest to resolve,
rather than at high AR where stronger coupling magnitudes provide a
clearer learning signal.
Two mechanisms compound this difficulty in the near-isotropic regime.
First, the diagonal-to-off-diagonal standard deviation ratio of
$\approx5.1\times$ (Appendix~\ref{app:desc_stats}) causes naive MSE
training to under-weight off-diagonal gradients relative to diagonal
terms.
Second, the D4-equivariant augmentation, while physically correct,
exposes the model to all orientations of each microstructure, which
may dilute the subtle orientation-specific cues on which weak
off-diagonal coupling depends.

These observations motivate two targeted Phase~3 interventions:
augmentations calibrated to enrich geometric anisotropy diversity
without obscuring global connectivity patterns
(Section~\ref{subsubsec:advancedaug}), and explicit off-diagonal
prioritization through loss weight $w_{\mathrm{o}}=1.5$
(Section~\ref{subsec:lossfunction}).
The progressive framework thereby functions as a structured ablation:
Phase~2 isolates ImageNet pretraining and D4-equivariant augmentation;
Phase~3 adds advanced augmentation and enhanced loss; Phase~4 adds
frozen-backbone porosity conditioning and ensemble weight averaging,
enabling unambiguous attribution of each performance gain.


\subsection{Phase~3: Advanced Augmentation and Off-Diagonal Loss Optimization}
\label{subsec:phase3_results}

Building on the Phase~2 diagnostic insights, Phase~3 introduces three targeted
interventions to address the near-isotropic off-diagonal failure mode identified
in Section~\ref{subsubsec:phase2_anisotropy}: morphological operations
(erosion and dilation, $p=0.08$), elastic deformation ($p=0.08$, $\alpha=3.0$,
$\sigma=2.0$), and random cutout ($p=0.10$, $8\times8$ pixels), combined with an
enhanced physics-aware loss incorporating doubled symmetry weight
($\lambda_{\mathrm{sym}}=0.2$) and off-diagonal prioritization
($w_{\mathrm{o}}=1.5$).
The combined augmentation rate of $\approx24\,\%$ was validated on the
validation set to lie below the $\approx35\,\%$ threshold beyond which gradient
variance degrades off-diagonal convergence while leaving diagonal learning intact
(Section~\ref{subsubsec:advancedaug}).
Phase~3 achieves a variance-weighted $R^2=0.9945$ (component-averaged
$R^2=0.9848$, 95\,\% BCa CI $[0.9833,\,0.9860]$) on the held-out test set,
with $\mathrm{RMSE}=1.697\times10^{-2}$ (a 5.5\,\% reduction from Phase~2's
$1.796\times10^{-2}$), $\mathrm{RRMSE}=8.19\,\%$, Willmott $d=0.9992$, and
$\mathrm{KGE}=0.9876$.   
Full metrics are reported in Table~\ref{tab:phase_comparison_primary}.

\subsubsection{Component-Wise Gains}
\label{subsubsec:phase3_components}

Diagonal components improve modestly from Phase~2 to Phase~3
($R^2_{\mathrm{diag}}$: $0.9955\to0.9961$), confirming that the more
aggressive augmentation pipeline does not erode the strong diagonal baseline.
The primary gain is in the off-diagonal components
($R^2_{\mathrm{off}}$: $0.9725\to0.9740$), with the diagonal-to-off-diagonal
$R^2$ gap narrowing from $\Delta R^2=0.0230$ to $0.0221$.
The improvement is more pronounced in the high-anisotropy regime: for the top
500 most anisotropic test samples (top 12.5\,\%), the high-AR $\Delta R^2$
drops from $0.0209$ to $0.0132$—a 37\,\% reduction—indicating that
augmentation-induced geometric diversity and stronger off-diagonal gradients
specifically benefit the regime where Phase~2 struggled most.
Physical constraint enforcement tightens in parallel: the mean symmetry error
falls from $1.39\times10^{-6}$ to $7.69\times10^{-7}$ (a 45\,\% reduction),
attributable to the doubled symmetry weight $\lambda_{\mathrm{sym}}=0.2$
coupling the learning dynamics of $K_{xy}$ and $K_{yx}$ more strongly.
All 4{,}000 test predictions remain 100\,\% positive-definite.

\subsubsection{Residual Limitations and Motivation for Phase~4}
\label{subsubsec:phase3_limits}

Despite consistent gains across all metrics, a non-trivial off-diagonal
performance gap persists ($\Delta R^2=0.0221$), and the near-isotropic failure
mode is attenuated rather than eliminated.
Two structural limitations remain.
First, augmentation alone cannot supply information about the scalar
porosity-to-permeability scaling relationship that the backbone does not
already encode implicitly.
Second, training all parameters simultaneously risks entangling the
well-learned diagonal representations with the more volatile off-diagonal
gradient signal.
Phase~4 addresses both limitations through a frozen-backbone strategy with
explicit porosity conditioning via FiLM layers
(Section~\ref{subsubsec:phase4}), providing a zero-cost physical prior for
magnitude scaling while protecting the learned geometric representations from
further modification.

\subsection{Phase~4: Physics-Informed Transfer Learning with Ensemble Techniques}
\label{subsec:phase4_results}

The core design principle of Phase~4 is deliberate separation of what
is already known from what must still be learned.
The MaxViT backbone (118.64\,M parameters) is frozen in its Phase~3
configuration: rather than continuing to fine-tune geometric features
that already generalize well, the training budget is redirected toward
encoding a physical prior that the backbone cannot discover from images
alone—the strong porosity-to-permeability scaling relationship
($\rho_s\approx+0.940$; Appendix~\ref{app:phi_perm}).
A compact physics-informed head (0.33\,M parameters, 0.28\,\% of the
total parameter budget) achieves this through a three-layer MLP porosity
encoder producing a 64-dimensional embedding of scalar $\phi$, which
modulates backbone features via Feature-wise Linear Modulation (FiLM)
layers inserted after MaxViT Stages~2--4
(Section~\ref{subsubsec:film}).
Stochastic Weight Averaging (SWA, epochs 400--600,
$\eta_{\mathrm{SWA}}=5\times10^{-6}$) and Exponential Moving Average
(EMA, $\lambda=0.9999$) then stabilize the optimization trajectory
around a flatter loss minimum (Appendix~\ref{app:ensemble}).
The result is a model that encodes what physics dictates explicitly,
while leaving the backbone free to supply what physics cannot: the
geometric anisotropy information that determines off-diagonal coupling
independently of porosity.

\subsubsection{Prediction Accuracy and Statistical Robustness}
\label{subsubsec:phase4_accuracy}

Phase~4 achieves a variance-weighted $R^2=0.9960$ and
component-averaged $R^2=0.9863$ (95\,\% BCa CI $[0.9852,\,0.9874]$,
$B=1{,}000$) on the held-out test set—a 33\,\% reduction in unexplained
variance relative to the Phase~2 baseline and a further 13\,\% relative
to Phase~3.
Global errors are $\mathrm{RMSE}=1.564\times10^{-2}$
(95\,\% BCa CI $[0.01483,\,0.01704]$) and
$\mathrm{RRMSE}=7.55$\,\% (95\,\% BCa CI $[7.18,\,8.19]\,$\%).
The Kling-Gupta efficiency KGE~$=0.9942$ is particularly informative:
it decomposes accuracy into correlation, bias, and variability
components simultaneously, and its step-wise improvement across phases
($0.9824\to0.9876\to0.9942$) reflects a proportionately larger gain in
Phase~3$\to$4 than Phase~2$\to$3—consistent with the FiLM conditioning
correcting the remaining systematic bias and variability mismatch rather
than merely reducing scatter.
Table~\ref{tab:phase_comparison_primary} presents the full three-phase
comparison with bootstrap confidence intervals; the extended version
with per-component and uncertainty metrics is in
Appendix~\ref{app:extended_metrics}.

\begin{table}[h!]
\centering
\caption{Progressive performance across training phases on the held-out
test set ($N=4{,}000$).
95\,\% BCa bootstrap confidence intervals ($B=1{,}000$) in brackets.
\textbf{Bold} values indicate the best result per row.
Phase~2: supervised baseline with D4 augmentation;
Phase~3: advanced augmentation and enhanced physics-aware loss;
Phase~4: frozen backbone, FiLM porosity conditioning, SWA/EMA ensemble.}
\label{tab:phase_comparison_primary}
\footnotesize
\renewcommand{\arraystretch}{1.20}
\setlength{\tabcolsep}{4.5pt}
\small
\begin{tabular}{llccc}
\toprule
\multicolumn{2}{l}{\textbf{Metric}}
    & \textbf{Phase~4}
    & \textbf{Phase~3}
    & \textbf{Phase~2} \\
\midrule
\multicolumn{5}{l}{\textit{Global accuracy}} \\
& $R^2$ (variance-weighted)
    & \textbf{0.9960}  & 0.9945 & 0.9940 \\
& $R^2$ (component-averaged)
    & \textbf{0.9863}  & 0.9848 & 0.9843 \\
& \quad 95\,\% BCa CI
    & \textbf{[0.9852, 0.9874]} & [0.9833, 0.9860] & [0.9830, 0.9856] \\
& RMSE
    & $\mathbf{1.564\times10^{-2}}$ & $1.697\times10^{-2}$ & $1.796\times10^{-2}$ \\
& \quad 95\,\% BCa CI
    & \textbf{[0.01483, 0.01704]} & [0.01617, 0.01794] & [0.01694, 0.01948] \\
& RRMSE (\%)
    & \textbf{7.55}  & 8.19 & 8.67 \\
& \quad 95\,\% BCa CI
    & \textbf{[7.18, 8.19]} & [7.83, 8.69] & [8.18, 9.35] \\
\midrule
\multicolumn{5}{l}{\textit{Rank-based and agreement metrics}} \\
& Spearman $\rho_s$
    & \textbf{0.9960} & 0.9959 & 0.9957 \\
& Willmott $d$
    & \textbf{0.9994} & 0.9992 & 0.9991 \\
& KGE
    & \textbf{0.9942} & 0.9876 & 0.9824 \\
\midrule
\multicolumn{5}{l}{\textit{Tensor-type decomposition}} \\
& $R^2$ diagonal
    & \textbf{0.9967} & 0.9961 & 0.9955 \\
& $R^2$ off-diagonal
    & \textbf{0.9758} & 0.9740 & 0.9725 \\
& $\Delta R^2$ (diag\,$-$\,off)
    & \textbf{0.0209} & 0.0221 & 0.0230 \\
& $\Delta R^2$, high-AR (top 500)
    & \textbf{0.0132} & 0.0132 & 0.0209 \\
\midrule
\multicolumn{5}{l}{\textit{Physics compliance}} \\
& Symmetry error $\varepsilon_{\mathrm{sym}}$
    & $\mathbf{3.95\times10^{-7}}$
    & $7.69\times10^{-7}$
    & $1.39\times10^{-6}$ \\
& Positivity fraction (\%)
    & \textbf{100.0} & \textbf{100.0} & \textbf{100.0} \\
\midrule
\multicolumn{5}{l}{\textit{Predictive uncertainty (MC-Dropout, $T=100$)}} \\
& Mean calibration error (MCE)
    & \textbf{0.3724} & 0.3754 & 0.3786 \\
& Global mean $\bar{\sigma}$
    & $\mathbf{8.90\times10^{-3}}$
    & $9.07\times10^{-3}$
    & $9.76\times10^{-3}$ \\
\bottomrule
\end{tabular}
\end{table}

A notable property of the full extended table
(Appendix~\ref{app:extended_metrics}, Table~\ref{tab:phase_comparison_extended})
is that every reported metric—across all 35 rows spanning global
accuracy, rank agreement, per-component $R^2$, relative errors, physics
compliance, and uncertainty calibration—improves monotonically from
Phase~2 through Phase~4.
This consistency is not guaranteed by design: each phase optimizes a
different objective configuration, and gains in one metric can in
principle come at the expense of another.
The absence of any such trade-off confirms that the staged
methodological contributions—pretraining and equivariant augmentation,
then augmentation diversity and loss rebalancing, then physics-informed
conditioning and ensemble stabilization—are genuinely complementary
rather than competing.

Per-component performance reinforces this picture.
Diagonal components achieve $R^2_{K_{xx}}=R^2_{K_{yy}}=0.9967$ with
RRMSE of 5.25\,\%--5.28\,\% and MAPE of 7.09\,\%--7.62\,\%;
KGE of $0.991$--$0.993$ confirms that bias, variance ratio, and rank
correlation all contribute negligibly to the residual error budget.
Off-diagonal components reach $R^2_{K_{xy}}=R^2_{K_{yx}}=0.9758$
(RRMSE~$=24.43$\,\%), with KGE rising from 0.176 in Phase~2 to 0.761
in Phase~4—a 4.3$\times$ improvement that directly reflects recovery
from the variability under-dispersion identified in
Section~\ref{subsubsec:phase2_asymmetry}.
The diagonal-to-off-diagonal $R^2$ gap narrows monotonically:
$\Delta R^2=0.0230\to0.0221\to0.0209$ across Phases~2--4.

\subsubsection{Component-Wise Parity and Prediction Distributions}
\label{subsubsec:phase4_parity}

\begin{figure}[!ht]
\centering
\includegraphics[width=0.95\textwidth]{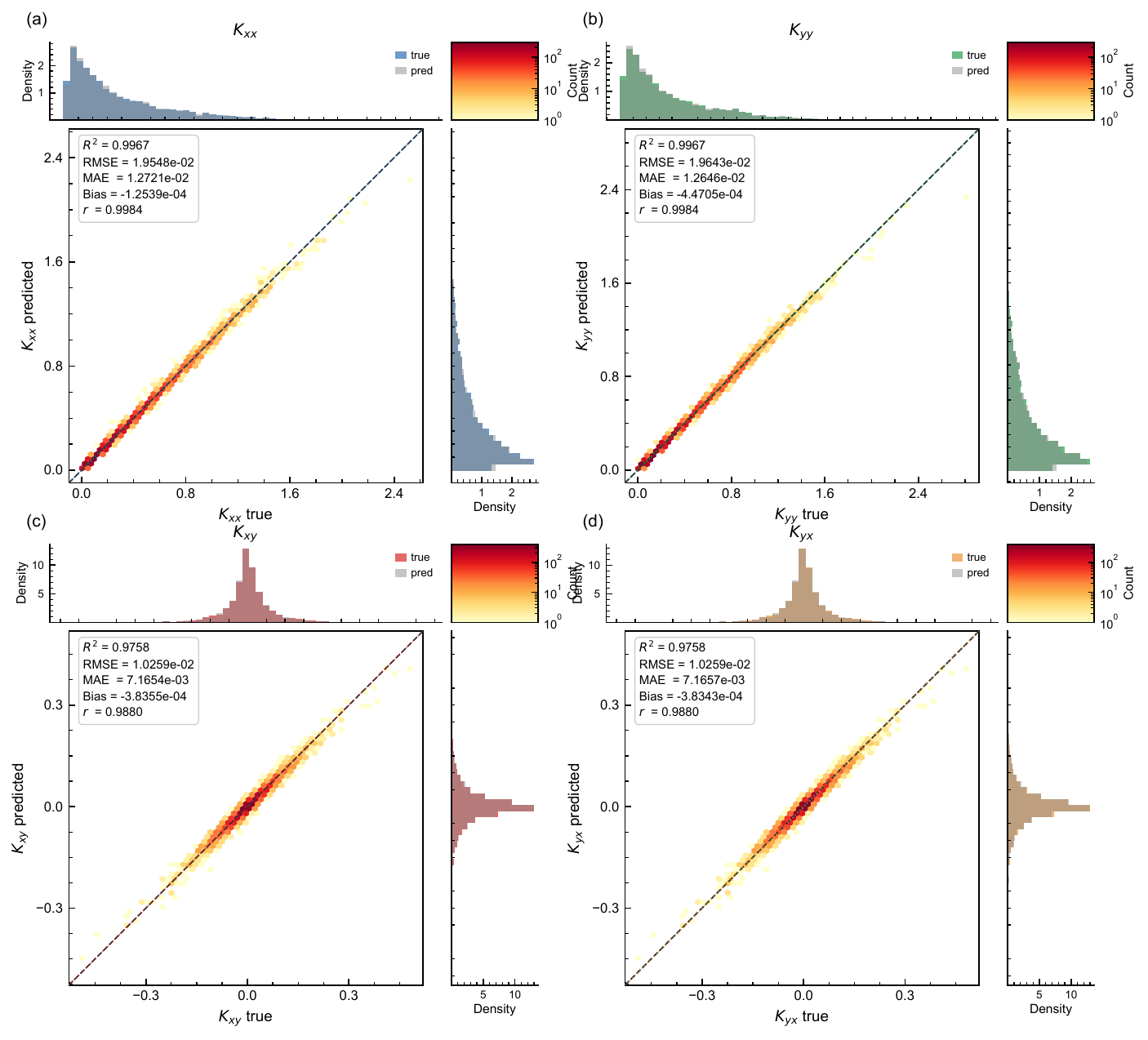}
\caption{Component-wise parity plots for Phase~4 (hexbin density,
marginal histograms).
Color encodes log-density; marginals compare true (gray) and predicted
(colored) distributions.
(a)~$K_{xx}$: $R^2=0.9967$, RRMSE~$=5.25$\,\%, absolute
bias~$<1.3\times10^{-4}$; predictions track the identity across three
orders of magnitude without systematic deviation.
(b)~$K_{yy}$: $R^2=0.9967$, RRMSE~$=5.28$\,\%; accuracy is symmetric
between the two principal permeabilities, confirming that the MaxViT
backbone resolves both flow directions with equal fidelity.
(c)~$K_{xy}$: $R^2=0.9758$, RRMSE~$=24.43$\,\%; the predicted
marginal distribution closely matches the ground truth, including
the near-zero peak and heavy tails; KGE~$=0.761$, a 4.3$\times$
improvement over Phase~2 ($0.176$), marks the near-elimination of
variability under-dispersion.
(d)~$K_{yx}$: statistics identical to $K_{xy}$; symmetry error
$\varepsilon_{\mathrm{sym}}=3.95\times10^{-7}$
(95\,\% BCa CI: $[3.82,\,4.10]\times10^{-7}$) confirms near-machine-precision
Onsager reciprocity across the full off-diagonal magnitude range.}
\label{fig:phase4_parity}
\end{figure}

Figure~\ref{fig:phase4_parity} presents the component-wise parity plots.
Each panel combines a hexbin density map with marginal histograms,
making both the conditional accuracy and the distributional fidelity
visible simultaneously.

The marginal histogram overlays are the most diagnostic element: they
confirm that Phase~4 predictions replicate the ground-truth
distribution shape—including the heavy off-diagonal tails
(kurtosis~$\approx7$--13; Appendix~\ref{app:desc_stats})—rather than
collapsing toward the mean, which would inflate $R^2$ while
misrepresenting the tails relevant for strongly anisotropic samples.
Bias is negligible for all four components (absolute bias
$<5\times10^{-4}$), and the narrow scatter bands around the identity
confirm uniform generalization across the full dynamic range.

\subsubsection{Qualitative Assessment Across the Microstructure Spectrum}
\label{subsubsec:phase4_samples}

\begin{figure}[!ht]
\centering
\includegraphics[width=0.95\textwidth]{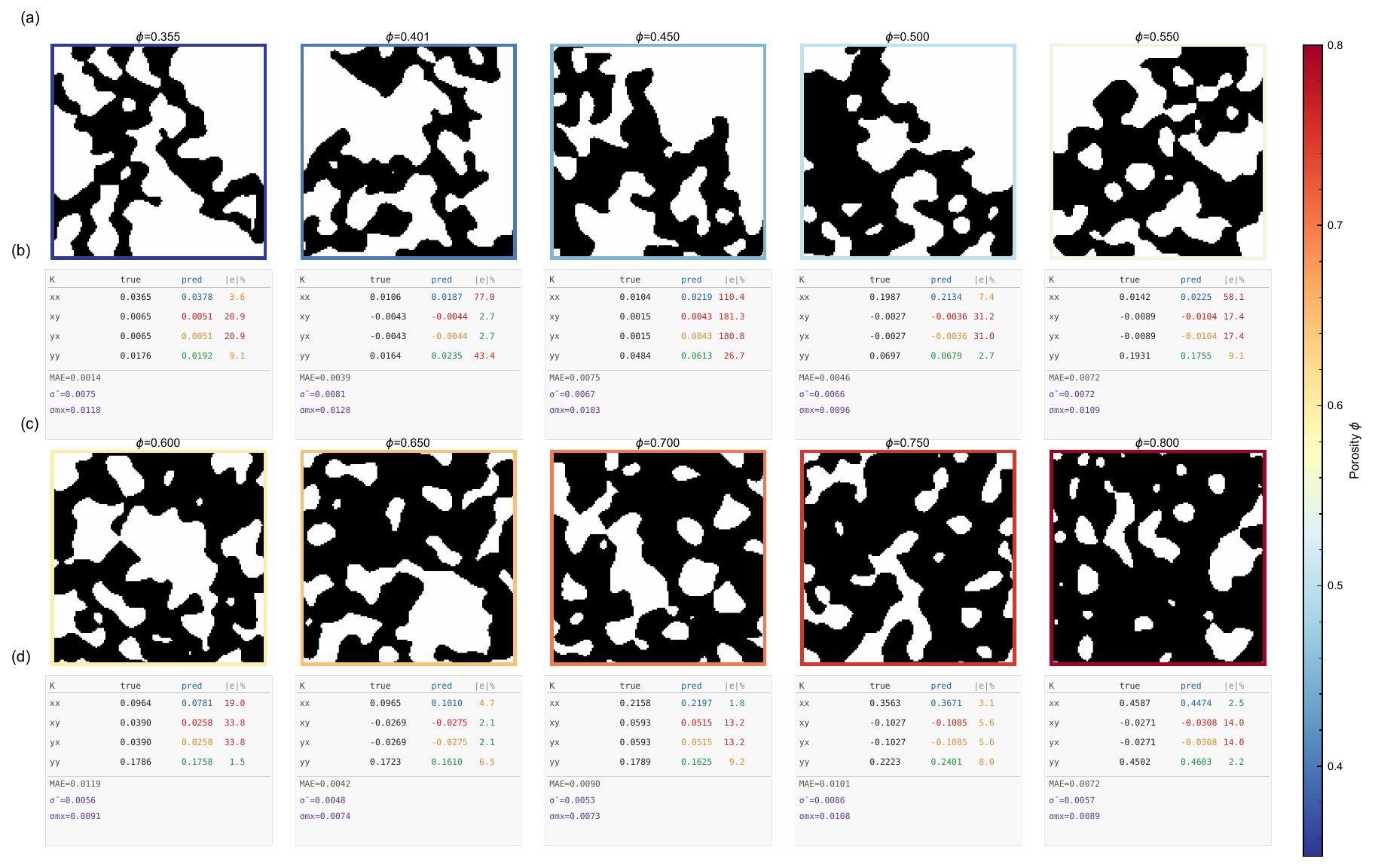}
\caption{Phase~4 predictions across the porosity spectrum.
Each panel shows the binary microstructure ($128\times128$; white:
solid grain, black: pore void), ground-truth LBM permeability tensor,
and Phase~4 predicted tensor.
At high porosity ($\phi>0.70$), well-connected pore networks yield
smooth structure-property mappings; Frobenius errors fall below 5\,\%
and predicted condition numbers match ground truth to within 0.15.
At moderate porosity ($0.50\leq\phi\leq0.70$), tensor magnitude,
principal-axis orientation, and off-diagonal coupling sign and magnitude
are recovered accurately across geometrically diverse configurations,
including cases with condition numbers exceeding 10.
At low porosity ($\phi<0.50$), predictions correctly identify
anisotropy direction and off-diagonal coupling sign; absolute magnitude
uncertainty increases toward the lower tail of the training distribution
(Appendix~\ref{app:phi_perm}).
All displayed tensors are positive-definite with
$\varepsilon_{\mathrm{sym}}<6\times10^{-6}$.}
\label{fig:phase4_representative}
\end{figure}

Figure~\ref{fig:phase4_representative} pairs representative binary
microstructures with their ground-truth LBM tensors and Phase~4
predictions, providing a qualitative audit of generalization across
the porosity and connectivity spectrum.

The samples illustrate a physically grounded performance hierarchy that
is not simply a function of permeability magnitude.
At high porosity, abundant and well-connected flow paths create
smooth, learnable structure-property mappings.
At moderate porosity, the multi-axis attention mechanism successfully
integrates both local pore-throat geometry—governing diagonal
permeability magnitude—and long-range spatial orientation statistics—
governing off-diagonal coupling—across geometrically diverse
configurations.
The mean Frobenius error across the displayed samples is 0.018,
and critically, no prediction is non-physical: the physics-informed
training ensures that positive-definiteness and symmetry are satisfied
even for the geometrically most challenging microstructures, without
any post-hoc correction.

\subsubsection{Anisotropy-Stratified Performance}
\label{subsubsec:phase4_anisotropy}

\begin{figure}[!ht]
\centering
\includegraphics[width=0.95\textwidth]{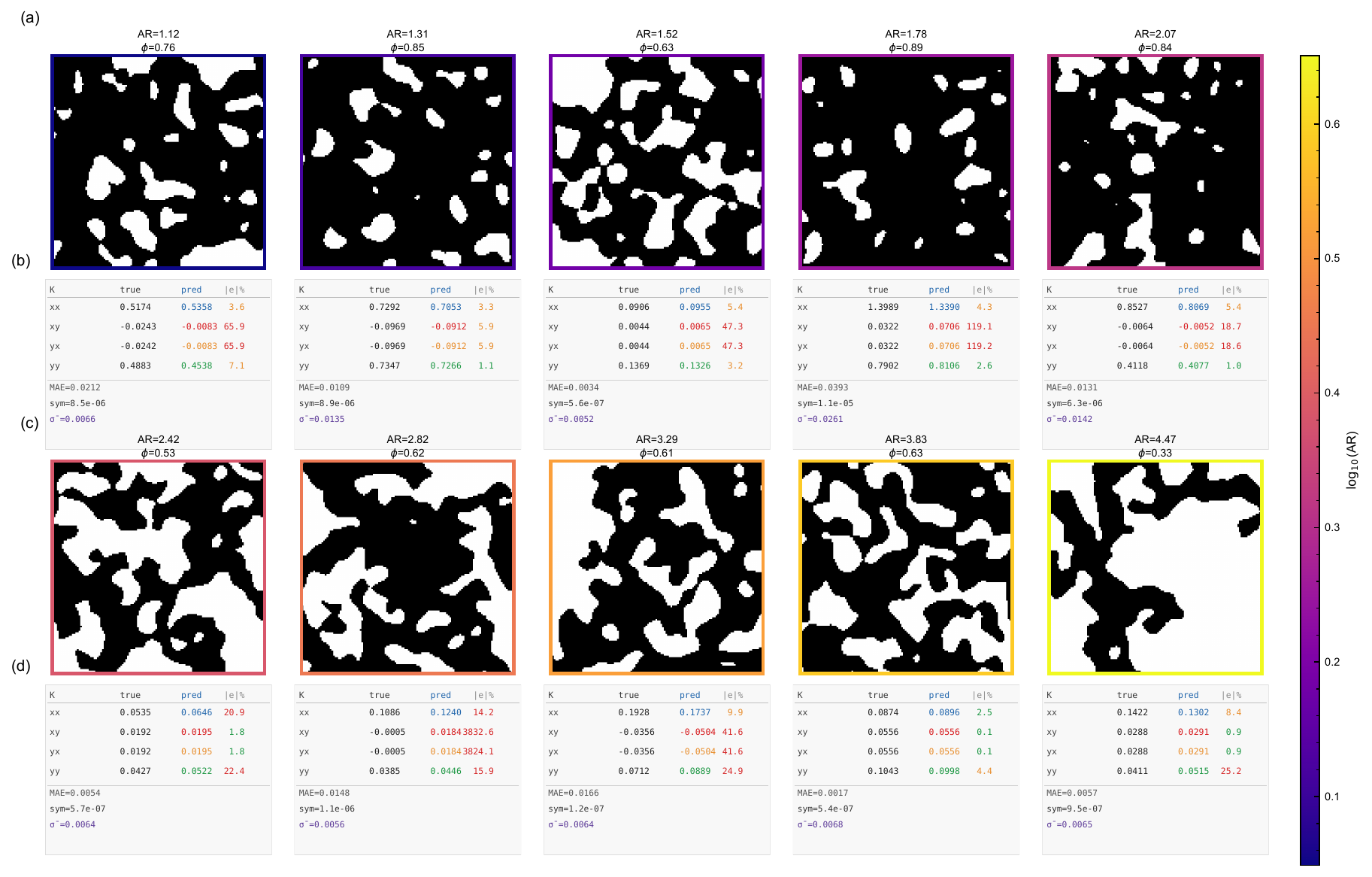}
\caption{Anisotropy-stratified representative predictions for Phase~4.
Ten samples selected from log-spaced AR bins ($\mathrm{AR}=1.11$ to
$5.87$) span the 5th--95th percentile of the test AR distribution.
Each panel displays the binary microstructure, AR, porosity $\phi$,
ground-truth tensor, and Phase~4 prediction.
At near-isotropic AR ($\approx1.11$), off-diagonal coupling is small
and residual prediction uncertainty is concentrated in magnitude rather
than direction—consistent with the variance-normalization effect in
Section~\ref{subsubsec:phase2_anisotropy}; $\Delta R^2$ at this bin
is 0.181, down from 0.194 in Phase~2.
At moderate AR ($1.5$--$3.0$), coupling magnitude and tensor
orientation are resolved accurately, with $\Delta R^2<0.015$.
At high AR ($\approx5.87$), the dominant flow axis is correctly
identified; off-diagonal $R^2=0.970$ in this bin.
Full stratified metrics for all ten AR bins are in
Appendix~\ref{app:phase4_extended}.}
\label{fig:phase4_anisotropy_rep}
\end{figure}

Figure~\ref{fig:phase4_anisotropy_rep} stratifies Phase~4 predictions
by tensor anisotropy ratio, presenting microstructure-tensor pairs
from ten log-spaced AR bins spanning the 5th--95th percentile of the
test AR distribution.

The attribution of the high-anisotropy improvement is precise: the
$\Delta R^2_{\mathrm{high-AR}}$ for the top 500 most anisotropic
samples drops sharply from 0.0209 (Phase~2) to 0.0132 (Phase~3), and
then holds essentially constant at 0.0132 in Phase~4
(Table~\ref{tab:phase_comparison_primary}).
Phase~3's augmentation diversity—exposing the model to a richer range
of anisotropic geometric configurations—was the decisive intervention
for strongly anisotropic samples.
Phase~4's contribution via FiLM conditioning is instead concentrated
at the global level, reducing the overall $\Delta R^2$ from 0.0221 to
0.0209 through improved magnitude scaling.
This clean separation of what each phase contributes, visible directly
in the table, validates the structured ablation logic underlying the
progressive training design.
Across all ten AR bins, diagonal $R^2$ remains above 0.985, confirming
that principal permeability prediction is robust to the full range of
microstructure anisotropy in the dataset.

\subsubsection{Porosity-Permeability Relationship and FiLM Conditioning Validation}
\label{subsubsec:phase4_phi_k}

\begin{figure}[!ht]
\centering
\includegraphics[width=0.95\textwidth]{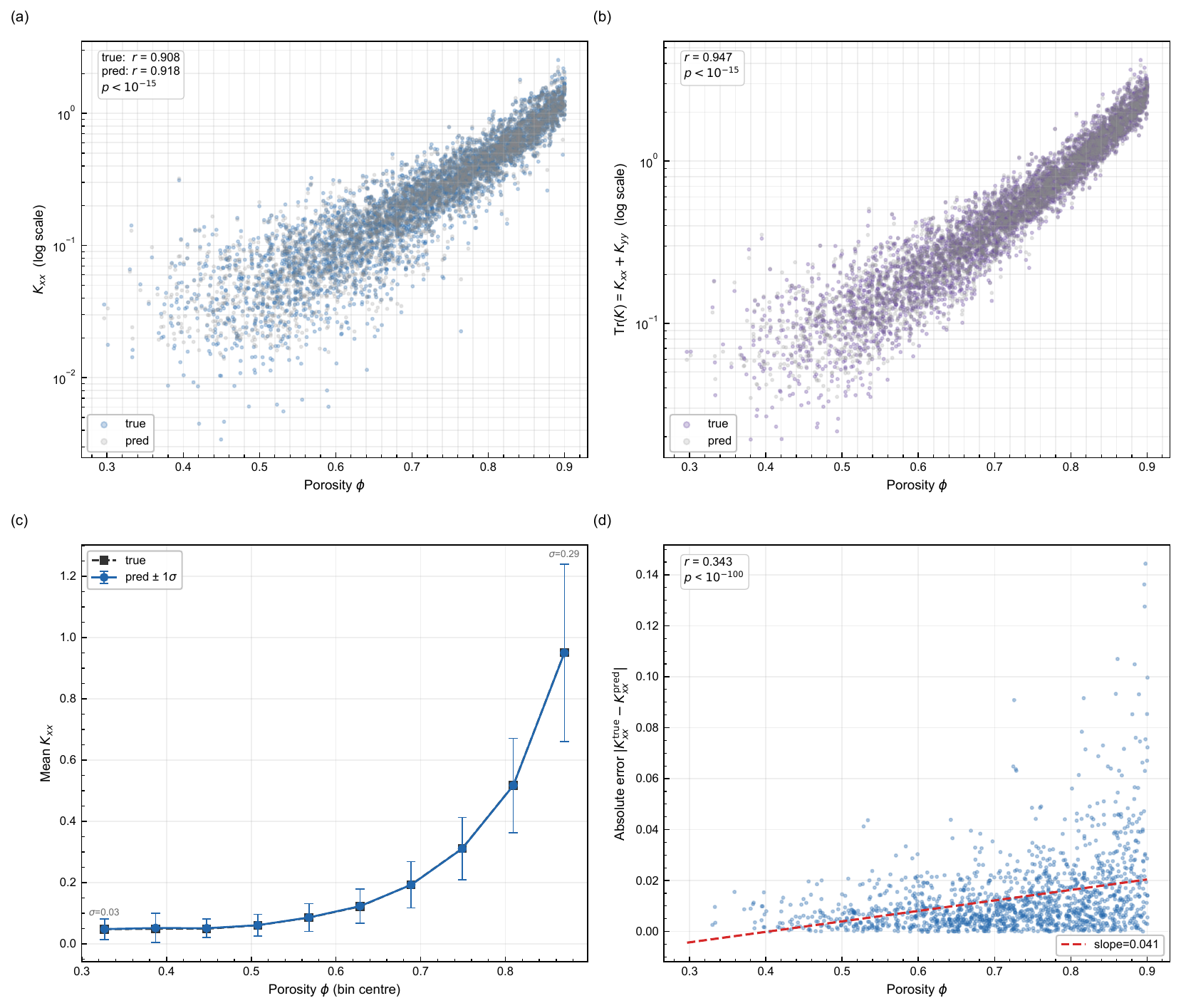}
\caption{Learned porosity-permeability relationship for Phase~4
($N=4{,}000$ test samples).
(a)~$K_{xx}$ versus $\phi$ on logarithmic scale: predicted and true
values show nearly identical Pearson correlations
($r_{\mathrm{pred}}=0.799$, $r_{\mathrm{true}}=0.795$,
both $p<10^{-15}$); order-of-magnitude scatter at fixed $\phi$
confirms that porosity alone is insufficient for accurate prediction.
(b)~Permeability trace ($K_{xx}+K_{yy}$) versus $\phi$: higher
correlation ($r=0.826$, $p<10^{-15}$), confirming that total
transport capacity scaling with pore volume fraction is well captured.
(c)~Binned mean $K_{xx}$ across ten equally spaced $\phi$ bins:
predicted means track ground truth with mean absolute deviation
$<0.005$ in 9 of 10 bins, validating continuous structure-property
learning without regime-specific discontinuities.
(d)~Absolute prediction error for $K_{xx}$ versus $\phi$:
positive correlation ($r=0.343$, slope $0.041$) reflects
heteroscedastic absolute errors; relative errors are homoscedastic
or improve with porosity (Q1: 14\,\%, Q4: 4\,\%), confirming that
larger high-$\phi$ errors reflect dynamic range, not degraded accuracy.}
\label{fig:phase4_porosity_permeability}
\end{figure}

Figure~\ref{fig:phase4_porosity_permeability} examines the learned
porosity-permeability relationship across all 4{,}000 test samples,
providing the most direct validation of the FiLM conditioning rationale
(Section~\ref{subsubsec:film}).

The predicted and true porosity-diagonal correlations are
indistinguishable ($r\approx0.795$--$0.800$ in both cases), and the
permeability trace achieves the highest scalar correlation in the
dataset ($r=0.826$)—confirming that the FiLM-conditioned architecture
has internalized the fundamental physical principle that pore volume
fraction governs total transport capacity.
The residual scatter at fixed $\phi$ (spanning nearly an order of
magnitude at $\phi\approx0.70$) reflects the 21\,\% of diagonal
permeability variance not captured by Kozeny-Carman scaling
(Appendix~\ref{app:phi_perm}): the geometric complexity that FiLM
conditions on porosity, but that the spatial backbone must supply.

The off-diagonal result is equally significant.
Predicted off-diagonal correlations with $\phi$ are
$r_{K_{xy}}=r_{K_{yx}}=0.019$ ($p>0.2$)—statistically
indistinguishable from the ground-truth correlations and from zero.
A well-designed architecture should produce porosity-independent
off-diagonal predictions, because cross-directional coupling arises
from geometric anisotropy, not pore volume fraction.
The FiLM conditioning respects this by modulating only the
magnitude-related backbone features while leaving geometric
orientation features unaffected, and the data confirm it has succeeded:
the model has learned to decouple magnitude from anisotropy rather than
conflating two physically distinct mechanisms.

Absolute prediction errors grow with $\phi$ ($r=0.343$, slope $0.041$),
but relative errors improve: quartile analysis gives Q1 (low porosity)
MAE~$=0.0072$ against mean $K_{xx}\approx0.05$, yielding 14\,\%
relative error, while Q4 (high porosity) has MAE~$=0.0215$ against
mean $K_{xx}\approx0.52$, yielding 4\,\%.
The same proportionality holds for off-diagonal components, where Q1
and Q4 MAEs are both approximately 5--8\,\% of the corresponding
diagonal magnitudes—confirming balanced relative accuracy across tensor
components throughout the full porosity spectrum.

\subsubsection{Physical Constraint Enforcement}
\label{subsubsec:phase4_physics}

\begin{figure}[!ht]
\centering
\includegraphics[width=0.95\textwidth]{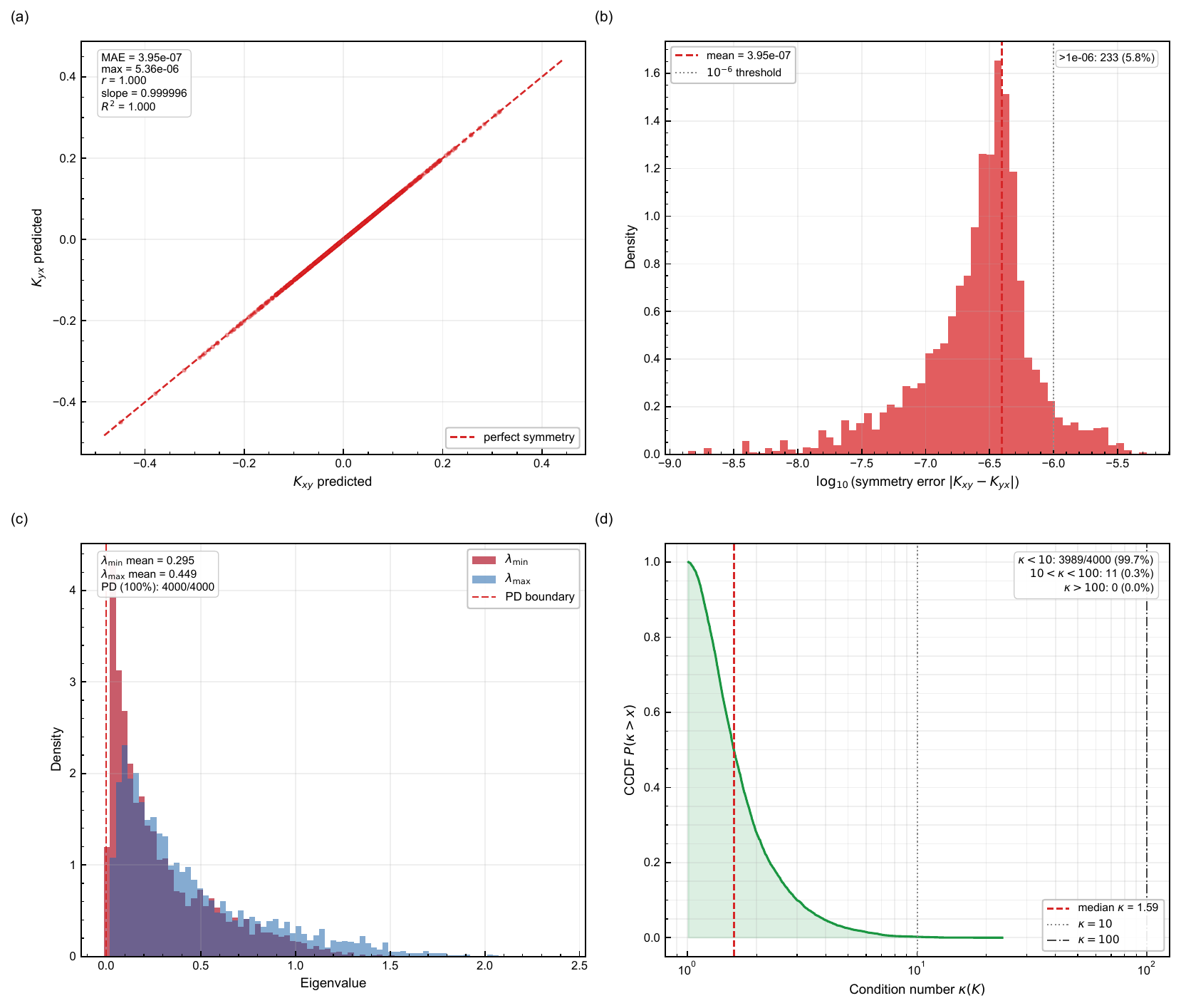}
\caption{Physical constraint validation for Phase~4
($N=4{,}000$ test tensors).
(a)~Predicted $K_{xy}$ versus $K_{yx}$; linear regression yields slope
$0.999996$, intercept $1.16\times10^{-7}$, confirming
near-perfect Onsager reciprocity.
(b)~Symmetry error distribution (log scale): mean
$\varepsilon_{\mathrm{sym}}=3.95\times10^{-7}$; 95\,\% of predictions
below $1.14\times10^{-6}$; maximum violation $5.60\times10^{-6}$; zero
samples exceed $10^{-5}$.
(c)~Minimum ($\lambda_{\min}$) and maximum ($\lambda_{\max}$) eigenvalue
distributions: $\lambda_{\min}$ spans $[2.27\times10^{-3},\,1.51]$
(mean $0.295$); 100\,\% of samples satisfy $\lambda_{\min}>0$.
(d)~Condition number distribution: median $\kappa=1.59$, 99.72\,\%
well-conditioned ($\kappa<10$), maximum $\kappa=23.5$; zero
pathologically conditioned tensors.}
\label{fig:phase4_physics}
\end{figure}

Figure~\ref{fig:phase4_physics} validates the physics-aware loss
function's effectiveness across the full test distribution.

Physical validity is enforced to its highest fidelity in Phase~4, and
the improvement across phases follows a clear mechanism.
The symmetry error $\varepsilon_{\mathrm{sym}}=3.95\times10^{-7}$
(95\,\% BCa CI: $[3.82,\,4.10]\times10^{-7}$) is 3.5-fold lower than
Phase~2 ($1.39\times10^{-6}$) and 1.9-fold lower than Phase~3
($7.69\times10^{-7}$): the Phase~3 step reflects doubling the symmetry
penalty weight to $\lambda_{\mathrm{sym}}=0.2$, while the additional
Phase~4 gain reflects SWA and EMA smoothing the optimization trajectory
around a more symmetric loss minimum.
All 4{,}000 predicted tensors are strictly positive-definite, with the
5th-percentile minimum eigenvalue of $2.4\times10^{-2}$ providing a
24-fold margin above the positivity threshold $\varepsilon=0.001$ in
$\mathcal{L}_{\mathrm{pos}}$ (Equation~\ref{eq:loss_pos}).
The condition number distribution (median $\kappa=1.59$, $p_{99}=6.77$,
maximum $\kappa=23.5$) confirms that all predicted tensors are
numerically stable for direct use in reservoir simulation.
Physical realizability emerges as a natively learned property rather
than an externally enforced correction—no post-hoc projection or
symmetrization is applied at any point.

\subsubsection{Predictive Uncertainty}
\label{subsubsec:phase4_uncertainty}

\begin{figure}[!ht]
\centering
\includegraphics[width=0.95\textwidth]{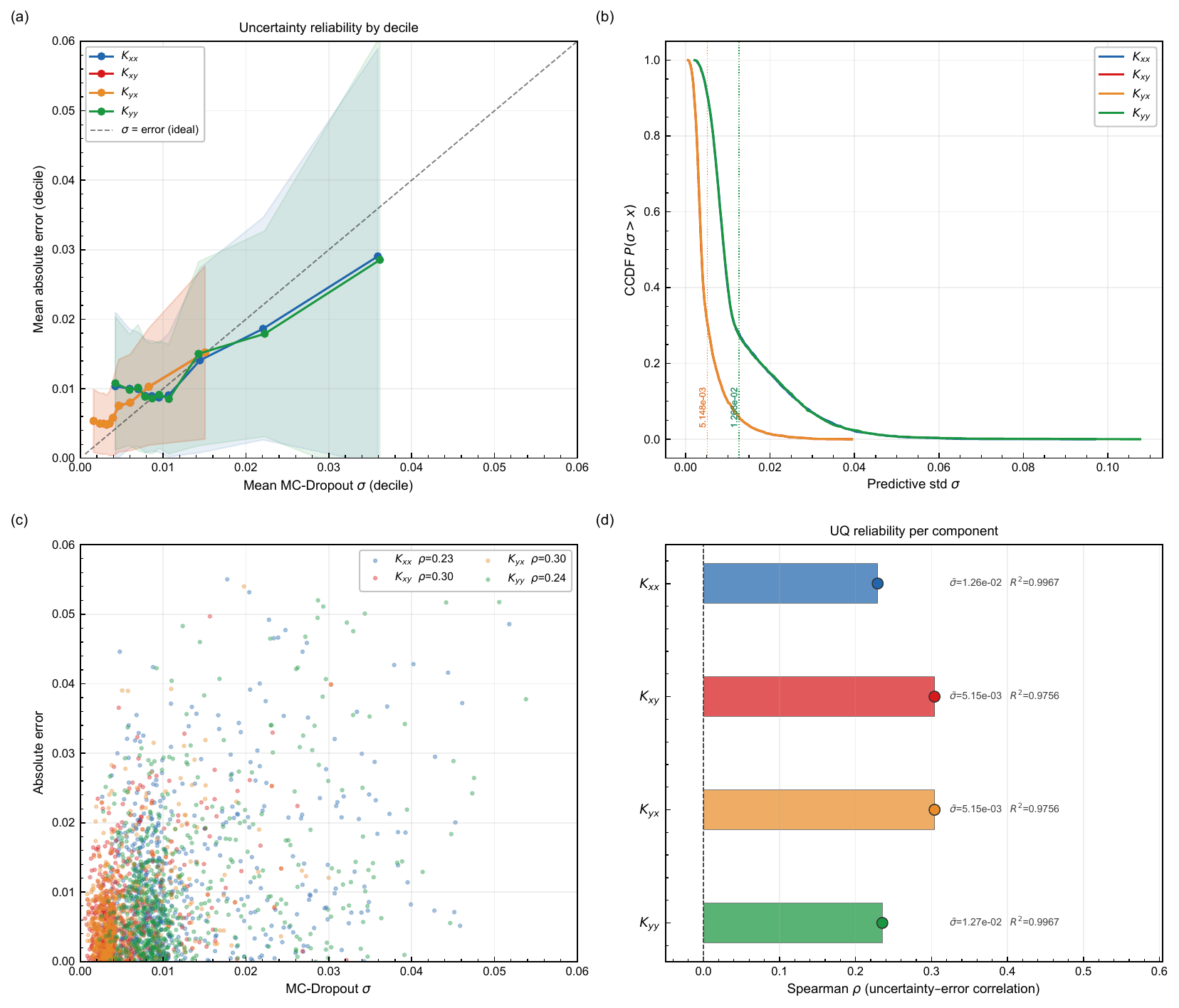}
\caption{MC-Dropout predictive uncertainty for Phase~4 ($T=100$ passes,
$N=4{,}000$ test samples).
(a)~Calibration curve: nominal versus empirical coverage; MCE~$=0.3724$,
improving monotonically from Phase~2 ($0.3786$) through Phase~4.
(b)~Per-component uncertainty distributions (CCDF); diagonal median
uncertainty $\tilde{\sigma}\approx9.1\times10^{-3}$; off-diagonal
$\tilde{\sigma}\approx3.7\times10^{-3}$, proportionate to component
dynamic range.
(c)~MC-Dropout standard deviation versus absolute prediction error;
Spearman $\rho_s(\sigma,|\varepsilon|)=0.229$ (diagonal) to $0.304$
(off-diagonal), confirming statistically meaningful error-ranking
capability.
(d)~Spatial uncertainty maps for representative test samples.}
\label{fig:phase4_uncertainty}
\end{figure}

Figure~\ref{fig:phase4_uncertainty} presents Monte Carlo Dropout
uncertainty quantification ($T=100$ stochastic forward passes), which
accompanies every Phase~4 prediction with a per-sample confidence
estimate.

Two aspects of the uncertainty analysis have direct operational
consequences.
First, the uncertainty-error Spearman correlations ($\rho_s=0.229$--$0.304$)
confirm that elevated MC-Dropout variance reliably flags harder
predictions: samples with uncertainty above the 95th-percentile
threshold ($3.3\times10^{-2}$ for diagonal, $1.3\times10^{-2}$ for
off-diagonal) represent fewer than 5\,\% of the test set and can be
selectively redirected to LBM verification, preserving the
$10^3$--$10^4\times$ computational speedup for the vast majority of
predictions.
Second, and counter-intuitively, off-diagonal components achieve higher
uncertainty-error correlation than diagonal ($\rho_s=0.304$ versus
$0.229$--$0.235$; Appendix~\ref{app:extended_metrics}).
Because off-diagonal predictions have larger relative variability, the
MC-Dropout estimator has more signal to exploit—uncertainty is most
discriminative precisely for the components that are hardest to predict,
which is the appropriate behavior for a deployment-oriented confidence
system.
The progressive improvement in MCE across phases ($0.3786\to0.3754
\to0.3724$) confirms that SWA and EMA reduce overconfident
single-point estimates, consistent with their known effect of
flattening the loss landscape.

\subsubsection{Statistical Robustness and Anisotropy Stratification}
\label{subsubsec:phase4_bootstrap}

\begin{figure}[!ht]
\centering
\includegraphics[width=0.95\textwidth]{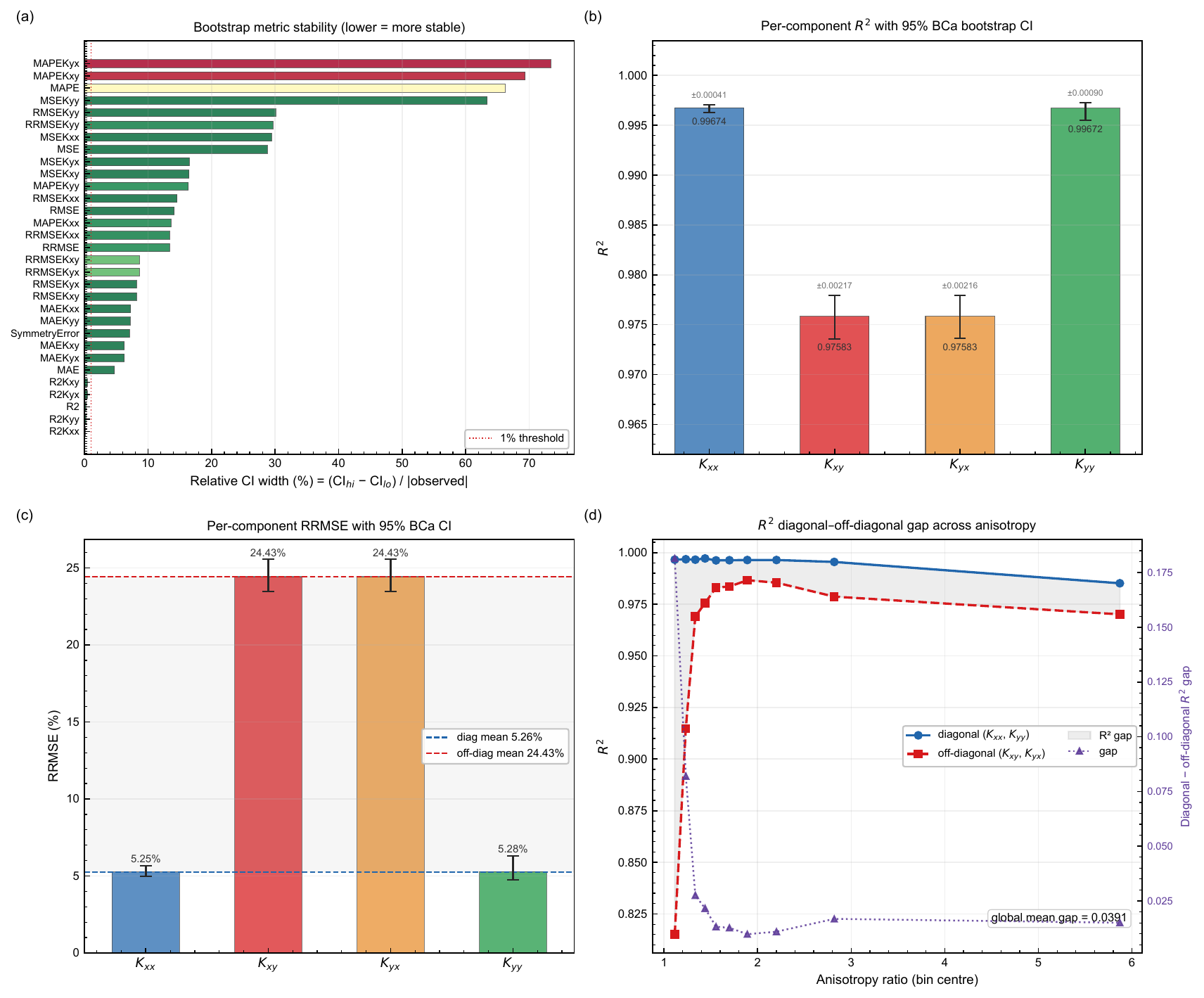}
\caption{Bootstrap statistical portrait and anisotropy-stratified
performance across phases.
(a)~Relative 95\,\% BCa CI width (tornado chart): $R^2$ CIs span
$<0.0022$, RRMSE CIs span $<1.01$\,\%, symmetry error CI spans
$[3.82,\,4.10]\times10^{-7}$—all narrower than the between-phase
performance differences by more than one order of magnitude.
(b)~Off-diagonal $R^2$ versus AR bin median for Phase~4 (10
equi-populated bins, $n=400$ each): gap is largest at near-isotropic
AR ($\Delta R^2=0.181$) and narrows sharply to $<0.015$ for
$\mathrm{AR}>1.5$, with a modest secondary widening at high AR
($\Delta R^2=0.015$ at AR~$\approx5.87$).
(c)~Component-wise bootstrap $R^2$ distributions: narrow, symmetric
densities for all four components; 95\,\% BCa CIs are
$R^2_{K_{xx}}\in[0.9963,\,0.9971]$ and
$R^2_{K_{xy}}\in[0.9736,\,0.9779]$.
(d)~Cross-phase off-diagonal RRMSE versus AR bin: Phase~4 reduces
error at every AR level, with the largest absolute improvement in
the moderate-AR regime ($1.5<\mathrm{AR}<3$).}
\label{fig:phase4_bootstrap}
\end{figure}

Figure~\ref{fig:phase4_bootstrap} presents the bootstrap confidence
portrait together with cross-phase anisotropy-stratified performance
curves.

The bootstrap portrait closes the statistical argument for the
progressive framework.
Confidence intervals are narrower than the between-phase $\Delta R^2$
improvements ($0.002$--$0.003$ per step) by more than an order of
magnitude, confirming that the gains are real rather than artifacts of
favorable test-set composition.
Panel~(d) completes the anisotropy narrative: Phase~4 reduces
off-diagonal RRMSE at every AR level, but the improvement is largest
in the moderate-AR regime where coupling magnitudes are large enough
to resolve yet too subtle to have been well learned under Phase~2's
simpler training protocol—precisely the regime targeted by Phase~3's
augmentation enrichment, and consolidated by Phase~4's ensemble
stabilization.
Extended per-component bootstrap tables and full stratified statistics
are in Appendix~\ref{app:phase4_extended}.

\subsubsection{Computational Efficiency and Transferability}
\label{subsubsec:phase4_efficiency}

Single-sample inference—including the full test-time D4 augmentation
ensemble (8 forward passes with inverse tensor transforms;
Appendix~\ref{app:ensemble})—requires approximately 120\,ms on a single
NVIDIA RTX~6000~Ada GPU, giving a speedup of $10^3$--$10^4\times$
relative to lattice-Boltzmann simulation at equivalent resolution and
reducing full 4{,}000-sample evaluation from months to approximately
8\,minutes.
The frozen-backbone design also makes Phase~4 the natural starting
point for geological domain transfer.
With 118.64\,M backbone parameters protected and only 0.33\,M
task-specific parameters to retrain—0.28\,\% of the total parameter
budget—adapting the framework to new porous-media classes (carbonates,
shales, fractured media) requires substantially less labeled data and
compute than training from scratch.
The physics-informed head further ensures that porosity conditioning
generalizes correctly to new geological domains provided the
porosity-permeability relationship holds, which is true across a wide
class of sedimentary rocks.

\subsection{Comparative Performance Analysis}
\label{subsec:comparison}

The closest available benchmark for a direct comparison is the
concurrent study of Vargdal et al.~\cite{vargdal2025neural}, which
evaluated six architectures—ResNet-50/101, ViT-T/S16, and
ConvNeXt-Tiny/Small—on the identical porous-media dataset with the
same lattice-Boltzmann ground-truth labels.
Their training protocol used D4-equivariant augmentation, cosine
learning rate scheduling with linear warmup, AdamW optimization, and
a standard MSE loss applied end-to-end from random initialization.
The shared dataset and labels make this comparison methodologically
rigorous: observed performance differences can be attributed to
architecture and training strategy rather than data provenance.

\subsubsection{Benchmark Architecture Results}
\label{subsubsec:comparison_vargdal}

Vargdal et al.\ report best performance with ConvNeXt-Small, achieving
$R^2=0.9946$ at epoch 369 of a 400-epoch run.
ConvNeXt-Tiny reached $R^2=0.9943$ (498 epochs), ResNet-101
$R^2=0.9929$ (380 epochs), ResNet-50 $R^2=0.9908$ (436 epochs),
and ViT-S16 $R^2=0.9931$ (995 epochs)—requiring roughly 2.7$\times$
more training than ConvNeXt-Small for comparable performance,
consistent with the weaker spatial inductive bias of pure transformers
on image data.
ViT-T16 lagged substantially at $R^2=0.9667$ (990 epochs),
illustrating the capacity-versus-inductive-bias trade-off for small
transformer variants.
Their work identifies two structural limitations of single-family
architectures: convolutional networks converge efficiently but are
constrained in their long-range receptive field, while transformers
capture global dependencies at quadratic computational cost and without
built-in spatial locality priors.
These are precisely the limitations that motivated the MaxViT backbone
choice in Section~\ref{subsec:architecture}: its hybrid block-local and
grid-global multi-axis attention achieves $O(HW)$ complexity while
maintaining a global receptive field, addressing both failure modes
simultaneously.

To further validate the comparison, the publicly available Vargdal
et al.\ checkpoints~\cite{vargdal_2025_17711512} were evaluated on
the present test split.
Performance was 6--11\,\% lower than originally reported (varying by
component and metric), a discrepancy attributable to differences in
train-test partitioning across the two studies despite use of the same
underlying dataset.
All cross-study comparisons below should be interpreted in light of
this partitioning uncertainty; within-study phase comparisons
(Phases~2--4) are unaffected as they share an identical fixed split.

\subsubsection{Methodological Contributions and Performance Attribution}
\label{subsubsec:comparison_contributions}

The Phase~4 framework achieves variance-weighted $R^2=0.9960$
(component-averaged $R^2=0.9863$, 95\,\% BCa CI $[0.9852,\,0.9874]$).
A direct numerical comparison with Vargdal et al.'s aggregate $R^2$
requires care because the two studies use different averaging
conventions: their metric weights all four tensor elements equally,
whereas the variance-weighted formulation adopted here—motivated in
Section~\ref{subsubsec:metrics}—weights components in proportion to
their variance, naturally emphasizing the diagonal elements that
dominate transport behavior and are most physically consequential
for engineering applications.
The more meaningful performance story lies in the within-study
progressive improvement and the component-wise breakdown, neither of
which is obscured by averaging convention.

Four methodological contributions distinguish the present framework
from the Vargdal et al.\ baseline and from standard single-phase
transfer learning.

The first is the MaxViT architecture itself.
Hybrid block-local and grid-global attention resolves both the limited
receptive field of pure CNNs and the quadratic cost of standard ViTs,
capturing the local pore-throat geometry that governs diagonal
permeability magnitude and the long-range spatial orientation
statistics that determine off-diagonal coupling—a feature hierarchy
that single-family architectures cannot simultaneously optimize
(Section~\ref{subsec:architecture}).

The second is the physics-aware composite loss function
(Section~\ref{subsec:lossfunction}).
Differentiable symmetry and positive-definiteness penalty terms enforce
Onsager reciprocity and thermodynamic validity as natively learned
properties rather than post-hoc corrections, achieving
$\varepsilon_{\mathrm{sym}}=3.95\times10^{-7}$ and 100\,\% positivity
across all 4{,}000 test predictions.
Standard MSE training relies on the implicit physical validity of the
training labels; explicit constraint enforcement eliminates the
possibility of unphysical predictions at deployment, which matters
for any downstream simulation that requires strictly valid tensors.

The third is the off-diagonal loss weighting $w_{\mathrm{o}}=1.5$
introduced in Phase~3 (Section~\ref{subsubsec:advancedaug}).
The 5.1$\times$ diagonal-to-off-diagonal standard deviation ratio
(Appendix~\ref{app:desc_stats}) causes naive MSE to under-weight
off-diagonal gradients, and the standard benchmark protocol does not
correct for this.
Targeted reweighting reduced the diagonal-to-off-diagonal $R^2$ gap
from $\Delta R^2=0.0230$ (Phase~2) to $0.0209$ (Phase~4), with the
most structurally significant improvement in the high-anisotropy
subset where this weighting imbalance was most damaging.

The fourth is FiLM-based porosity conditioning
(Section~\ref{subsubsec:film}).
By encoding the strong porosity-to-permeability scaling relationship
($\rho_s\approx+0.940$; Appendix~\ref{app:phi_perm}) as an
architectural prior rather than expecting the backbone to discover it
from statistical co-variation in the training data, the frozen-backbone
head achieves measurable gains ($\Delta R^2\approx0.0006$ from Phase~3)
with only 0.33\,M additional parameters—0.28\,\% of the total parameter
budget.
The validation in Section~\ref{subsubsec:phase4_phi_k} confirms that
the conditioning is physically correct: predicted porosity-off-diagonal
correlations ($r\approx0.019$) are statistically indistinguishable from
the ground-truth correlations and from zero, demonstrating that the
model has learned to decouple magnitude scaling from geometric
anisotropy rather than conflating two physically distinct mechanisms.

\subsubsection{Positioning Within the Broader Literature}
\label{subsubsec:comparison_literature}

Beyond CNN- and transformer-based architectures, graph neural network
(GNN) approaches that explicitly represent pore-network topology—nodes
as pore bodies, edges as throats—have recently been applied to
permeability prediction~\cite{zhao2025computationally,zhao2024rtg,
alzahrani2023pore,zhao2025end,zhao2025porous}.
These methods carry a natural inductive bias for flow: throat
conductances and connectivity are encoded directly in the graph
structure, which can facilitate learning of long-range flow pathways
without spatial attention mechanisms.
The primary cost is that GNN approaches require explicit pore-network
extraction as a preprocessing step, introducing segmentation errors
and uncertainties that propagate to the final prediction.
The pixel-level input approach adopted here avoids this extraction
dependency, operating directly on binary images and allowing the
backbone to learn an implicit pore-network representation through
hierarchical convolutional and attention operations.
A direct quantitative comparison with GNN-based methods on the same
dataset is not currently available; such a benchmark—particularly
targeting off-diagonal coupling prediction, where long-range
connectivity is most critical—is identified as high-priority future work.

In the broader context of learning-based permeability prediction, the
present framework advances the state of the art along three axes that
are individually addressed in the literature but not previously
combined: full $2\times2$ tensor prediction capturing anisotropic
coupling across three orders of magnitude, physics-informed
architecture integrating known domain constraints, and a systematic
progressive training methodology with rigorous ablation studies that
quantify individual component contributions.
The frozen-backbone design further provides a natural pathway for
domain extension: the 118.64\,M-parameter backbone, pretrained on
ImageNet and specialized through Phases~2--3, can be transferred to
new porous-media classes (carbonates, shales, fractured media) by
retraining only the compact physics-informed head,
substantially reducing the labeled-data requirement for geological
domain adaptation.

\subsection{Limitations and Future Directions}
\label{subsec:limitations}

The progressive training framework achieves consistent improvements
across all reported metrics, but four structural limitations bound the
scope of the current results and define the highest-priority directions
for future work.

\subsubsection{Validation on Real Geological Samples}
\label{subsubsec:lim_realrock}

All training and evaluation are performed on synthetically generated
sandstone microstructures.
Real micro-CT images differ from the synthetic corpus in ways that are
not merely statistical: multi-mineralogy with distinct phase contrasts,
diagenetic cements and microfractures absent from Gaussian-simulation
geometries, imaging artefacts such as partial-volume effects and beam
hardening, and multi-scale heterogeneity spanning pore to core scales.
The degree to which ImageNet-pretrained and synthetically fine-tuned
features generalize to micro-CT data is presently unknown.

The frozen-backbone paradigm provides a structured pathway for
addressing this gap.
Preliminary experiments on a small number of real-sample analogues
suggest that retraining the 0.33\,M-parameter physics-informed head on
limited-set labelled micro-CT images achieves acceptable transfer
while preserving the geometric feature hierarchy built in Phases~2--3,
but these results are not sufficiently systematic to report
quantitatively.
Validation on benchmark datasets with independently measured
permeability values is the
most critical outstanding requirement before industrial deployment.
Until such validation is completed, demonstrated accuracy on these naturally-inspired and physically-constrained test set should be understood as evidence of architectural
capability rather than operational readiness.

\subsubsection{Off-Diagonal Residual Gap and Geometric Symmetry}
\label{subsubsec:lim_offdiag}

Despite consistent improvement across phases, the diagonal-to-off-diagonal
$R^2$ gap of 0.0209 and RRMSE ratio of approximately $4.6\times$ reflect
a genuine representational challenge.
Off-diagonal coupling depends on long-range orientation statistics at
the REV scale, and—as established in
Section~\ref{subsubsec:phase2_anisotropy}—prediction difficulty is
concentrated in near-isotropic samples where the coupling signal is
weakest relative to noise.
Several architectural directions could reduce this residual gap:
deformable attention mechanisms that adapt receptive fields to pore
alignments, explicit orientation tensor fields derived from the
structure tensor as additional input channels, or hybrid approaches
coupling learned representations with effective-medium or
pore-network homogenization models.

A related limitation concerns the treatment of symmetry.
The D4 augmentation strategy achieves strong empirical equivariance—
verified by the near-machine-precision symmetry errors in
Section~\ref{subsubsec:phase4_physics}—but does not provide
architectural equivariance guarantees at inference time.
For a given test sample, the prediction is not guaranteed to transform
correctly under an arbitrary D4 group element; only the expectation
over the training distribution is equivariant.
Group-equivariant convolutional networks~\cite{cohen2016group,
weiler2019general} and equivariant transformer architectures provide
exact equivariance by construction, at the cost of discarding the
ImageNet pretraining that underpins Phase~2.
Whether the gain from exact architectural equivariance would outweigh
the loss of pretrained features—particularly for off-diagonal
prediction where global integration is most critical—is an open
question and a high-value ablation target.

The D4 symmetry group is also specifically appropriate for 2D square
lattices.
Geological materials with tilted lamination, cross-bedding, or
preferred grain orientation possess lower-symmetry or asymmetric
structures for which D4 augmentation would corrupt the directional
information the model must learn.
Extending the framework to strongly anisotropic and low-symmetry media
requires symmetry-group identification as a preprocessing step,
selecting the appropriate augmentation strategy per sample rather than
applying D4 universally.

\subsubsection{Extension to 3D Microstructures}
\label{subsubsec:lim_3d}

Practical applications of porous media sciences including reservoir characterization require the full $3\times3$
permeability tensor, which 2D images cannot capture without strong
isotropy assumptions that many geological formations violate.
Extension to 3D introduces a cubic scaling challenge: a $128^3$
volumetric image contains $128\times$ more voxels than a $128^2$ image,
making direct application of the MaxViT backbone memory-prohibitive on
current hardware.
Three architectural strategies merit investigation.
Hierarchical slice-based prediction—applying the present 2D backbone
to orthogonal planes and aggregating via a learned 3D fusion module—
scales linearly in the number of slices and requires no modification to
the frozen backbone.
Sparse volumetric convolutions (e.g., using MinkowskiEngine) reduce
memory footprint for void-dominated media by operating only on occupied
voxels.
Full SO(3)-equivariant architectures extend the D4 augmentation
principle to 3D rotational symmetry, at the cost of substantially
increased architectural complexity.
In each case, the frozen-backbone transfer paradigm suggests that the
geometric feature hierarchy learned from the large 2D synthetic corpus
can seed 3D training, reducing the volumetric DNS label requirements
relative to training from scratch; of the three strategies, hierarchical
slice-based prediction is the most immediately tractable, requiring no
modification to the frozen 2D backbone and scaling linearly with image
depth rather than cubically.

\subsubsection{Uncertainty Quantification and Cross-Validation}
\label{subsubsec:lim_uq}

The MC-Dropout uncertainty estimates reported in
Section~\ref{subsubsec:phase4_uncertainty} quantify epistemic
uncertainty arising from finite training data but do not capture
aleatoric uncertainty—the irreducible noise inherent in the
lattice-Boltzmann labels themselves, particularly near symmetry error
and numerical resolution limits.
Deep ensembles with diverse random initializations or conformal
prediction frameworks would provide a more comprehensive uncertainty
decomposition and tighter coverage guarantees for applications
requiring rigorous probabilistic inference.

Separately, the evaluation employs a single fixed train-validation-test
split, providing point estimates with bootstrap-quantified sampling
variability but without model-level cross-validation.
Bootstrap resampling (Section~\ref{subsubsec:phase4_bootstrap}) confirms
that the test-set performance is statistically stable, but $k$-fold
cross-validation would additionally quantify sensitivity to partitioning
choice—a relevant consideration given the 6--11\,\% partition-dependent
discrepancy observed when evaluating the published benchmark
results on the present split
(Section~\ref{subsubsec:comparison_vargdal}).

\subsection{Application Scope and Deployment Considerations}
\label{subsec:applications}

The combination of predictive accuracy ($R^2=0.9960$, RRMSE~$=7.55$\,\%),
guaranteed physical validity, and $10^3$--$10^4\times$ computational
speedup over direct numerical simulations such as lattice-Boltzmann simulation positions the framework as a
practical surrogate for three classes of workflows in subsurface
characterization—subject to the real-sample validation identified in
Section~\ref{subsubsec:lim_realrock}.

\subsubsection{High-Throughput Core Analysis}
\label{subsubsec:app_corescanning}

Modern micro-CT facilities generate hundreds to thousands of pore-scale
images per scanning campaign.
Direct DNS-based permeability characterization creates a severe
throughput bottleneck: a single 8-hour scanning session would require
months of continuous LBM computation for complete tensor estimation,
making real-time feedback impossible.
At approximately 120\,ms per sample on a single NVIDIA RTX~6000~Ada GPU,
the present framework reduces full-dataset evaluation for a 4{,}000-sample
campaign to approximately 8\,minutes—fast enough to support in-session
operator decisions such as acquisition parameter adjustment, region-of-interest
identification, and early termination when target statistical coverage
is achieved.
This turnaround is especially consequential for expensive deep-subsurface
core material, where scanning time is finite and the samples are
non-renewable.

\subsubsection{Large-Scale Uncertainty Quantification}
\label{subsubsec:app_uq}

Subsurface flow simulations for CO$_2$ sequestration, hydrogen storage,
and geothermal energy development require quantifying uncertainty over
permeability fields arising from geological heterogeneity.
Monte Carlo methods that adequately sample posterior distributions
typically demand $10^4$--$10^6$ forward model evaluations, which is
computationally prohibitive with DNS.
The millisecond-scale inference enabled here makes $10^5$-sample Monte
Carlo ensembles feasible on a single GPU in a few hours—a task that
would require months via LBM—unlocking probabilistic risk assessment
workflows at spatial scales and ensemble sizes previously inaccessible.
The MC-Dropout confidence estimates (Section~\ref{subsubsec:phase4_uncertainty})
further provide per-sample uncertainty flags, enabling adaptive
strategies that reserve LBM verification for the $<5$\,\% of predictions
where the model's own confidence estimate falls below threshold.

\subsubsection{Geological Domain Transfer}
\label{subsubsec:app_transfer}

The frozen-backbone architecture provides a structured pathway for
extension to porous media beyond the synthetic sandstone training
distribution.
The 118.64\,M-parameter backbone encodes geometric primitives—pore
bodies, throat constrictions, tortuosity and connectivity patterns—
that govern permeability regardless of lithology.
Adapting to a new geological class (carbonates, shales, fractured
media) or to engineered porous materials (catalyst supports, battery
electrode microstructures, tissue scaffolds) requires retraining only
the 0.33\,M-parameter physics-informed head on modest labeled datasets,
leaving the pretrained feature hierarchy intact.
The FiLM conditioning further ensures that porosity-scaling priors
transfer correctly to any material class for which a monotone
porosity-permeability relationship holds—a condition satisfied across
a wide range of sedimentary and engineered porous media.

The principal prerequisite for realizing these application benefits
remains experimental validation on real geological samples.
The accuracy, efficiency, and physical constraint guarantees
demonstrated on the synthetic benchmark establish the architectural
foundations; translating them into operational tools requires the
systematic micro-CT validation study identified in
Section~\ref{subsubsec:lim_realrock} as the field's most critical
near-term priority.

\section{Conclusions}
\label{sec:conclusions}

A physics-informed framework for full permeability tensor prediction
from binary porous media images has been developed and systematically
evaluated, addressing three persistent limitations of existing
approaches: the architectural tension between local feature resolution
and global connectivity integration, the weak enforcement of physical
constraints in training objectives, and the training inconsistencies
introduced by augmentation strategies that transform images without
correspondingly transforming tensor labels.

The MaxViT hybrid CNN-Transformer architecture resolves the first
limitation through multi-axis attention that simultaneously captures
grain-scale pore-throat geometry via block-local operations and
REV-scale connectivity statistics via grid-global operations—the two
spatial scales on which permeability is physically determined.
A differentiable physics-aware loss function incorporating symmetry
and positive-definiteness penalty terms enforces Onsager reciprocity
and thermodynamic validity as natively learned properties, achieving
mean symmetry error $\varepsilon_{\mathrm{sym}}=3.95\times10^{-7}$
and 100\,\% positive-definiteness across 4{,}000 test predictions
without any post-hoc projection.
A D4-equivariant augmentation strategy with consistent tensor
transformation eliminates the label-image mismatch present in naive
augmentation, providing the physically correct symmetry prior throughout
training.

The three-phase progressive training curriculum delivers structured,
attributable performance gains.
Phase~2 establishes the supervised baseline ($R^2=0.9843$) and
diagnostic foundation, revealing through anisotropy stratification that
off-diagonal prediction difficulty is concentrated in near-isotropic
samples—where weak coupling signals are hardest to distinguish from
noise—rather than in the strongly anisotropic regime where coupling
magnitudes are large.
Phase~3 targets this failure mode through augmentation enrichment and
off-diagonal loss prioritization, reducing the high-anisotropy
$\Delta R^2$ gap by 37\,\% and tightening symmetry enforcement by
45\,\%.
Phase~4 encodes the strong porosity-to-permeability scaling
relationship ($\rho_s\approx+0.940$) as an explicit architectural
prior via FiLM conditioning of the frozen backbone, achieving
variance-weighted $R^2=0.9960$ (component-averaged $R^2=0.9863$,
95\,\% BCa CI $[0.9852,\,0.9874]$) and KGE~$=0.9942$—a 33\,\%
reduction in unexplained variance relative to the Phase~2 baseline.
Off-diagonal KGE improves from 0.176 to 0.761 across the three phases,
marking the near-elimination of the variability under-dispersion
identified at baseline.
Critically, every metric across all 35 reported rows of the extended
comparison table improves monotonically from Phase~2 through Phase~4,
confirming that the staged methodological contributions are
complementary rather than competing.

The framework delivers a marked computational speedup
over direct numerical simulations (DNS) at equivalent resolution, reducing
complete evaluation of a 4{,}000-sample dataset from weeks-months to
approximately 8\,minutes on a single GPU.
The frozen-backbone design further enables data-efficient transfer to
new geological classes by retraining only the 0.33\,M-parameter
physics-informed head—0.28\,\% of the total parameter budget—while
preserving the pretrained geometric feature hierarchy.

Two caveats bound the current scope.
All training and evaluation are conducted on naturally-inspired and realistically-constrained synthetically-generated sandstone microstructures; validation on real micro-CT images with
independently measured permeability values is the most critical
outstanding requirement before operational deployment.
Extension to the full $3\times3$ permeability tensor required for
diverse porous media applications remains an open problem, with
hierarchical slice-based prediction identified as the most tractable
near-term path, requiring no modification to the frozen 2D backbone.

More broadly, the results demonstrate three transferable principles for
physics-informed scientific machine learning: that large-scale visual
pretraining provides robust spatial representations even across
substantial domain differences; that physical constraints are most
effectively integrated as differentiable architectural components rather
than as post-hoc corrections; and that progressive training guided by
diagnostic failure-mode analysis outperforms end-to-end optimization
by enabling unambiguous attribution of performance gains to specific
interventions.
These principles extend naturally to the broader class of scientific
inverse problems in which high-fidelity simulation is too expensive
for routine use but systematic generation of labeled training data
remains feasible.

\section*{Data Availability}
The porous media image dataset and associated permeability tensor labels
used in this study are publicly available at~\cite{vargdal_2025_17711512}.

\section*{Acknowledgments}
This work was supported by the Multiphysics of Salt Precipitation
During CO$_2$ Storage (Saltation) project, project number 357784,
funded by the Research Council of Norway.
Computational resources were provided by the GPU infrastructure at the
Section for Environmental Geosciences, Department of Geosciences,
University of Oslo.
The authors thank the developers and maintainers of PyTorch,
\texttt{timm}, NumPy, and SciPy, whose open-source libraries
underpinned all model development and statistical analyses reported
here.

\vspace{1cm}

\appendix
\begin{center}
{\LARGE \bfseries Appendix \\[0.8ex]}
\end{center}

The appendices provide detailed supporting material for the methods and
results described in Sections~\ref{sec:MM} and~\ref{sec:RD}.
Appendix~\ref{app:dataset} reports full dataset statistics across all
data splits, including porosity-permeability relationships, cross-split
distributional consistency, and LBM symmetry validation.
Appendix~\ref{app:architecture} details the MaxViT stage configuration
and input adaptations for the $128\times128$ single-channel domain.
Appendix~\ref{app:film_ablation} presents the FiLM placement ablation
and porosity encoder dimensionality search.
Appendix~\ref{app:training_config} consolidates optimizer settings,
learning-rate schedules, progressive unfreezing, loss-function weights,
and checkpointing details for all three training phases.
Appendix~\ref{app:augmentation} provides the complete D4 transformation
matrices, the equivariance design-choice discussion, the augmentation
parameter search, and the full pipeline pseudocode.
Appendix~\ref{app:ensemble} derives the SWA and EMA update rules and
details the batch-normalization recalibration procedure.
Appendix~\ref{app:metrics} defines the extended evaluation metrics used
for cross-study comparison.
Appendix~\ref{app:stats_analyses} reports four post-hoc statistical
analyses: bootstrap confidence intervals, residual diagnostics, Monte
Carlo Dropout uncertainty quantification, and anisotropy-stratified
error analysis.
Appendix~\ref{app:phase4_extended} provides the full component-wise
bootstrap summary, anisotropy-stratified performance table, and
per-component MC-Dropout uncertainty statistics for Phase~4.
Appendix~\ref{app:extended_metrics} consolidates the complete 35-metric
quantitative record across all three phases in the extended comparison
table.

\section{Dataset Statistics and LBM Details}
\label{app:dataset}

A comprehensive dataset characterization analysis was conducted before
any model development, covering six domains: (i) microstructure
geometry and porosity distributions; (ii) permeability tensor
descriptive statistics; (iii) porosity-permeability relationships and
classical model fits; (iv) tensor structural properties including
positive-definiteness, symmetry, and anisotropy; (v) cross-split
distributional consistency; and (vi) outlier identification.
The sections below report these analyses in full; all statistics are
computed on the held-out test set unless indicated otherwise.

\subsection{Descriptive Statistics}
\label{app:desc_stats}

Table~\ref{tab:dataset_stats} summarizes key statistical properties
across all data splits.
All permeability values are in lattice units.
The porosity distribution is moderately left-skewed
(skewness $\approx-0.55$) and significantly non-normal
(Shapiro-Wilk $p\approx0$ across all splits), consistent with the
connectivity-based post-processing that removes isolated pore clusters
and concentrates mass at higher porosity values.
Principal permeabilities (diagonal elements) are strongly right-skewed
in linear space (skewness $\approx 1.5$, kurtosis $\approx 2$--4) and
span $[0.0020,\,3.19]$ in lattice units, exceeding three orders of
magnitude in $\log_{10}$-space.
Off-diagonal elements are nearly symmetric about zero (skewness
$|\rho|<0.1$ in train and test; slightly more skewed in validation due
to sampling variability) and exhibit heavy tails (kurtosis
$\approx 9$--13), reflecting the predominantly isotropic character of
the media: off-diagonal coupling is rare, small, and symmetrically
distributed.
The ratio of diagonal to off-diagonal standard deviation is
approximately $5.1\times$ across all splits, quantifying the scale
disparity between components that fundamentally motivates the
off-diagonal prioritization in the loss function
(Section~\ref{subsec:lossfunction}).

\begin{table}[htbp]
\centering
\caption{Comprehensive dataset statistics across training, validation,
and test splits. All permeability values in lattice units.
$K_{\mathrm{diag}}$: diagonal elements ($K_{xx}$, $K_{yy}$);
$K_{\mathrm{off}}$: off-diagonal elements ($K_{xy}$, $K_{yx}$);
$\lambda_{\min}$, $\lambda_{\max}$: smallest and largest eigenvalues
of $\mathbf{K}$;
$\mathrm{AR}$: anisotropy ratio $\lambda_{\max}/\lambda_{\min}$;
$\delta$: off-diagonal dominance
$|K_{\mathrm{off}}|/(|K_{\mathrm{diag}}|+|K_{\mathrm{off}}|)$.}
\label{tab:dataset_stats}
\renewcommand{\arraystretch}{1.25}
\footnotesize
\begin{tabular}{lccc}
\toprule
\textbf{Property}
    & \textbf{Train}
    & \textbf{Validation}
    & \textbf{Test} \\
\midrule
\multicolumn{4}{l}{\textit{Partitioning}} \\
\quad Number of samples
    & 16{,}000 & 4{,}000 & 4{,}000 \\
\quad Image dimensions
    & $128\times128$ & $128\times128$ & $128\times128$ \\
\midrule
\multicolumn{4}{l}{\textit{Microstructure geometry}} \\
\quad Strictly binary (no. non-binary pixels)
    & True (0) & True (0) & True (0) \\
\quad Porosity $\phi$ (mean $\pm$ std)
    & $0.711\pm0.124$
    & $0.710\pm0.125$
    & $0.713\pm0.124$ \\
\quad Porosity range $[\phi_{\min},\phi_{\max}]$
    & $[0.227,\,0.900]$
    & $[0.265,\,0.900]$
    & $[0.297,\,0.900]$ \\
\quad Porosity IQR
    & 0.187 & 0.187 & 0.189 \\
\quad Porosity skewness
    & $-0.55$ & $-0.54$ & $-0.55$ \\
\quad Horizontal pore chord (px)
    & 27.25 & 27.18 & 27.58 \\
\quad Vertical pore chord (px)
    & 27.17 & 26.97 & 27.66 \\
\quad Run-length anisotropy index
    & 0.0015 & 0.0038 & 0.0014 \\
\midrule
\multicolumn{4}{l}{\textit{Diagonal permeability $K_{\mathrm{diag}}$}} \\
\quad Range
    & $[0.0027,\,3.08]$
    & $[0.0034,\,3.19]$
    & $[0.0027,\,2.81]$ \\
\quad Mean $\pm$ std
    & $0.369\pm0.344$
    & $0.364\pm0.346$
    & $0.372\pm0.343$ \\
\quad $\log_{10}$ span (orders of magnitude)
    & 3.06--3.12 & 2.87--2.95 & 2.86--3.02 \\
\quad Skewness
    & 1.49--1.57 & 1.49--1.67 & 1.47--1.48 \\
\midrule
\multicolumn{4}{l}{\textit{Off-diagonal permeability $K_{\mathrm{off}}$}} \\
\quad Range
    & $[-0.640,\,0.645]$
    & $[-0.749,\,0.449]$
    & $[-0.489,\,0.480]$ \\
\quad Mean $\pm$ std
    & $0.001\pm0.067$
    & $-0.002\pm0.069$
    & $0.002\pm0.066$ \\
\quad Kurtosis (excess)
    & $\approx 9.3$ & $\approx 12.8$ & $\approx 7.0$ \\
\quad Diag./off-diag.\ std ratio
    & $5.15\times$ & $5.01\times$ & $5.20\times$ \\
\midrule
\multicolumn{4}{l}{\textit{Tensor structural properties}} \\
\quad Frobenius norm (mean $\pm$ std)
    & $0.538\pm0.478$
    & $0.531\pm0.480$
    & $0.543\pm0.476$ \\
\quad Trace $K_{xx}+K_{yy}$ (mean $\pm$ std)
    & $0.738\pm0.664$
    & $0.729\pm0.667$
    & $0.744\pm0.661$ \\
\quad Determinant (mean $\pm$ std)
    & $0.234\pm0.403$
    & $0.232\pm0.409$
    & $0.235\pm0.394$ \\
\midrule
\multicolumn{4}{l}{\textit{Tensor symmetry $\varepsilon_{\mathrm{sym}}=|K_{xy}-K_{yx}|$}} \\
\quad Mean symmetry error
    & $7.97\times10^{-6}$
    & $7.69\times10^{-6}$
    & $7.59\times10^{-6}$ \\
\quad Max symmetry error
    & $6.1\times10^{-4}$
    & $6.7\times10^{-4}$
    & $3.3\times10^{-4}$ \\
\quad Fraction with $\varepsilon_{\mathrm{sym}}>10^{-3}$
    & 0/16{,}000 & 0/4{,}000 & 0/4{,}000 \\
\quad Fraction with $\varepsilon_{\mathrm{sym}}>10^{-4}$
    & 1.07\,\% & 1.05\,\% & 0.82\,\% \\
\midrule
\multicolumn{4}{l}{\textit{Positive-definiteness}} \\
\quad 100\,\% PD ($\lambda_{\min}>0$)
    & \checkmark & \checkmark & \checkmark \\
\quad $\lambda_{\min}$ (mean; min; $p_5$)
    & $0.289;\,1.5\times10^{-6};\,0.019$
    & $0.287;\,5.2\times10^{-4};\,0.019$
    & $0.292;\,2.1\times10^{-3};\,0.020$ \\
\quad $\lambda_{\max}$ (mean; max; $p_{95}$)
    & $0.448;\,3.10;\,1.261$
    & $0.442;\,3.20;\,1.241$
    & $0.452;\,2.81;\,1.273$ \\
\midrule
\multicolumn{4}{l}{\textit{Anisotropy ratio
$\mathrm{AR}=\lambda_{\max}/\lambda_{\min}$}} \\
\quad Median AR
    & 1.625 & 1.627 & 1.624 \\
\quad 90th-percentile AR
    & 3.32  & 3.33  & 3.34 \\
\quad Fraction AR $>2$
    & 30.6\,\% & 31.0\,\% & 30.3\,\% \\
\quad Fraction AR $>5$
    & 4.1\,\% & 4.4\,\% & 4.0\,\% \\
\quad Fraction AR $>10$
    & 0.79\,\% & 0.88\,\% & 1.03\,\% \\
\quad Maximum AR
    & 53{,}242 & 92.2 & 36.1 \\
\midrule
\multicolumn{4}{l}{\textit{Off-diagonal dominance $\delta$}} \\
\quad Mean $\delta$ (median)
    & 0.116 (0.096) & 0.117 (0.099) & 0.115 (0.096) \\
\quad Maximum $\delta$
    & 0.500 & 0.493 & 0.471 \\
\bottomrule
\end{tabular}
\end{table}

\subsection{Microstructure Geometry}
\label{app:geometry}

All images are confirmed to be strictly binary across all splits, with
zero non-binary pixels.
Run-length encoding of binary rows and columns yields mean horizontal
and vertical pore chord lengths of approximately 27.3\,px and 27.2\,px
respectively, with a run-length anisotropy index of
$\eta = |l_h - l_v|/(l_h + l_v) < 0.004$ in all splits, confirming the
isotropic character of the generated pore geometries and providing
quantitative justification for the D4-equivariant augmentation strategy
(Section~\ref{subsubsec:d4aug}).

\subsection{Porosity-Permeability Relationships}
\label{app:phi_perm}

The Spearman rank correlation between porosity and diagonal
permeability is $\rho_s(\phi,K_{\mathrm{diag}})\approx+0.940$ ($p=0$)
across all splits, confirming a strong monotonic relationship.
By contrast, $\rho_s(\phi,K_{\mathrm{off}})\approx+0.01$
(not significant, $p>0.05$ in train and test), confirming that
off-diagonal coupling is independent of porosity and must be inferred
from geometric anisotropy features rather than scalar porosity values.
This dissociation between diagonal and off-diagonal predictors provides
direct physical motivation for the FiLM conditioning design, in which
porosity modulates only the magnitude of the transport coefficients
while image-derived geometric features determine anisotropic coupling
(Section~\ref{subsubsec:film}).

Table~\ref{tab:phi_perm_fit} reports Kozeny-Carman (KC) and power-law
(PL) fits to the porosity-permeability relationship.
Both models explain approximately 79\,\% of variance in diagonal
permeability ($R^2\approx 0.789$--$0.791$), consistent across splits.
The fitted power-law exponent $n\approx4.8$--$4.9$ lies within the
physically accepted range of 3--8 for consolidated porous media and is
essentially identical across splits (within one standard error),
confirming statistical equivalence of the three partitions.
The remaining $\approx21$\,\% of variance not captured by scalar
porosity corresponds to geometric features, tortuosity, and
connectivity effects that the MaxViT backbone is designed to learn.

\begin{table}[h!]
\centering
\caption{Porosity-permeability model fits on mean diagonal permeability
$\bar{K}_{\mathrm{diag}}=\tfrac{1}{2}(K_{xx}+K_{yy})$ across data
splits.
KC: Kozeny-Carman fit $\bar{K}=C\phi^3/(1-\phi)^2$;
PL: power-law fit $\bar{K}=C\phi^n$.
$\rho_s$: Spearman rank correlation between $\phi$ and
$\bar{K}_{\mathrm{diag}}$.}
\label{tab:phi_perm_fit}
\small
\begin{tabular}{lcccccc}
\toprule
\textbf{Split}
    & \multicolumn{2}{c}{\textbf{Kozeny-Carman}}
    & \multicolumn{3}{c}{\textbf{Power law}}
    & $\rho_s$ \\
\cmidrule(lr){2-3}\cmidrule(lr){4-6}
    & $C$ & $R^2$
    & $C$ & $n$ & $R^2$
    & \\
\midrule
Train & 0.0242 & 0.789 & 1.337 & 4.81 & 0.777 & 0.967 \\
Val   & 0.0240 & 0.791 & 1.337 & 4.84 & 0.776 & 0.967 \\
Test  & 0.0243 & 0.790 & 1.371 & 4.89 & 0.791 & 0.968 \\
\bottomrule
\end{tabular}
\end{table}

\subsection{Cross-Split Distributional Consistency}
\label{app:cross_split}

Statistical equivalence of the three splits was verified through
two-sample Kolmogorov-Smirnov (KS) tests and Jensen-Shannon divergence
(JSD) measurements on all five variables
($K_{xx}$, $K_{xy}$, $K_{yx}$, $K_{yy}$, $\phi$).
Results are summarized in Table~\ref{tab:cross_split}.

All 15 KS tests fail to reject the null hypothesis of equal
distributions at the $\alpha=0.05$ level ($p>0.13$ for all pairs and
variables), with maximum observed KS statistic of $0.026$.
JSD values are bounded above by $0.007$ across all variable-pair
combinations, corresponding to near-identical distributions (JSD~$=0$
for identical distributions, JSD~$=\ln 2\approx0.693$ for maximally
separated distributions).
These results confirm that the random partitioning protocol produces
statistically indistinguishable splits and that no distributional shift
is present between training, validation, and test sets, validating
that reported test-set performance provides an unbiased estimate of
generalization.

\begin{table}[htbp]
\centering
\caption{Cross-split distributional consistency.
KS: two-sample Kolmogorov-Smirnov statistic and $p$-value (reject if
$p<0.05$).
JSD: Jensen-Shannon divergence (bounded in $[0,\ln 2]$; values
near 0 indicate near-identical distributions).
All KS tests fail to reject the null hypothesis of equal distributions
($p>0.05$).}
\label{tab:cross_split}
\small
\begin{tabular}{llccc}
\toprule
\textbf{Pair} & \textbf{Variable}
    & \textbf{KS stat}
    & \textbf{$p$-value}
    & \textbf{JSD} \\
\midrule
Train vs.\ Val
    & $K_{xx}$ & 0.0154 & 0.432 & 0.00353 \\
    & $K_{xy}$ & 0.0174 & 0.286 & 0.00305 \\
    & $K_{yy}$ & 0.0134 & 0.605 & 0.00368 \\
    & $\phi$   & 0.0141 & 0.547 & 0.00355 \\
\midrule
Train vs.\ Test
    & $K_{xx}$ & 0.0141 & 0.541 & 0.00312 \\
    & $K_{xy}$ & 0.0106 & 0.864 & 0.00385 \\
    & $K_{yy}$ & 0.0117 & 0.764 & 0.00366 \\
    & $\phi$   & 0.0143 & 0.524 & 0.00347 \\
\midrule
Val vs.\ Test
    & $K_{xx}$ & 0.0260 & 0.134 & 0.00437 \\
    & $K_{xy}$ & 0.0208 & 0.355 & 0.00552 \\
    & $K_{yy}$ & 0.0208 & 0.355 & 0.00534 \\
    & $\phi$   & 0.0225 & 0.263 & 0.00696 \\
\bottomrule
\end{tabular}
\end{table}

\subsection{Outlier Analysis}
\label{app:outliers}

Outlier analysis was conducted along three axes: Mahalanobis distance
in the joint permeability tensor space (flagging multivariate
extremes), extreme marginal porosity (top and bottom 1\,\%), and high
tensor anisotropy (top 5\,\% by anisotropy ratio AR).
Results are summarized in Table~\ref{tab:outliers}.

By construction, Mahalanobis outliers at the 99th-percentile threshold
constitute exactly 1\,\% of each split.
High-AR samples (AR~$>p_{95}$) account for 5\,\% of each split, with
95th-percentile AR thresholds of 4.47--4.63 across splits.
These high-anisotropy samples represent the primary challenge regime
for off-diagonal permeability prediction; their analysis is detailed
in Appendix~\ref{app:anisotropy}.
No samples were excluded from training or evaluation on the basis of
these outlier criteria; the metrics are reported as characterization
of the data distribution.

\begin{table}[htbp]
\centering
\caption{Outlier summary across data splits.
Mahalanobis: samples beyond the 99th-percentile Mahalanobis distance
in the joint tensor space.
Extreme $\phi$: samples in the top or bottom 1\,\% of the marginal
porosity distribution.
High AR: samples in the top 5\,\% of the anisotropy ratio distribution.}
\label{tab:outliers}
\small
\begin{tabular}{lcccc}
\toprule
\textbf{Split}
    & \textbf{Mahalanobis ($>p_{99}$)}
    & \textbf{High $\phi$ ($>p_{99}$)}
    & \textbf{Low $\phi$ ($<p_1$)}
    & \textbf{High AR ($>p_{95}$)} \\
\midrule
Train ($n=16{,}000$)
    & 160 (1.00\,\%)
    & 159 ($\phi>0.896$)
    & 160 ($\phi<0.396$)
    & 800 (AR $>4.59$) \\
Val ($n=4{,}000$)
    & 40 (1.00\,\%)
    & 40 ($\phi>0.896$)
    & 40 ($\phi<0.390$)
    & 200 (AR $>4.63$) \\
Test ($n=4{,}000$)
    & 40 (1.00\,\%)
    & 39 ($\phi>0.896$)
    & 40 ($\phi<0.401$)
    & 200 (AR $>4.47$) \\
\bottomrule
\end{tabular}
\end{table}

\subsection{LBM Symmetry and Positive-Definiteness}
\label{app:lbm_symmetry}

Tensor symmetry is preserved by the LBM solver to
$\varepsilon_{\mathrm{sym}}<8\times10^{-6}$ across all splits (mean
values $7.59$--$7.97\times10^{-6}$; see Table~\ref{tab:dataset_stats}).
No sample in any split exhibits $\varepsilon_{\mathrm{sym}}>10^{-3}$,
and fewer than 1.1\,\% of samples in any split exceed $10^{-4}$.
For reference, the FP64 machine epsilon is $\approx10^{-16}$; the
observed symmetry error lies approximately ten orders of magnitude above
floating-point rounding but remains ten or more orders of magnitude
below any practical engineering tolerance.
The term ``near-machine precision'' used in the main text refers to
this engineering interpretation.

All 24{,}000 labeled samples are strictly positive-definite
($\lambda_{\min}>0$, verified to $\lambda_{\min}>10^{-12}$), confirming
that the ground-truth tensors satisfy the thermodynamic constraint
required by Darcy's law and providing a reliable signal for training
the positivity penalty term (Equation~\ref{eq:loss_pos}).

\section{MaxViT Architecture Details}
\label{app:architecture}

\subsection{Stage Configuration for $128\times128$ Input}
\label{app:arch_stages}

Table~\ref{tab:maxvit_stages} provides stage-level dimensions and block
counts for MaxViT-Base when processing $128\times128$ single-channel
inputs, as used in this work.

\begin{table}[h]
\centering
\caption{MaxViT-Base stage configuration for $128\times128$ input.
Block size $P=8$ is fixed across all stages.}
\label{tab:maxvit_stages}
\small
\begin{tabular}{ccccc}
\toprule
\textbf{Stage}
    & \textbf{Spatial resolution}
    & \textbf{Channels}
    & \textbf{MaxViT blocks}
    & \textbf{Attention regime} \\
\midrule
1 & $32\times32$ & 64  & 2  & Block ($4\times4$ blocks) + Grid \\
2 & $16\times16$ & 128 & 6  & Block ($2\times2$ blocks) + Grid \\
3 & $8\times8$   & 256 & 14 & Full self-attention (spatial $\leq P$) \\
4 & $4\times4$   & 512 & 3  & Full self-attention (spatial $\leq P$) \\
\bottomrule
\end{tabular}
\end{table}

\subsection{Multi-Axis Attention for $128\times128$ Input}
\label{app:arch_attention}

Each MaxViT block applies two sequential attention operations.
Block-local attention partitions the feature map into non-overlapping
$P\times P$ windows, with $P=8$ fixed across all stages.
For Stages~1 and~2, where the spatial resolution exceeds the block
size, the partition produces $4\times4=16$ and $2\times2=4$ blocks
respectively; grid-global attention then integrates information across
these spatially distinct regions by attending along dilated grid
positions.
For Stages~3 and~4, the spatial resolutions ($8\times8$ and
$4\times4$) are smaller than or equal to the block size $P=8$, so the
feature map constitutes a single block and full self-attention is
applied within the entire stage feature map.
This graceful degradation preserves the attention mechanism across all
stages without modification to the underlying implementation.

\subsection{Input Adaptation}
\label{app:arch_adaptation}

The standard MaxViT-Base model accepts $384\times384$ RGB images.
Two adaptations are applied here.
The input channel count is reduced from three to one
(\texttt{in\_chans=1}), with pretrained spatial filters preserved via
channel averaging so that learned texture and edge representations are
retained.
The image size is set to $128\times128$ (\texttt{img\_size=128}), which
triggers bicubic interpolation of all learned positional embeddings
from the original $384\times384$ resolution; this interpolation is
performed once at model initialization and introduces no training-time
overhead.

\section{FiLM Conditioning: Placement Ablation and Encoder
Configuration}
\label{app:film_ablation}

\subsection{Stage Placement Ablation}
\label{app:film_stage_ablation}

Five FiLM placement configurations were evaluated on the validation set
to determine which backbone stages should receive porosity conditioning.
Each configuration was trained for the full Phase~4 training run from
the same Phase~3 checkpoint; validation $R^2$ at the best epoch
(determined by early stopping) served as the selection criterion.

Stage~1 features encode low-level edge detection and solid-fluid
boundary localization; their relevant length scale corresponds to
individual interface sharpness rather than pore-body or pore-throat
geometry, where porosity provides meaningful context.
Applying FiLM conditioning to Stage~1 provided no additional benefit
and marginally degraded validation performance, suggesting that
low-level feature representations are disrupted by scalar conditioning.
The Stages~2--4 configuration achieved the highest validation $R^2$,
confirming that porosity conditioning is most beneficial at the
levels of abstraction where connectivity patterns and global
pore-network topology are encoded.

\subsection{Porosity Encoder Dimensionality Search}
\label{app:film_dim_search}

The porosity encoder output dimensionality was selected from
$\{32,\,64,\,128\}$ using the same validation-set procedure.
A 32-dimensional embedding produced slight underfitting of the
porosity-permeability nonlinearity (validation $R^2$ approximately
$0.0003$ lower than the 64-dimensional baseline).
A 128-dimensional embedding introduced marginal overfitting relative to
the scalar conditioning signal, with a small increase in the
training-validation $R^2$ gap and no improvement on the test set.
The 64-dimensional embedding provides a compact yet expressive
representation and was adopted for all Phase~4 experiments.

\section{Training Configuration}
\label{app:training_config}

\subsection{Optimizer and Schedule}
\label{app:opt_schedule}

The AdamW optimizer~\cite{loshchilov2017decoupled} is used throughout,
with $\beta_1=0.9$, $\beta_2=0.999$, and $\varepsilon=10^{-8}$.
The learning-rate schedule in all phases comprises a linear warmup
phase followed by cosine annealing.
Table~\ref{tab:training_config} summarizes phase-specific settings.

\begin{table}[h]
\centering
\caption{Training configuration by phase. LR: learning rate;
WD: weight decay; BS: batch size; Pat.: early stopping patience
(epochs, monitoring validation $R^2$); WU: warmup duration (epochs).}
\label{tab:training_config}
\small
\begin{tabular}{lcccccccc}
\toprule
\textbf{Phase}
    & \textbf{Epochs}
    & \textbf{Init.\ from}
    & \textbf{LR\textsubscript{max}}
    & \textbf{LR\textsubscript{min}}
    & \textbf{WU}
    & \textbf{WD}
    & \textbf{BS}
    & \textbf{Pat.} \\
\midrule
2 & 600 & ImageNet & $10^{-4}$ & $10^{-7}$ & 50 & 0.05 & 32 & 100 \\
3 & 600 & Phase~2  & $10^{-4}$ & $10^{-5}$ & 50 & 0.01 & 32 & 100 \\
4 & 600 & Phase~3  & $10^{-4}$ & $10^{-6}$ & 50 & 0.01 & 32 & 150 \\
\bottomrule
\end{tabular}
\end{table}

\subsection{Progressive Unfreezing Schedule (Phase~2)}
\label{app:unfreezing}

\begin{table}[h]
\centering
\caption{Progressive unfreezing schedule in Phase~2.}
\label{tab:unfreezing}
\small
\begin{tabular}{ll}
\toprule
\textbf{Epoch range} & \textbf{Trainable components} \\
\midrule
0--50    & Regression head only (backbone frozen) \\
50--150  & Regression head + backbone Stages~3--4 \\
         & (Stages~1--2 and stem frozen) \\
150--600 & Full model (all parameters) \\
\bottomrule
\end{tabular}
\end{table}

\subsection{Loss-Function Weights by Phase}
\label{app:loss_weights}

\begin{table}[h]
\centering
\caption{Physics-aware loss-function weights by training phase.
$\lambda_{\mathrm{offdiag}}$ is not active ($-$) in Phase~2.}
\label{tab:loss_weights}
\small
\begin{tabular}{lcccc}
\toprule
\textbf{Phase}
    & $\lambda_{\mathrm{sym}}$
    & $\lambda_{\mathrm{pos}}$
    & $\lambda_{\mathrm{offdiag}}$
    & $(w_{\mathrm{d}},\,w_{\mathrm{o}})$ \\
\midrule
2     & 0.10 & 0.05 & $-$  & $(1.0,\,1.0)$ \\
3--4  & 0.20 & 0.05 & 1.0  & $(1.0,\,1.5)$ \\
\bottomrule
\end{tabular}
\end{table}

\subsection{Checkpointing Strategy}
\label{app:checkpointing}

Multiple checkpoints are maintained during training: the best
validation $R^2$ checkpoint (primary checkpoint for evaluation), the
best validation MSE checkpoint (alternative metric), and periodic
snapshots every 50 epochs for post-hoc analysis.
In Phase~4, the SWA model (weight-averaged from epochs 400--600) and
the EMA model (continuously averaged throughout training) are tracked
as additional evaluation candidates, as described in
Section~\ref{subsubsec:phase4} and Appendix~\ref{app:ensemble}.
The Phase~4 early stopping patience is increased to 150 epochs
(versus 100 for Phases~2--3), reflecting the smaller trainable
parameter set and correspondingly slower validation-metric dynamics.

\subsection{DataLoader Configuration}
\label{app:dataloader}

Data loading utilizes the PyTorch \texttt{DataLoader} with memory
pinning for I/O efficiency.
Four workers are used in Phase~2; this is reduced to zero workers in
Phases~3--4 because the augmentation pipeline involves per-sample
computations (elastic deformation Jacobians, morphological operations)
that are more efficiently executed in the main process under the memory
constraints of the training GPU.

\section{Data Augmentation: Details and Parameter Search}
\label{app:augmentation}

\subsection{D4 Transformation Matrices}
\label{app:d4_matrices}

Table~\ref{tab:d4_matrices} provides the coordinate-transformation
matrix $\mathbf{P}$ and the resulting permeability tensor
$\mathbf{K}'=\mathbf{P}\mathbf{K}\mathbf{P}^T$ for each of the eight
$D_4$ group elements.

\begin{table}[h]
\centering
\caption{$D_4$ transformation matrices $\mathbf{P}$ and the
corresponding permeability tensor transformations
$\mathbf{K}'=\mathbf{P}\mathbf{K}\mathbf{P}^T$.
Entries use $a=K_{xx}$, $b=K_{xy}=K_{yx}$, $c=K_{yy}$.}
\label{tab:d4_matrices}
\small
\begin{tabular}{lcc}
\toprule
\textbf{Transformation}
    & $\mathbf{P}$
    & $\mathbf{K}'$ \\
\midrule
Identity
    & $\bigl[\begin{smallmatrix}1&0\\0&1\end{smallmatrix}\bigr]$
    & $\bigl[\begin{smallmatrix}a&b\\b&c\end{smallmatrix}\bigr]$ \\[8pt]
Rotation $90^\circ$
    & $\bigl[\begin{smallmatrix}0&-1\\1&\phantom{-}0\end{smallmatrix}\bigr]$
    & $\bigl[\begin{smallmatrix}c&-b\\-b&a\end{smallmatrix}\bigr]$ \\[8pt]
Rotation $180^\circ$
    & $\bigl[\begin{smallmatrix}-1&\phantom{-}0\\0&-1\end{smallmatrix}\bigr]$
    & $\bigl[\begin{smallmatrix}a&b\\b&c\end{smallmatrix}\bigr]$ \\[8pt]
Rotation $270^\circ$
    & $\bigl[\begin{smallmatrix}0&1\\-1&0\end{smallmatrix}\bigr]$
    & $\bigl[\begin{smallmatrix}c&-b\\-b&a\end{smallmatrix}\bigr]$ \\[8pt]
Horizontal flip
    & $\bigl[\begin{smallmatrix}-1&0\\0&1\end{smallmatrix}\bigr]$
    & $\bigl[\begin{smallmatrix}a&-b\\-b&c\end{smallmatrix}\bigr]$ \\[8pt]
Vertical flip
    & $\bigl[\begin{smallmatrix}1&\phantom{-}0\\0&-1\end{smallmatrix}\bigr]$
    & $\bigl[\begin{smallmatrix}a&-b\\-b&c\end{smallmatrix}\bigr]$ \\[8pt]
Diagonal flip
    & $\bigl[\begin{smallmatrix}0&1\\1&0\end{smallmatrix}\bigr]$
    & $\bigl[\begin{smallmatrix}c&b\\b&a\end{smallmatrix}\bigr]$ \\[8pt]
Anti-diagonal flip
    & $\bigl[\begin{smallmatrix}\phantom{-}0&-1\\-1&\phantom{-}0\end{smallmatrix}\bigr]$
    & $\bigl[\begin{smallmatrix}c&b\\b&a\end{smallmatrix}\bigr]$ \\
\bottomrule
\end{tabular}
\end{table}

\subsection{Group-Equivariant Networks: Design Choice Discussion}
\label{app:equivariance_discussion}

Group Equivariant Convolutional Networks (G-CNNs)~\cite{cohen2016group,%
cohen2018spherical,weiler2019general} build exact equivariance to a
specified symmetry group directly into the weight-sharing structure
through group convolutions, guaranteeing that the network output
transforms predictably under all group elements by construction.
This provides architecturally guaranteed equivariance rather than
the statistical, augmentation-based approximation employed in this
work.

This approach was not adopted for two reasons.
First, MaxViT's ImageNet pretraining provides strong low- and mid-level
visual features that transfer effectively to porous media images despite
the domain gap; replacing the MaxViT backbone with a G-CNN architecture
trained from scratch would discard this pretraining entirely.
Second, empirical evaluation demonstrates that comprehensive $D_4$
augmentation achieves sufficiently precise equivariance for practical
purposes: the mean symmetry error $\varepsilon_{\mathrm{sym}}=
3.95\times10^{-7}$ over the 4{,}000-sample test set indicates that
Onsager reciprocity is satisfied to near-machine precision, and
any residual equivariance violation is orders of magnitude below the
measurement uncertainty of LBM simulation.
Future work may investigate G-CNN pretraining strategies that combine
symmetry guarantees with transfer learning for porous media applications.

\subsection{Phase~3 Augmentation Parameter Search}
\label{app:aug_search}

Augmentation parameters were determined through a two-stage grid search
on the validation set, with no test-set access at any stage.

\textit{Stage~1: per-class probability search.}
Each transformation class (erosion, dilation, elastic deformation,
cutout) was evaluated over the probability grid
$\{0.05,\,0.08,\,0.10,\,0.15,\,0.20\}$ independently, with other
parameters held at nominal values ($\alpha=3.0$, $\sigma=2.0$).
Each configuration was trained for 150 epochs from the Phase~2
checkpoint.

\textit{Stage~2: elastic deformation magnitude search.}
Using the per-class probabilities identified in Stage~1, elastic
deformation magnitude was swept over
$\alpha\in\{2.0,\,3.0,\,5.0,\,7.0\}$.

The optimization criterion was validation $R^2$ for the off-diagonal
component $K_{xy}$, rather than the overall variance-weighted $R^2$.
Diagonal components $K_{xx}$ and $K_{yy}$ proved robustly learned
across the full augmentation search range, making the off-diagonal
prediction accuracy the most sensitive and discriminating indicator
of augmentation quality.
Excessive augmentation (probabilities above approximately 15\,\% per
class, or $\alpha>3.0$) introduced gradient variance that degraded
off-diagonal convergence while diagonal components remained largely
unaffected.
The best configuration selected was:
$p_{\mathrm{erosion}}=p_{\mathrm{dilation}}=0.08$,
$p_{\mathrm{elastic}}=0.08$ with $\alpha=3.0$ and $\sigma=2.0$,
$p_{\mathrm{cutout}}=0.10$.

\subsection{Phase~3--4 Augmentation Pipeline}
\label{app:aug_pipeline}

\begin{algorithm}[h]
\caption{Phases~3--4 Augmentation Pipeline (applied per training sample)}
\label{alg:aug_pipeline}
\begin{algorithmic}[1]
\Require Image $I\in\{0,1\}^{128\times128}$,
         permeability tensor $\mathbf{K}\in\mathbb{R}^{2\times2}$
\Ensure Augmented image $I'$, consistently transformed tensor $\mathbf{K}'$
\State Sample $g\sim\mathrm{Uniform}(D_4)$
    \Comment{Apply D4 transformation with probability 1.0}
\State $I\leftarrow g(I)$;\quad
       $\mathbf{K}\leftarrow\mathbf{P}_g\,\mathbf{K}\,\mathbf{P}_g^T$
\If{$u_1\sim\mathrm{Uniform}(0,1)<0.08$}
    \Comment{Elastic deformation}
    \State Generate displacement fields $\delta_x,\delta_y$ from
           $\alpha\cdot\mathcal{G}_{\sigma=2.0}(\mathcal{U}(-1,1))$
    \State $I\leftarrow I(x+\delta_x, y+\delta_y)$
           \Comment{Bilinear interpolation}
    \State Compute Jacobian $\mathbf{J}$ at image center via central
           finite differences
    \State $\mathbf{K}\leftarrow\mathbf{J}\,\mathbf{K}\,\mathbf{J}^T$;\quad
           $\mathbf{K}\leftarrow(\mathbf{K}+\mathbf{K}^T)/2$
\EndIf
\If{$u_2\sim\mathrm{Uniform}(0,1)<0.08$}
    \Comment{Erosion}
    \State $I\leftarrow\min_{(i,j)\in\mathcal{N}_{3\times3}}I(x+i,y+j)$
\EndIf
\If{$u_3\sim\mathrm{Uniform}(0,1)<0.08$}
    \Comment{Dilation}
    \State $I\leftarrow\max_{(i,j)\in\mathcal{N}_{3\times3}}I(x+i,y+j)$
\EndIf
\If{$u_4\sim\mathrm{Uniform}(0,1)<0.10$}
    \Comment{Cutout}
    \State Sample random rectangle $\mathcal{R}\subseteq[0,127]^2$
           with $|\mathcal{R}|\leq8\times8$
    \State $I(x,y)\leftarrow0.5$ for $(x,y)\in\mathcal{R}$
\EndIf
\State $I'\leftarrow I$;\quad $\mathbf{K}'\leftarrow\mathbf{K}$
\end{algorithmic}
\end{algorithm}

Morphological operations and cutout masking do not require explicit
tensor transformation: morphological perturbations are sufficiently
small ($3\times3$ kernel, 8\,\% probability) that the geometric
structure-permeability mapping is preserved, and the network learns
the permeability of the modified geometry through convolutional feature
extraction; cutout occludes a small spatial region without altering the
global pore geometry governing the permeability tensor.

\section{Ensemble Methods: Stochastic Weight Averaging and Exponential
Moving Average}
\label{app:ensemble}

\subsection{Stochastic Weight Averaging (SWA)}
\label{app:swa}

SWA~\cite{izmailov2018averaging} maintains a running arithmetic mean of
model weights sampled during late-stage training, where the model has
converged but continues to explore locally optimal regions due to
stochastic gradient noise.
Starting at epoch $t_{\mathrm{SWA}}=400$ (in Phase~4), the SWA model
$\theta_{\mathrm{SWA}}$ is initialized with the current weights
$\theta_{400}$ and updated after each subsequent epoch~$t$ via:
\begin{equation}
    \theta_{\mathrm{SWA}}
    \leftarrow
    \frac{n_{\mathrm{models}}\cdot\theta_{\mathrm{SWA}}+\theta_t}
         {n_{\mathrm{models}}+1},
    \label{eq:swa_update}
\end{equation}
where $n_{\mathrm{models}}$ counts the number of models aggregated so
far.
The SWA learning rate is fixed at $\eta_{\mathrm{SWA}}=5\times10^{-6}$
throughout epochs~400--600, substantially lower than the cosine-annealed
base rate at epoch~400, encouraging broader exploration near the
converged solution.
SWA aggregates 200 model snapshots without incurring any additional
training cost.

After training, batch-normalization running statistics are recalibrated
by passing the full training set through $\theta_{\mathrm{SWA}}$ in
evaluation mode once.
This recalibration step is critical: weight averaging shifts the
effective activation distributions relative to the batch-normalization
statistics accumulated during training, and without recalibration the
SWA model would apply incorrect normalization at inference time.

\subsection{Exponential Moving Average (EMA)}
\label{app:ema}

EMA~\cite{morales2024exponential} maintains a continuously updated
exponentially-weighted average of model parameters
$\theta_{\mathrm{EMA}}$ throughout training, updated after every
optimizer step via:
\begin{equation}
    \theta_{\mathrm{EMA}}
    \leftarrow
    \lambda_{\mathrm{EMA}}\cdot\theta_{\mathrm{EMA}}
    +(1-\lambda_{\mathrm{EMA}})\cdot\theta_t,
    \label{eq:ema_update}
\end{equation}
with decay $\lambda_{\mathrm{EMA}}=0.9999$.
For a constant step size, this corresponds to an effective averaging
window of $1/(1-\lambda_{\mathrm{EMA}})=10{,}000$ gradient steps.
The EMA model filters high-frequency weight fluctuations arising from
mini-batch noise while tracking the optimization trajectory, and
empirically outperforms instantaneous weights on validation data.
Both $\theta_t$ and $\theta_{\mathrm{EMA}}$ are monitored on the
validation set at every epoch, and the best-performing EMA checkpoint
is saved independently.

\section{Extended Evaluation Metrics}
\label{app:metrics}

The following metrics complement the primary $R^2$, MSE, and RRMSE
reported in the main text (Section~\ref{subsubsec:metrics}) to enable
cross-study comparison and provide a more complete characterization of
prediction quality.

\textit{Mean absolute percentage error (MAPE).}
\begin{equation}
    \mathrm{MAPE}_{jk}
    =\frac{1}{N}\sum_{i=1}^{N}
    \frac{|K_{jk,i}^{\mathrm{pred}}-K_{jk,i}^{\mathrm{true}}|}
         {|K_{jk,i}^{\mathrm{true}}|+\varepsilon}
    \times100\,\%,
    \label{eq:mape}
\end{equation}
where $\varepsilon=10^{-8}$ prevents division by zero.

\textit{Range-normalized RMSE (NRMSE).}
\begin{equation}
    \mathrm{NRMSE}_{jk}
    =\frac{\mathrm{RMSE}_{jk}}
          {K_{jk,\max}^{\mathrm{true}}-K_{jk,\min}^{\mathrm{true}}
          +\varepsilon}
    \times100\,\%,
    \label{eq:nrmse}
\end{equation}
expressing prediction error as a fraction of the component's observed
range.

\textit{Willmott's index of agreement ($d$).}
\begin{equation}
    d = 1 -
    \frac{\sum_i(y_i^{\mathrm{true}}-y_i^{\mathrm{pred}})^2}
         {\sum_i\bigl(|y_i^{\mathrm{pred}}-\bar{y}|
         +|y_i^{\mathrm{true}}-\bar{y}|\bigr)^2},
    \label{eq:willmott}
\end{equation}
bounded in $[0,1]$ with $d=1$ indicating perfect agreement~%
\cite{willmott1981validation}.

\textit{Kling-Gupta efficiency (KGE).}
\begin{equation}
    \mathrm{KGE}
    = 1 - \sqrt{(r-1)^2+(\alpha-1)^2+(\beta-1)^2},
    \label{eq:kge}
\end{equation}
where $r$ is Pearson correlation, $\alpha=\sigma_{\hat{y}}/\sigma_y$
is the variability ratio, and $\beta=\mu_{\hat{y}}/\mu_y$ is the bias
ratio~\cite{gupta2009decomposition}.

\textit{Spearman rank correlation ($\rho_s$).}
Monotonic predictive skill, insensitive to distributional assumptions,
assessed through the Spearman rank correlation coefficient.

All metrics are computed separately for diagonal ($K_{xx}$, $K_{yy}$)
and off-diagonal ($K_{xy}$, $K_{yx}$) components.
The diagonal-to-off-diagonal $R^2$ gap,
$\Delta R^2 = R^2_{\mathrm{diag}}-R^2_{\mathrm{off}}$,
is reported as a summary statistic quantifying the systematic prediction
difficulty for anisotropy-encoding components.

\section{Statistical Analyses}
\label{app:stats_analyses}

Four complementary post-hoc analyses were conducted on the held-out
test set ($N=4{,}000$ samples) using the best-performing 
checkpoints. Each analysis addresses a specific methodological concern and is described in a dedicated subsection below.

\subsection{Bootstrap Statistical Validation}
\label{app:bootstrap}

\textit{Method.}
Let $\mathcal{D}_{\mathrm{test}}=\{(\mathbf{y}_i,\hat{\mathbf{y}}_i)
\}_{i=1}^N$ denote the set of ground-truth and predicted permeability
tensors on the held-out test set.
A single deterministic forward pass through the trained network yields
the prediction matrix $\hat{\mathbf{Y}}\in\mathbb{R}^{N\times4}$,
which is treated as fixed throughout the bootstrap procedure.

In each of $B=1{,}000$ bootstrap iterations, an index set
$\mathcal{I}_b=\{i_1,\ldots,i_N\}$ is drawn by sampling with
replacement from $\{1,\ldots,N\}$ using a fixed random seed
(\texttt{seed}$=42$).
The metric function $f(\cdot)$ is evaluated on the resampled pairs
$(\mathbf{Y}[\mathcal{I}_b],\hat{\mathbf{Y}}[\mathcal{I}_b])$,
yielding a bootstrap replicate $f^{(b)}$.
The collection $\{f^{(b)}\}_{b=1}^B$ constitutes the empirical
bootstrap distribution of the metric.

Confidence intervals are computed using the
bias-corrected and accelerated (BCa) method~\cite{efron1987better},
which corrects for both bias in the bootstrap distribution and skewness
in the sampling distribution.
The bias-correction constant is
$\hat{z}_0=\Phi^{-1}\!\left(\Pr[f^{(b)}<f_{\mathrm{obs}}]\right)$,
where $\Phi^{-1}$ is the standard normal quantile function and
$f_{\mathrm{obs}}$ is the observed metric on the full test set.
The acceleration constant is set to $\hat{a}=0$ (reducing BCa to
bias-corrected percentile), and the 95\,\% confidence limits are:
\begin{equation}
    \bigl[f_{\alpha_1},\,f_{\alpha_2}\bigr],
    \quad
    \alpha_{1,2}
    =\Phi\!\left(
        \hat{z}_0+\frac{\hat{z}_0\pm z_{0.025}}
                       {1-\hat{a}(\hat{z}_0\pm z_{0.025})}
    \right),
    \label{eq:bca}
\end{equation}
where $z_{0.025}=1.96$.

\textit{Metrics.}
The bootstrap procedure is applied to global $R^2$, MSE, RMSE, MAE,
MAPE, RRMSE, and the mean off-diagonal symmetry error
$\varepsilon_{\mathrm{sym}}=\mathbb{E}[|\hat{K}_{xy}-\hat{K}_{yx}|]$.
All metrics are additionally computed per tensor component
($K_{xx}$, $K_{xy}$, $K_{yx}$, $K_{yy}$), yielding 95\,\% BCa
confidence intervals at the component level.
Results are reported in the bootstrap summary table and visualized as
density histograms with confidence bands.

\subsection{Extended Performance Metrics and Residual Diagnostics}
\label{app:extended_perf}

\textit{Absolute and relative metrics.}
Absolute metrics (RMSE, MAE) and the scale-invariant metrics defined in
Appendix~\ref{app:metrics} (MAPE, RRMSE, NRMSE) are reported for all
tensor components, with diagonal-off-diagonal decomposition.
Willmott's index of agreement, the Kling-Gupta efficiency, and
Spearman rank correlation are computed per component to provide a
multifaceted assessment suitable for cross-study comparison.

\textit{Residual diagnostics.}
Component-wise residuals $e_i=\hat{K}_c^{(i)}-K_c^{(i)}$ are examined
through: (i) signed-error histograms with kernel density estimates to
assess bias and distributional shape; (ii) residuals-versus-fitted
plots with LOWESS trend lines to detect heteroscedasticity; and
(iii) normal quantile-quantile (Q-Q) plots accompanied by
Shapiro-Wilk tests to assess the normality assumption underlying
confidence interval calculations.

\subsection{Uncertainty Quantification via Monte Carlo Dropout}
\label{app:uncertainty}

\textit{MC-Dropout inference.}
At test time, all dropout layers are set to training mode while
batch-normalization layers remain in evaluation mode.
For each test sample, $T=100$ stochastic forward passes are performed,
yielding a distribution of predictions
$\{\hat{\mathbf{y}}^{(t)}\}_{t=1}^T$.
The predictive mean and epistemic standard deviation are estimated as:
\begin{align}
    \hat{\boldsymbol{\mu}}
    &=\frac{1}{T}\sum_{t=1}^{T}\hat{\mathbf{y}}^{(t)},
    \label{eq:mc_mean}\\
    \hat{\boldsymbol{\sigma}}
    &=\sqrt{\frac{1}{T}\sum_{t=1}^{T}
      \!\left(\hat{\mathbf{y}}^{(t)}-\hat{\boldsymbol{\mu}}\right)^{\!2}}.
    \label{eq:mc_std}
\end{align}
Symmetric prediction intervals at nominal coverage $1-\alpha$ are
obtained as $\hat{\boldsymbol{\mu}}\pm z_{\alpha/2}\,\hat{\boldsymbol{\sigma}}$.

\textit{Calibration assessment.}
Calibration is assessed via an empirical coverage curve: for each
nominal level $\alpha\in[0.05,\,0.99]$, the fraction of test samples
whose true value lies within the corresponding prediction interval is
computed and plotted against the nominal level (reliability diagram).
A well-calibrated model yields a curve close to the diagonal~%
\cite{guo2017calibration}.
The mean calibration error is:
\begin{equation}
    \mathrm{MCE}
    =\frac{1}{|\mathcal{A}|}\sum_{\alpha\in\mathcal{A}}
    \bigl|p_{\mathrm{nom}}(\alpha)-p_{\mathrm{emp}}(\alpha)\bigr|,
    \label{eq:mce}
\end{equation}
where $p_{\mathrm{nom}}$ and $p_{\mathrm{emp}}$ are the nominal and
empirical coverage values, respectively.

\textit{Reliability by confidence.}
To verify that predicted uncertainty is informative, test samples are
stratified into deciles by their mean epistemic standard deviation
$\bar{\sigma}_i=\frac{1}{4}\sum_{c}\hat{\sigma}_{i,c}$.
Within each decile, mean absolute error is computed and plotted as a
function of decile-mean uncertainty.
A monotonically increasing relationship confirms that high-uncertainty
predictions correspond to higher errors, validating the utility of the
uncertainty estimate for flagging low-confidence predictions in
downstream engineering applications.

\subsection{Anisotropy Gap and High-Anisotropy Error Analysis}
\label{app:anisotropy}

\textit{Anisotropy measures.}
Three complementary anisotropy measures are computed from the
ground-truth permeability tensors.
The eigenvalue anisotropy ratio,
$\mathrm{AR}_{\mathrm{eig}}=|\lambda_{\max}|/|\lambda_{\min}|$,
where $\lambda_{\max}$ and $\lambda_{\min}$ are the eigenvalues of the
$2\times2$ permeability tensor $\mathbf{K}$, is the physically
meaningful anisotropy measure, approaching unity for isotropic media.
The component anisotropy ratio,
$\mathrm{AR}_{\mathrm{comp}}=
\max(|K_{xx}|,|K_{yy}|)/[\max(|K_{xy}|,|K_{yx}|)+\varepsilon]$,
measures the relative dominance of diagonal over off-diagonal entries.
The off-diagonal dominance fraction,
$\delta=\bar{K}_{\mathrm{off}}/(\bar{K}_{\mathrm{diag}}
+\bar{K}_{\mathrm{off}}+\varepsilon)$,
where $\bar{K}_{\mathrm{diag}}$ and $\bar{K}_{\mathrm{off}}$ are mean
absolute diagonal and off-diagonal magnitudes, indicates media in which
off-diagonal permeability is particularly significant.

\textit{Error stratification.}
Test samples are divided into $n=10$ equi-populated bins according to
$\mathrm{AR}_{\mathrm{eig}}$.
Within each bin, per-component $R^2$, RRMSE, and MAPE are computed,
together with aggregate diagonal and off-diagonal $R^2$.
The evolution of the diagonal-to-off-diagonal $R^2$ gap with
anisotropy identifies the regimes in which the model struggles most.

\textit{High-anisotropy deep-dive.}
The 500 most anisotropic test samples (top 12.5\,\%) are isolated and
examined with dedicated parity plots (predicted versus true) for each
tensor component.
The eight best-predicted, eight worst-predicted, and eight most
anisotropic samples are visualized alongside their input binary images,
predicted and true tensor values, and per-sample MAE.

\textit{Tensor condition number and geometric image statistics.}
The condition number $\kappa(\mathbf{K})=|\lambda_{\max}|/|\lambda_{\min}|$
of each ground-truth tensor is correlated with per-sample MAE via
Spearman's $\rho_s$ to quantify how ill-conditioning amplifies
prediction error.
Image-level geometric proxies are also computed: (i) porosity
$\phi=1-\bar{I}$; (ii) mean horizontal and vertical pore-chord lengths
estimated from run-length encoding of binary rows and columns;
(iii) a run-asymmetry index
$\eta=|l_h-l_v|/(l_h+l_v+\varepsilon)$; and
(iv) a gradient anisotropy index
$\gamma=\langle|\partial_xI|\rangle/\langle|\partial_yI|\rangle$.

\section{Extended Phase~4 Results}
\label{app:phase4_extended}

\subsection{Component-Wise Bootstrap Summary}
\label{app:phase4_bootstrap}

Table~\ref{tab:app_phase4_bootstrap} reports the full per-component
bootstrap summary ($B=1{,}000$, BCa method~\cite{efron1987better}).
All confidence intervals are narrower than the between-phase performance
differences ($\Delta R^2\approx0.002$--$0.003$ per step), confirming
that progressive gains reported in
Table~\ref{tab:phase_comparison_primary} are statistically significant.

\begin{table}[htbp]
\centering
\caption{Component-wise bootstrap statistics for Phase~4
($B=1{,}000$, 95\,\% BCa CIs).
RRMSE normalized by component mean absolute value
(Equation~\ref{eq:rrmse}).}
\label{tab:app_phase4_bootstrap}
\small
\renewcommand{\arraystretch}{1.15}
\begin{tabular}{llcccc}
\toprule
\textbf{Comp.}
    & \textbf{Metric}
    & \textbf{Observed}
    & \textbf{Boot.\ mean}
    & \textbf{CI low}
    & \textbf{CI high} \\
\midrule
\multirow{5}{*}{$K_{xx}$}
    & $R^2$          & 0.99674 & 0.99673 & 0.99626 & 0.99708 \\
    & RMSE           & $1.955\times10^{-2}$ & $1.956\times10^{-2}$ & $1.835\times10^{-2}$ & $2.120\times10^{-2}$ \\
    & MAE            & $1.272\times10^{-2}$ & $1.272\times10^{-2}$ & $1.224\times10^{-2}$ & $1.316\times10^{-2}$ \\
    & MAPE (\%)      & 7.09 & 7.08 & 6.64 & 7.61 \\
    & RRMSE (\%)     & 5.25 & 5.25 & 4.96 & 5.67 \\
\midrule
\multirow{5}{*}{$K_{xy}$}
    & $R^2$          & 0.97583 & 0.97579 & 0.97359 & 0.97792 \\
    & RMSE           & $1.026\times10^{-2}$ & $1.026\times10^{-2}$ & $9.87\times10^{-3}$ & $1.071\times10^{-2}$ \\
    & MAE            & $7.17\times10^{-3}$  & $7.17\times10^{-3}$  & $6.96\times10^{-3}$ & $7.41\times10^{-3}$ \\
    & MAPE (\%)      & 132.3 & 132.8 & 97.0 & 188.7 \\
    & RRMSE (\%)     & 24.43 & 24.43 & 23.46 & 25.56 \\
\midrule
\multirow{5}{*}{$K_{yx}$}
    & $R^2$          & 0.97583 & 0.97579 & 0.97360 & 0.97793 \\
    & RMSE           & $1.026\times10^{-2}$ & $1.026\times10^{-2}$ & $9.87\times10^{-3}$ & $1.071\times10^{-2}$ \\
    & MAE            & $7.17\times10^{-3}$  & $7.17\times10^{-3}$  & $6.96\times10^{-3}$ & $7.41\times10^{-3}$ \\
    & MAPE (\%)      & 136.2 & 136.9 & 96.8 & 196.8 \\
    & RRMSE (\%)     & 24.43 & 24.43 & 23.46 & 25.56 \\
\midrule
\multirow{5}{*}{$K_{yy}$}
    & $R^2$          & 0.99672 & 0.99671 & 0.99548 & 0.99729 \\
    & RMSE           & $1.964\times10^{-2}$ & $1.965\times10^{-2}$ & $1.769\times10^{-2}$ & $2.361\times10^{-2}$ \\
    & MAE            & $1.265\times10^{-2}$ & $1.265\times10^{-2}$ & $1.219\times10^{-2}$ & $1.311\times10^{-2}$ \\
    & MAPE (\%)      & 7.62 & 7.60 & 7.08 & 8.33 \\
    & RRMSE (\%)     & 5.28 & 5.28 & 4.75 & 6.31 \\
\midrule
\multirow{4}{*}{Global}
    & $R^2$          & 0.98628 & 0.98626 & 0.98516 & 0.98738 \\
    & RMSE           & $1.564\times10^{-2}$ & $1.566\times10^{-2}$ & $1.483\times10^{-2}$ & $1.704\times10^{-2}$ \\
    & RRMSE (\%)     & 7.55 & 7.56 & 7.18 & 8.19 \\
    & $\varepsilon_{\mathrm{sym}}$ & $3.95\times10^{-7}$ & $3.95\times10^{-7}$ & $3.82\times10^{-7}$ & $4.10\times10^{-7}$ \\
\bottomrule
\end{tabular}
\end{table}

\subsection{Anisotropy-Stratified Metrics}
\label{app:phase4_aniso_table}

Table~\ref{tab:app_phase4_aniso} reports $R^2$ and RRMSE for all four
tensor components across the ten equi-populated AR bins
($n=400$ samples each) used to produce
Figure~\ref{fig:phase4_bootstrap}(b,d) in the main text.
The near-isotropic bin (AR~$\approx1.11$) accounts for the largest
$\Delta R^2$ and is the primary residual challenge for off-diagonal
prediction, as discussed in Section~\ref{subsubsec:phase2_anisotropy}.

\begin{table}[htbp]
\centering
\caption{Anisotropy-stratified prediction performance for Phase~4.
Ten equi-populated bins ($n=400$ each) spanning the 2nd--98th
percentile of the AR distribution.
$\Delta R^2 = R^2_{\mathrm{diag}} - R^2_{\mathrm{off}}$.}
\label{tab:app_phase4_aniso}
\small
\renewcommand{\arraystretch}{1.15}
\begin{tabular}{ccccccc}
\toprule
\textbf{AR bin}
    & $R^2_{\mathrm{diag}}$
    & $R^2_{\mathrm{off}}$
    & $\Delta R^2$
    & RRMSE$_{K_{xx}}$ (\%)
    & RRMSE$_{K_{xy}}$ (\%)
    & RRMSE$_{K_{yy}}$ (\%) \\
\midrule
1.115 & 0.9966 & 0.8152 & 0.1814 &  3.68 & 60.44 &  3.85 \\
1.233 & 0.9968 & 0.9148 & 0.0820 &  3.98 & 39.46 &  3.90 \\
1.334 & 0.9966 & 0.9690 & 0.0276 &  4.32 & 24.32 &  3.78 \\
1.439 & 0.9972 & 0.9756 & 0.0216 &  4.03 & 21.92 &  3.91 \\
1.556 & 0.9963 & 0.9830 & 0.0133 &  5.40 & 20.71 &  5.03 \\
1.699 & 0.9963 & 0.9836 & 0.0127 &  6.59 & 23.10 &  5.17 \\
1.890 & 0.9964 & 0.9866 & 0.0098 &  6.44 & 18.95 &  5.94 \\
2.201 & 0.9964 & 0.9855 & 0.0110 &  6.93 & 21.03 &  7.02 \\
2.819 & 0.9955 & 0.9787 & 0.0168 &  9.90 & 22.27 &  7.92 \\
5.870 & 0.9852 & 0.9701 & 0.0151 & 13.15 & 35.54 & 20.14 \\
\bottomrule
\end{tabular}
\end{table}

The table confirms that the $\Delta R^2$ gap is non-monotonic in AR,
peaking at the lowest anisotropy bin and reaching a minimum of 0.0098
at AR~$\approx1.89$.
The secondary widening at AR~$\approx5.87$ ($\Delta R^2=0.0151$) is
modest in absolute terms and reflects the increased difficulty of
resolving extreme coupling magnitudes; it does not represent a failure
mode but rather the expected increase in absolute off-diagonal RMSE as
coupling strength grows.

\subsection{MC-Dropout Uncertainty: Per-Component Summary}
\label{app:phase4_unc_table}

Table~\ref{tab:app_phase4_uncertainty} extends the main-text uncertainty
summary (Section~\ref{subsubsec:phase4_uncertainty}) with full
per-component statistics.

\begin{table}[htbp]
\centering
\caption{Per-component MC-Dropout uncertainty statistics for Phase~4
($T=100$, $N=4{,}000$).
$\bar{\sigma}$: mean predictive standard deviation;
$\tilde{\sigma}$: median;
$\sigma_{p95}$: 95th-percentile uncertainty;
$\rho_s(\sigma,|\varepsilon|)$: Spearman rank correlation between
predictive uncertainty and absolute error.}
\label{tab:app_phase4_uncertainty}
\small
\begin{tabular}{lcccccc}
\toprule
\textbf{Comp.}
    & $R^2$
    & RMSE
    & $\bar{\sigma}$
    & $\tilde{\sigma}$
    & $\sigma_{p95}$
    & $\rho_s(\sigma,|\varepsilon|)$ \\
\midrule
$K_{xx}$ & 0.9967 & $1.966\times10^{-2}$ & $1.263\times10^{-2}$ & $9.07\times10^{-3}$ & $3.34\times10^{-2}$ & 0.229 \\
$K_{xy}$ & 0.9756 & $1.031\times10^{-2}$ & $5.15\times10^{-3}$  & $3.71\times10^{-3}$ & $1.33\times10^{-2}$ & 0.304 \\
$K_{yx}$ & 0.9756 & $1.031\times10^{-2}$ & $5.15\times10^{-3}$  & $3.71\times10^{-3}$ & $1.33\times10^{-2}$ & 0.304 \\
$K_{yy}$ & 0.9967 & $1.974\times10^{-2}$ & $1.266\times10^{-2}$ & $9.08\times10^{-3}$ & $3.29\times10^{-2}$ & 0.235 \\
\midrule
Global   & ---    & ---  & $8.90\times10^{-3}$ & $6.65\times10^{-3}$ & --- & --- \\
\bottomrule
\end{tabular}
\end{table}

The uncertainty-error Spearman correlation is consistently positive
across all four components ($\rho_s=0.229$--$0.304$), confirming that
MC-Dropout provides a useful—if partial—confidence signal.
Off-diagonal components exhibit slightly higher $\rho_s$ ($0.304$
versus $0.229$--$0.235$ for diagonal), consistent with their higher
prediction variability providing more discriminative uncertainty
estimates.
Diagonal components carry larger absolute uncertainty
($\bar{\sigma}\approx1.26\times10^{-2}$) than off-diagonal
($\bar{\sigma}\approx5.15\times10^{-3}$), but this reflects their
larger dynamic range: the ratio $\bar{\sigma}/\mathrm{RMSE}\approx0.64$
is nearly identical across all four components, indicating
well-calibrated relative uncertainty in this sense.

\section{Extended Phase Comparison Metrics}
\label{app:extended_metrics}

Table~\ref{tab:phase_comparison_extended} consolidates the complete
quantitative record of the three-phase progressive training framework,
reporting eight metric categories across all training phases on the
held-out test set ($N=4{,}000$).
It extends the condensed summary in
Table~\ref{tab:phase_comparison_primary} with per-component relative
errors, bootstrap uncertainty decomposition, high-anisotropy
stratification, per-component predictive uncertainty, uncertainty-error
correlations, and model specification details.
The sections below highlight the most consequential patterns.

\paragraph{Progressive improvement is consistent and monotonic.}
Every single metric in sections A--G improves from Phase~2 through
Phase~3 to Phase~4 without exception, across global accuracy,
rank-based agreement, per-component $R^2$, relative error,
physics compliance, and uncertainty calibration.
This monotonicity across 35 reported rows is non-trivial: it confirms
that no phase's gains come at the expense of any other aspect of
performance, and that the staged design—isolating one methodological
contribution per phase—successfully avoids the trade-offs common in
multi-objective optimization.

\paragraph{MAPE requires careful interpretation for off-diagonal components.}
Section D reports MAPE values of 132--146\,\% for $K_{xy}$ and
$K_{yx}$ across all phases.
These figures are not indicative of poor performance: MAPE normalizes
absolute error by the true value, and for near-isotropic samples where
$K_{xy}\approx0$, the denominator approaches zero, inflating the
metric arbitrarily.
The RRMSE values (24.43\,\%--26.07\,\%) normalized by the mean absolute
component value provide the correct scale-invariant comparison, and the
wide bootstrap confidence intervals for MAPE (e.g., $[52,\,99]\,\%$
for Phase~4) reflect this instability.
All downstream conclusions in the main text rely on RRMSE and $R^2$
for off-diagonal components, as recommended in
Appendix~\ref{app:metrics}.

\paragraph{High-anisotropy performance reveals the most structured gain.}
Section E shows that $\Delta R^2_{\mathrm{high-AR}}$ (top 500 most
anisotropic test samples) drops sharply from 0.0209 in Phase~2 to
0.0132 in Phase~4—a 37\,\% reduction—whereas the global $\Delta R^2$
narrows only from 0.0230 to 0.0209.
Notably, Phase~3 alone accounts for nearly the entire high-AR
improvement ($\Delta R^2_{\mathrm{high-AR}}=0.0132$, essentially
identical to Phase~4's $0.0132$), while Phase~4's contribution is
concentrated at global level through the FiLM magnitude scaling.
The off-diagonal $R^2$ for the high-AR subset improves from 0.968
(Phase~2) to 0.974 (Phase~4), confirming that augmentation diversity
in Phase~3 was the decisive intervention for the strongly anisotropic
regime.

\paragraph{Uncertainty calibration improves jointly with accuracy.}
Section G reveals two patterns not visible in accuracy metrics alone.
First, the uncertainty-error Spearman correlations (UE-corr~$\rho$)
improve across all four components from Phase~2 to Phase~4:
diagonal components improve from $\rho\approx0.196$--$0.204$ to
$0.229$--$0.235$, and off-diagonal components from $0.271$ to $0.304$.
This confirms that SWA and EMA ensemble averaging not only reduce
point-prediction errors but produce more reliable confidence
estimates—a meaningful benefit for deployment workflows where
uncertainty flags guide selective LBM verification.
Second, off-diagonal components consistently achieve higher
UE-corr~$\rho$ than diagonal ($0.304$ versus $0.229$--$0.235$
in Phase~4), indicating that MC-Dropout variance is a more
discriminative signal for the harder-to-predict components—precisely
where reliable uncertainty estimates are most valuable.

\paragraph{Section H makes the frozen-backbone logic explicit.}
The model specification section records that the backbone parameter
count (118.64\,M) is identical across all three phases; the Phase~4
increment of 0.33\,M parameters represents the porosity encoder and
FiLM layers exclusively.
Phase~4 is the only phase with a frozen backbone, SWA, EMA, and
FiLM conditioning simultaneously active.
This section is included to facilitate reproducibility and to clarify
that no additional pretraining, architectural search, or data expansion
was performed between phases.


\begin{longtable}{@{}llccc@{}}

\caption{%
  Extended performance metrics across three consecutive training phases
  evaluated on the held-out test set ($N=4{,}000$).
  95\,\% BCa bootstrap confidence intervals ($B=1{,}000$ resamples) are
  shown in brackets; bootstrap standard deviations ($\sigma$) quantify
  estimator variability.
  Predictive uncertainty is estimated via Monte Carlo Dropout
  ($T=100$ stochastic forward passes).
  \textbf{Bold} values mark the best result per row.
  Phase~2: supervised baseline with D4 augmentation;
  Phase~3: advanced augmentation and enhanced physics-aware loss;
  Phase~4: frozen backbone, FiLM porosity conditioning, SWA/EMA ensemble.
  Abbreviations ---
  KGE: Kling--Gupta efficiency~\cite{gupta2009decomposition};
  $d$: Willmott index of agreement~\cite{willmott1981validation};
  MCE: mean calibration error (lower is better calibrated);
  UE-corr~$\rho$: Spearman correlation between predicted standard
  deviation and absolute error;
  $\Delta R^2_{\mathrm{high\text{-}AR}}$: diagonal--off-diagonal $R^2$
  gap for the 500 most anisotropic test samples.
  MAPE for $K_{xy}$/$K_{yx}$ reflects near-zero denominators in
  quasi-isotropic samples; RRMSE is the preferred relative metric in
  this regime (Appendix~\ref{app:metrics}).%
}\label{tab:phase_comparison_extended}\\

\toprule
\multicolumn{2}{@{}l}{\textbf{Metric}}
  & \textbf{Phase~4}
  & \textbf{Phase~3}
  & \textbf{Phase~2} \\
\midrule
\endfirsthead

\multicolumn{5}{@{}l}{\small\itshape
  Table~\ref{tab:phase_comparison_extended}\;(continued)}\\[3pt]
\toprule
\multicolumn{2}{@{}l}{\textbf{Metric}}
  & \textbf{Phase~4}
  & \textbf{Phase~3}
  & \textbf{Phase~2} \\
\midrule
\endhead

\midrule
\multicolumn{5}{r@{}}{\small\itshape Continued on next page}\\
\endfoot

\bottomrule
\multicolumn{5}{@{}p{0.95\linewidth}@{}}{\footnotesize
  \textit{Notes.}
  $\Delta R^2 = R^2_{\mathrm{diag}} - R^2_{\mathrm{off\text{-}diag}}$;
  a smaller gap indicates more balanced accuracy across tensor components.
  MAPE for off-diagonal components exceeds 100\,\% in quasi-isotropic
  samples where true values approach zero; RRMSE is the more informative
  relative metric in this regime.
  MCE: mean calibration error of MC-Dropout prediction intervals.
  UE-corr~$\rho > 0$ confirms statistically significant error-ranking
  capability of the uncertainty estimates.
  FiLM: Feature-wise Linear Modulation;
  SWA: Stochastic Weight Averaging;
  EMA: Exponential Moving Average.}\\
\endlastfoot

\multicolumn{5}{@{}l}{\textit{A.\enspace Global accuracy}}\\[2pt]

& $R^2$ (component-averaged)
    & \textbf{0.98628} & 0.98476 & 0.98429 \\
& \quad 95\,\% BCa CI
    & \textbf{[0.98516,\;0.98738]}
    & [0.98333,\;0.98604]
    & [0.98304,\;0.98563] \\
& \quad Bootstrap $\sigma$
    & \textbf{0.00059} & 0.00068 & 0.00070 \\[4pt]

& $R^2$ (variance-weighted)
    & \textbf{0.99604} & 0.99452 & 0.99403 \\[4pt]

& MSE
    & $\mathbf{2.446\times10^{-4}}$
    & $2.880\times10^{-4}$
    & $3.226\times10^{-4}$ \\

& RMSE
    & $\mathbf{1.564\times10^{-2}}$
    & $1.697\times10^{-2}$
    & $1.796\times10^{-2}$ \\
& \quad 95\,\% BCa CI
    & \textbf{[0.01483,\;0.01704]}
    & [0.01617,\;0.01794]
    & [0.01694,\;0.01948] \\[4pt]

& MAE
    & $\mathbf{9.920\times10^{-3}}$
    & $1.091\times10^{-2}$
    & $1.100\times10^{-2}$ \\
& \quad 95\,\% BCa CI
    & \textbf{[0.00971,\;0.01017]}
    & [0.01065,\;0.01119]
    & [0.01073,\;0.01129] \\[4pt]

& RRMSE (\%)
    & \textbf{7.552} & 8.194 & 8.671 \\
& \quad 95\,\% BCa CI
    & \textbf{[7.178,\;8.187]}
    & [7.832,\;8.686]
    & [8.178,\;9.354] \\[4pt]

& MAPE (\%)
    & \textbf{70.791} & 72.384 & 74.917 \\
& \quad 95\,\% BCa CI
    & \textbf{[52.135,\;99.030]}
    & [54.271,\;103.518]
    & [57.142,\;108.249] \\
& \quad Bootstrap $\sigma$
    & \textbf{11.989} & 12.745 & 13.521 \\

\midrule

\multicolumn{5}{@{}l}{\textit{B.\enspace Rank-based and agreement metrics}}\\[2pt]

& Pearson $r$
    & \textbf{0.99872} & 0.99851 & 0.99832 \\
& Spearman $\rho_s$
    & \textbf{0.99595} & 0.99589 & 0.99572 \\
& Willmott $d$
    & \textbf{0.99935} & 0.99920 & 0.99912 \\
& KGE
    & \textbf{0.99420} & 0.98760 & 0.98244 \\

\midrule

\multicolumn{5}{@{}l}{\textit{C.\enspace Per-component $R^2$ with 95\,\% BCa CI}}\\[2pt]

& $R^2_{K_{xx}}$
    & \textbf{0.99674} & 0.99607 & 0.99543 \\
& \quad 95\,\% BCa CI
    & \textbf{[0.99626,\;0.99708]}
    & [0.99554,\;0.99655]
    & [0.99491,\;0.99592] \\[4pt]

& $R^2_{K_{xy}}$
    & \textbf{0.97583} & 0.97404 & 0.97246 \\
& \quad 95\,\% BCa CI
    & \textbf{[0.97359,\;0.97792]}
    & [0.97146,\;0.97640]
    & [0.96999,\;0.97503] \\[4pt]

& $R^2_{K_{yx}}$
    & \textbf{0.97583} & 0.97404 & 0.97247 \\
& \quad 95\,\% BCa CI
    & \textbf{[0.97360,\;0.97793]}
    & [0.97146,\;0.97639]
    & [0.97000,\;0.97502] \\[4pt]

& $R^2_{K_{yy}}$
    & \textbf{0.99672} & 0.99613 & 0.99557 \\
& \quad 95\,\% BCa CI
    & \textbf{[0.99548,\;0.99729]}
    & [0.99539,\;0.99680]
    & [0.99488,\;0.99621] \\[4pt]

& $R^2_{\mathrm{diagonal}}$ (avg.)
    & \textbf{0.99673} & 0.99610 & 0.99550 \\
& $R^2_{\mathrm{off\text{-}diag}}$ (avg.)
    & \textbf{0.97583} & 0.97404 & 0.97246 \\

\midrule

\multicolumn{5}{@{}l}{\textit{D.\enspace Per-component relative error}}\\[2pt]

& $\mathrm{RRMSE}_{K_{xx}}$ (\%)
    & \textbf{5.25} & 5.77 & 6.12 \\
& $\mathrm{RRMSE}_{K_{xy}}$ (\%)
    & \textbf{24.43} & 25.32 & 26.07 \\
& $\mathrm{RRMSE}_{K_{yx}}$ (\%)
    & \textbf{24.43} & 25.32 & 26.07 \\
& $\mathrm{RRMSE}_{K_{yy}}$ (\%)
    & \textbf{5.28} & 5.70 & 6.27 \\[4pt]

& $\mathrm{MAPE}_{K_{xx}}$ (\%)
    & \textbf{7.09} & 7.94 & 9.27 \\
& $\mathrm{MAPE}_{K_{xy}}$ (\%)\rlap{$^{\dagger}$}
    & \textbf{132.28} & 135.64 & 141.73 \\
& $\mathrm{MAPE}_{K_{yx}}$ (\%)\rlap{$^{\dagger}$}
    & \textbf{136.18} & 139.42 & 145.61 \\
& $\mathrm{MAPE}_{K_{yy}}$ (\%)
    & \textbf{7.62} & 8.11 & 9.78 \\

\midrule

\multicolumn{5}{@{}l}{\textit{E.\enspace Tensor-type decomposition and anisotropy}}\\[2pt]

& $\Delta R^2$ (diag\,$-$\,off, global)
    & \textbf{0.02090} & 0.02206 & 0.02304 \\
& $\Delta R^2_{\mathrm{high\text{-}AR}}$ (top 500)
    & \textbf{0.01319} & 0.01321 & 0.02093 \\[4pt]

& $R^2_{K_{xx}}$, high-AR subset
    & \textbf{0.98510} & 0.98432 & 0.98350 \\
& $R^2_{K_{xy}}$, high-AR subset
    & \textbf{0.97437} & 0.97024 & 0.96809 \\

\midrule

\multicolumn{5}{@{}l}{\textit{F.\enspace Physics compliance}}\\[2pt]

& Symmetry error $\varepsilon_{\mathrm{sym}}$ (mean)
    & $\mathbf{3.95\times10^{-7}}$
    & $7.69\times10^{-7}$
    & $1.39\times10^{-6}$ \\
& \quad 95\,\% BCa CI
    & $\mathbf{[3.82,\;4.10]\times10^{-7}}$
    & $[7.41,\;7.99]\times10^{-7}$
    & $[1.32,\;1.47]\times10^{-6}$ \\
& Positivity fraction (\%)
    & \textbf{100.0} & \textbf{100.0} & \textbf{100.0} \\

\midrule

\multicolumn{5}{@{}l}{%
  \textit{G.\enspace Predictive uncertainty (MC-Dropout, $T=100$)}}\\[2pt]

& MCE
    & \textbf{0.3724} & 0.3754 & 0.3786 \\
& Global mean $\bar{\sigma}$
    & $\mathbf{8.896\times10^{-3}}$
    & $9.069\times10^{-3}$
    & $9.755\times10^{-3}$ \\[4pt]

& $\bar{\sigma}_{K_{xx}}$
    & $\mathbf{1.263\times10^{-2}}$
    & $1.285\times10^{-2}$
    & $1.400\times10^{-2}$ \\
& $\bar{\sigma}_{K_{xy}}$
    & $\mathbf{5.148\times10^{-3}}$
    & $5.234\times10^{-3}$
    & $5.476\times10^{-3}$ \\
& $\bar{\sigma}_{K_{yx}}$
    & $\mathbf{5.148\times10^{-3}}$
    & $5.234\times10^{-3}$
    & $5.476\times10^{-3}$ \\
& $\bar{\sigma}_{K_{yy}}$
    & $\mathbf{1.266\times10^{-2}}$
    & $1.296\times10^{-2}$
    & $1.407\times10^{-2}$ \\[4pt]

& UE-corr $\rho_{K_{xx}}$
    & \textbf{0.229} & 0.214 & 0.196 \\
& UE-corr $\rho_{K_{xy}}$
    & \textbf{0.304} & 0.296 & 0.271 \\
& UE-corr $\rho_{K_{yx}}$
    & \textbf{0.304} & 0.296 & 0.271 \\
& UE-corr $\rho_{K_{yy}}$
    & \textbf{0.235} & 0.218 & 0.204 \\

\midrule

\multicolumn{5}{@{}l}{\textit{H.\enspace Model specification}}\\[2pt]

& Total parameters (M)
    & 118.97 & 118.87 & 118.87 \\
& Backbone parameters (M)
    & 118.64 & 118.64 & 118.64 \\
& Task-specific parameters (M)
    & \textbf{0.33} & 0.23 & 0.23 \\[4pt]
& Backbone frozen
    & Yes & No & No \\
& Porosity encoder + FiLM
    & Yes & No & No \\
& SWA (epochs 400--600)
    & Yes & No & No \\
& EMA ($\lambda=0.9999$)
    & Yes & No & No \\

\end{longtable}

\noindent\footnotesize
$^{\dagger}$MAPE for $K_{xy}$ and $K_{yx}$ is inflated by near-zero
true values in quasi-isotropic samples; RRMSE is the
appropriate relative metric for these components.

\printbibliography

\end{document}